\documentclass[10pt,twocolumn,letterpaper]{article}

\usepackage{cvpr}
\usepackage{times}
\usepackage{epsfig}
\usepackage{graphicx}
\usepackage{amsmath}
\usepackage{amssymb}

\usepackage{xcolor}
\usepackage{multirow}
\usepackage{bbding}
\usepackage[utf8]{inputenc}

\DeclareMathOperator*{\argmax}{arg\,max}
\DeclareMathOperator*{\argmin}{arg\,min}

\usepackage{diagbox}
\usepackage{textcomp}
\usepackage{booktabs}
\usepackage{colortbl}

\usepackage{comment}
\usepackage{enumitem}
\usepackage{mathrsfs}

\usepackage[outline]{contour}
\usepackage{amssymb}
\usepackage{framed}
\usepackage{comment}
\usepackage{stfloats}

\usepackage[pagebackref=true,breaklinks=true,letterpaper=true,colorlinks,bookmarks=false]{hyperref}

 \cvprfinalcopy 

\def\cvprPaperID{121} 

\begin{document}

\title{Adversarial Ranking Attack and Defense}

\author{Mo Zhou$^{1}$, Zhenxing Niu$^2$, Le Wang$^3$, Qilin Zhang$^4$, Gang Hua$^5$\\
{\normalsize $^1$Xidian University,
$^2$Alibaba DAMO MIIL,
$^3$Xi'an Jiaotong University,
$^4$HERE Technologies,
$^5$Wormpex AI Research}\\
{\tt\small \{cdluminate,zhenxingniu,samqzhang,ganghua\}@gmail.com, lewang@xjtu.edu.cn}
}

\maketitle

\begin{abstract}

Deep Neural Network (DNN) classifiers are vulnerable
to adversarial attack, where an imperceptible perturbation
could result in misclassification.
However, the vulnerability
of DNN-based image ranking systems remains under-explored.
In this paper, we propose two attacks against deep
ranking systems, \ie, Candidate Attack and Query Attack,
that can raise or lower the rank of chosen candidates by
adversarial perturbations.
Specifically, the expected ranking
order is first represented as a set of inequalities, and then a
triplet-like objective function is designed to obtain the optimal
perturbation.
Conversely, a defense method is also proposed to improve the
ranking system robustness, which can mitigate all the proposed attacks
simultaneously.
Our adversarial ranking attacks and defense are evaluated on datasets
including MNIST, Fashion-MNIST, and Stanford-Online-Products.
Experimental results demonstrate that a typical deep ranking
system can be effectively compromised by our attacks.
Meanwhile, the
system robustness can be moderately improved with our defense.
Furthermore, the transferable and universal properties of our
adversary illustrate the possibility of realistic
black-box attack.

\end{abstract}

\section{Introduction}


Despite the successful application in computer vision tasks such as
image classification~\cite{alexnet,resnet}, Deep Neural Networks (DNNs) have been
found vulnerable to adversarial attacks. In particular, the
DNN's prediction can be arbitrarily changed by just applying an imperceptible
perturbation to the input image~\cite{l-bfgs,fgsm}. Moreover, such adversarial
attacks can effectively compromise the recent state-of-the-art DNNs such as
Inception~\cite{googlenet,inceptionv2} and ResNet~\cite{resnet}.  This poses a
serious security risk on many DNN-based applications such as face recognition,
where recognition evasion or impersonation can be easily
achieved~\cite{faceblack,phy-crime,advhat,advpattern}.

\begin{figure}[t!]
\centering
\includegraphics[width=1.0\columnwidth]{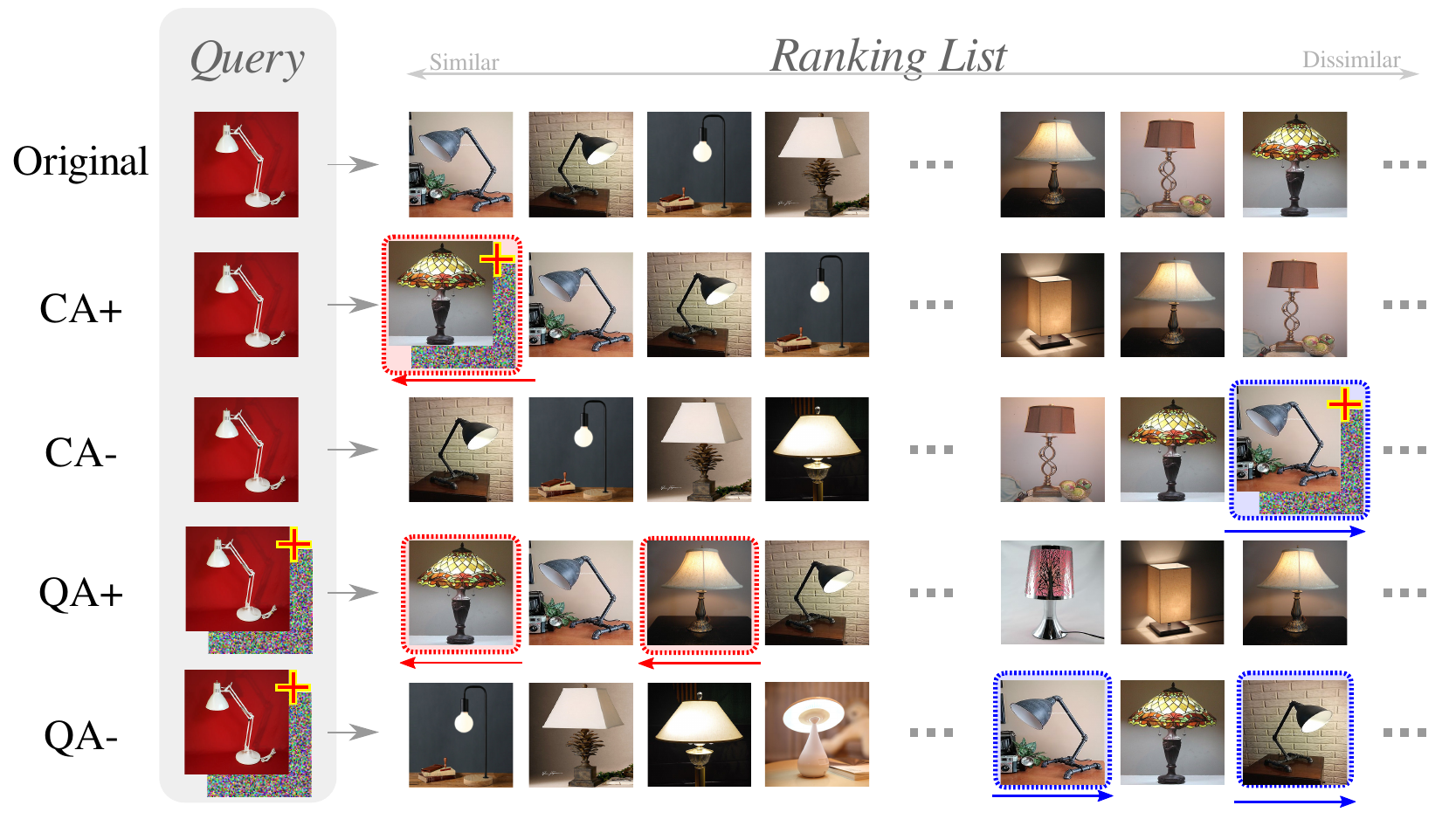}
	\caption{Adversarial ranking attack that can \emph{raise} or \emph{lower}
	the rank of chosen candidates by adversarial perturbations. In Candidate
	Attack, adversarial perturbation is added to the candidate and its rank is
	\emph{raised} (CA+) or \emph{lowered} (CA-).
	In Query Attack, adversarial perturbation is added to the query image,
	and the ranks of chosen candidates are \emph{raised} (QA+)
	or \emph{lowered} (QA-).
	}
\label{fig:advranking}
\end{figure}


Previous adversarial attacks primarily focus on \emph{classification},
however, we speculate that DNN-based image ranking systems~\cite{sbir1,imagesimilarity,imagesim2,kiros-unifying,hm-lstm,vse++,scan} also
suffer from similar vulnerability.
Taking the image-based product search as an example,
a fair ranking system should rank the database products according to their
visual similarity to the query, as shown in Fig.~\ref{fig:advranking} (row 1).
Nevertheless, a malicious seller may attempt to raise the rank of his/her
own product by adding perturbation to the image (CA+, row 2), or
lower the rank of his competitor’s product (CA-, row 3); Besides,
a ``man-in-the-middle'' attacker (\eg., a malicious advertising company) could
hijack and imperceptibly perturb the query image in order to promote
(QA+, row 4) or impede (QA-, row 5) the sales of specific products.



Unlike classification tasks where images are predicted independently,
the rank of one candidate is related to the query as well as other
candidates for image ranking. The relative relations among candidates
and queries determine the final ranking order. Therefore, we argue
that the existing adversarial classification
attacks are incompatible with the ranking scenario. Thus, we need to
thoroughly study the \emph{adversarial ranking attack}.


In this paper, adversarial ranking attack aims to \emph{raise} or
\emph{lower} the ranks of some chosen candidates
$C=\{c_1,c_2,\ldots,c_m\}$ with respect to a specific query set
$Q=\{q_1,q_2,\ldots,q_w\}$.
This can be achieved by either Candidate Attack (CA) or Query Attack (QA).
In particular, CA is defined as to raise (\emph{abbr.} CA+) or lower (\emph{abbr.} CA-)
the rank of a single candidate $c$ with respect to the query set $Q$ by perturbing $c$ itself;
while QA is defined as to raise (\emph{abbr.} QA+) or lower (\emph{abbr.} QA-) the
ranks of a candidate set $C$ with respect to a single query $q$ by perturbing $q$.
Thus, adversarial ranking attack can be achieved by performing CA on each
$c\in C$, or QA on each $q\in Q$.
In practice, the
choice of CA or QA depends on the  accessibility to the candidate or
query respectively, \ie, CA is feasible for modifiable candidate,
while QA is feasible for modifiable query.


An effective implementation of these attacks is proposed in this paper.
As we know, a typical DNN-based ranking model maps
objects (\ie, queries and candidates) to a common embedding space,
where the distances among them determine the final ranking order.
Predictably, the object's position in the embedding space will be
changed by adding a perturbation to it.
Therefore, the essential of
adversarial ranking attack is to find a proper perturbation, which
could push the object to a desired position that leads to the expected
ranking order. Specifically, we first represent the expected ranking
order as a set of inequalities. Subsequently, a triplet-like objective
function is designed according to those inequalities, and
combined with Projected Gradient Descent (PGD) to efficiently obtain
the desired adversarial perturbation.


Opposed to the proposed attacks, \emph{adversarial ranking defense}
is worth being investigated especially for security-sensitive deep ranking applications.
Until now, the Madry defense~\cite{madry} is regarded as the most effective method for
classification defense. However, we empirically discovered a primary
challenge of diverging training loss while directly adapting such
mechanism for ranking defense, possibly due to the generated
adversarial examples being too ``strong''.
In addition, such defense
mechanism needs to defend against distinct ranking attacks individually, but a
\emph{generic} defense method against all CA+, CA-, QA+ and QA-
attacks is preferred.


To this end, a shift-distance based ranking defense is proposed, which
could simultaneously defend against all attacks. Note that the
position shift of objects in the embedding space is the key for all
ranking attacks. Although different attacks prefer distinct shift
directions (\eg, CA+ and CA- often prefer opposed shifting directions), a
large shift distance is their common preference. If we could reduce
the shift distance of embeddings incurred by adversarial perturbation,
all attacks can be simultaneously defensed.
Specifically, we first propose a shift-distance based
ranking attack, which aims to push the objects as far from their
original positions as possible. And then, the adversarial examples
generated from such attack is involved in the adversarial
training. Experimental results manifest that our ranking defense can
converge and moderately improve model robustness.


In addition, our ranking attacks have some good properties for
realistic applications. First, our adversary is
transferable, \ie, the adversary obtained from a known DNN ranker can
be directly used to attack an unknown DNN ranker (\ie, the network
architecture and parameters are unknown). Second, our
attacks can be extended to \emph{universal} ranking attacks with
slight performance drop, \ie, we could learn a
\emph{universal} perturbation to all candidates for CA, or a
\emph{universal} perturbation to all queries for QA. Such properties
illustrate the possibility of practical black-box
attack.


To the best of our knowledge, this is the first work that thoroughly
studies the adversarial ranking attack and defense. In brief, our
contributions are:

\begin{enumerate}[noitemsep,leftmargin=*]
\item The adversarial ranking attack is defined and implemented, which can
intentionally change the ranking results by perturbing the candidates
or queries.
\item An adversarial ranking defense method is proposed to improve the
ranking model robustness, and mitigate all the proposed attacks
simultaneously.
\end{enumerate}



\section{Related Works}

\textbf{Adversarial Attacks.} Szegedy~\etal~\cite{l-bfgs} claimed that DNN is
susceptible to imperceptible adversarial
perturbations added to inputs, due to the intriguing ``blind
spot'' property, which was later ascribed to the local
linearity~\cite{fgsm} of neural networks. Following these findings, many
white-box (model architecture and parameters are known to the adversary) attacking
methods~\cite{deepfool,jsma,i-fgsm,cw,ead,mi-fgsm,featureadversary,onepixel,madry,spatially,hsja,fda}
are proposed to effectively compromise the state-of-the-art DNN classifiers.
Among them, PGD~\cite{madry} is regarded as one of the most powerful
attacks~\cite{pgdbeatscvpr18}.
Notably, adversarial examples are discovered to be transferable~\cite{black-frame,mltransfer}
among different neural network classifiers, which inspired a series of black-box
attacks~\cite{curlswhey,vr-igsm,di-fgsm,ensemble,ti-fgsm,ila}. 
On the other hand, universal (\ie, image-agnostic) adversarial perturbations are also
discovered~\cite{universal,unsupuap}. The existence of adversarial examples
stimulated research interests in areas such as object detection~\cite{objdet1,objdet2,objdet3},
semantic segmentation~\cite{semseg1},
and automatic speech recognition~\cite{asr1}, \etc.
It is even possible to create physical adversarial examples
~\cite{phy-crime,phy-synth,phy-robust,advpattern}.

\textbf{Deep Ranking.} Different from the traditional ``learning to
rank''~\cite{LTR,ranksvm} methods, DNN-based ranking methods often embed data
samples (including both queries and candidates) of all modalities into a common
embedding space, and subsequently determine the ranking order based on
distance.  Such workflow has been adopted in
distance metric learning~\cite{imagesimilarity,imagesim2,sop,horde},
image retrieval~\cite{sbir1},
cross-modal retrieval~\cite{hm-lstm,vse++,scan,kiros-unifying},
and face recognition~\cite{facenet}.


\textbf{Adversarial Attacks in Deep Ranking.}
For information retrieval and ranking systems, the risk of malicious users
manipulating the ranking always exists~\cite{advrank-doc,advrank-rec}.
However, only a few research efforts have been made in adversarial attacks
in deep ranking.
Liu~\etal~\cite{advrank-ut2} proposed adversarial queries
leading to incorrect retrieval results; while Li~\etal~\cite{universalret}
staged similar attack with universal perturbation that corrupts listwise ranking results.
None of the aforementioned research efforts explore the \emph{adversarial ranking attack}.
%
Besides, adaptation of distance-based attacks (\eg~\cite{featureadversary})
are unsuitable for our scenario.

\textbf{Adversarial Defenses.} Adversarial attacks
and defenses are consistently engaged in an arms race~\cite{overview1}.
Gradient masking-based defenses can be circumvented~\cite{obfuscated}.
Defensive distillation~\cite{distill1,distill2}
has been compromised by C\&W~\cite{cw,cw-pre}. As claimed in~\cite{ensembleweak},
ensemble of weak defenses are insufficient against adversarial examples.
Notably, as an early defense method~\cite{l-bfgs}, adversarial
training~\cite{fgsm,madry,advtrain,deffeatureadv,advscale,understandadvtrain,
	bilateral,advtrain-triplet,advtrainuniversal}
remains to be one of the most effective defenses.
Other types of defenses include
adversarial detection~\cite{safetynet,ondetecting},
input transformation/reconstruction/replacement~\cite{deflecting,foveation,inputtrans,magnet,nndef},
randomization~\cite{adv-bnn,self-ensemble},
feature denoising~\cite{featuredenoise},
network verification~\cite{reluplex,deepsafe},
evidential deep learning~\cite{edl}, \etc.
However, defense in deep ranking systems remains mostly uncharted.

%
%

\section{Adversarial Ranking}


Generally, a DNN-based ranking task could be formulated as a metric
learning problem. Given the query $q$ and candidate set
$X=\{c_1,c_2,\ldots,c_n\}$, deep ranking is to learn a mapping $f$
(usually implemented as a DNN) which maps all candidates and query
into a common embedding space, such that the relative distances among
the embedding vectors could satisfy the expected ranking order. For
instance, if candidate $c_i$ is more similar to the query $q$ than
candidate $c_j$, it is encouraged for the mapping $f$ to satisfy the
inequality $\|f(q)-f(c_i)\|<\|f(q)-f(c_j)\|$\footnote{Sometimes cosine distance is used instead.},
where $\|\cdot\|$ denotes $\ell_2$ norm.
For brevity, we denote $\|f(q)-f(c_i)\|$ as
$d(q,c_i)$ in following text.


Therefore, adversarial ranking attack is to find a proper adversarial
perturbation which leads the ranking order to be changed as
expected. For example, if a less relevant $c_j$ is expected to be ranked \emph{ahead}
of a relevant $c_i$, it is desired to
find a proper perturbation $r$ to perturb $c_j$, \ie
$\tilde{c}_j=c_j+r$, such that the inequality $d(q,c_i)<d(q,c_j)$
could be changed into $d(q,c_i)>d(q,\tilde{c}_j)$. In the next, we will
describe Candidate Attack and Query Attack in detail.


\subsection{Candidate Attack}


Candidate Attack (\textbf{CA}) aims to raise (\emph{abbr.} \textbf{CA+}) or
lower (\emph{abbr.} \textbf{CA-}) the rank of a \emph{single} candidate $c$ with respect to
a set of queries $Q=\{q_1,q_2,\ldots,q_w\}$ by adding perturbation $r$
to the candidate itself, \ie $\tilde{c}=c+r$.



Let $\text{Rank}_X(q,c)$ denote the rank of the candidate $c$ with respect to
the query $q$, where $X$ indicates the set of all candidates, and a smaller
rank value represents a higher ranking. Thus,
the \textbf{CA+} that {\it raises} the rank of $c$ with respect to every query $q\in Q$
by perturbation $r$ could be formulated as the following problem,
\begin{align}
	r &= \argmin_{r\in\Gamma}\sum_{q\in Q}\text{Rank}_X(q,c+r), \label{eq:ca_intuitive} \\
	\Gamma &= \{r \big| \|r\|_\infty\leqslant\varepsilon; r,c+r\in [0,1]^N \}, \label{eq:gamma}
\end{align}
where $\Gamma$ is a $\ell_\infty$-bounded $\varepsilon$-neighbor of $c$,
$\varepsilon \in [0,1]$ is a predefined small positive constant,
the constraint $\|r\|_\infty \leqslant \varepsilon$ limits the perturbation $r$
to be ``visually imperceptible'', and $c+r\in [0,1]^N$ ensures the adversarial example
remains a valid input image. Although alternative ``imperceptible'' constraints exist
(\eg, $\ell_0$~\cite{onepixel,sparse}, $\ell_1$~\cite{ead} and
$\ell_2$~\cite{cw,deepfool} variants),
we simply follow~\cite{fgsm,i-fgsm,madry} and use the $\ell_\infty$ constraint.



However, the optimization problem Eq.~\eqref{eq:ca_intuitive}--\eqref{eq:gamma}
cannot be directly solved due to the discrete nature of the rank value $\text{Rank}_X(q,c)$.
In order to solve the problem, a surrogate objective function is needed.


In metric learning, given two candidates $c_p, c_n \in X$ where $c_p$ is ranked
ahead of $c_n$, \ie $\text{Rank}_X(q,c_p) < \text{Rank}_X(q,c_n)$, the ranking
order is represented as an inequality $d(q,c_p)<d(q,c_n)$ and formulated
in triplet loss:
\begin{equation}
	L_{\text{triplet}}(q,c_p,c_n)=\left[\beta + d(q,c_p) - d(q,c_n)\right]_+,
	\label{eq:triplet}
\end{equation}
where $[\cdot]_+$ denotes $\max(0,\cdot)$, and $\beta$ is
a manually defined constant margin.
This function is known as the triplet (\ie pairwise ranking)
loss~\cite{imagesimilarity,facenet}.


Similarly, the attacking goal of \textbf{CA+} in Eq.~\eqref{eq:ca_intuitive} can
be readily converted into a series of inequalities, and subsequently turned
into a sum of triplet losses,
\begin{equation}
	L_{\text{CA+}}(c,Q;X)=\sum_{q\in Q}\sum_{x\in X}\big[d(q,c)-d(q,x)\big]_{+}.
	\label{eq:ca+_loss}
\end{equation}
In this way, the original problem in Eq.~\eqref{eq:ca_intuitive}--\eqref{eq:gamma}
can be reformulated into the following constrained optimization problem:
\begin{equation}
	r=\argmin_{r\in\Gamma}L_{\text{CA+}}(c+r,Q;X).
	\label{eq:ca_opt}
\end{equation}

%
%
%
%




To solve the optimization problem, Projected
Gradient Descent (PGD) method~\cite{madry,i-fgsm} (\emph{a.k.a}
the iterative version of FGSM~\cite{fgsm}) can be used.
Note that PGD is one of the most effective first-order
gradient-based algorithms~\cite{pgdbeatscvpr18}, popular among related works
about adversarial attack.

Specifically, in order to find an adversarial perturbation $r$ to create a desired
adversarial candidate $\tilde{c}=c+r$,
the PGD algorithm alternates two steps at every iteration $t=1, 2,\ldots,\eta$.
Step one updates $\tilde{c}$ according to the gradient of Eq.~\eqref{eq:ca+_loss};
while step two clips the result of step one to fit in the $\varepsilon$-neighboring region $\Gamma$:
\begin{equation}
	\tilde{c}_{t+1}=\text{Clip}_{c,\Gamma}\big\{\tilde{c}_{t}
	-\alpha\text{sign}(\nabla_{\tilde{c}_t}L_{\text{CA+}}(\tilde{c}_t,Q,X))\big\}, \label{eq:PGDsteps}
\end{equation}
where $\alpha$ is a constant hyper-parameter indicating the PGD step size,
and $\tilde{c}_1$ is initialized as $c$.
After $\eta$ iterations, the desired adversarial candidate $\tilde{c}$ is obtained as $\tilde{c}_{\eta}$,
which is optimized to satisfy as many inequalities
as possible. Each inequality represents a pairwise ranking sub-problem,
hence the adversarial candidate $\tilde{c}$ will be ranked ahead of
other candidates with respect to every specified query $q\in Q$.



Likewise, the \textbf{CA-} that {\it lowers} the rank of a candidate
$c$ with respect to a set of queries $Q$ can be obtained in similar way:
\begin{equation}
L_{\text{CA-}}(c,Q;X)=\sum_{q\in Q}\sum_{x\in X}\big[-d(q,c)+d(q,x)\big]_{+}.
	\label{eq:ca-_loss}
\end{equation}

\subsection{Query Attack}


Query Attack (\textbf{QA}) is supposed to raise (\emph{abbr.} \textbf{QA+}) or lower (\emph{abbr.} \textbf{QA-})
the rank of a set of candidates
$C=\{c_1,c_2,\ldots,c_m\}$ with respect to the query $q$,
by adding adversarial perturbation $r$ to the query $\tilde{q}=q+r$.
Thus, \textbf{QA} and \textbf{CA} are two ``symmetric'' attacks.
The \textbf{QA-} for {\it lowering} the rank could be formulated as follows:
\begin{equation}
	r=\argmax_{r\in\Gamma}\sum_{c\in C}\text{Rank}_X(q+r,c),
	\label{eq:qa_intuitive}
\end{equation}
where $\Gamma$ is the $\varepsilon$-neighbor of $q$. Likewise,
this attacking objective can also be transformed into the following constrained optimization problem:
%
\begin{align}
	L_{\text{QA-}}(q,C;X)&=\sum_{c\in C}\sum_{x\in X}\big[-d(q,c)+d(q,x)\big]_{+}, \label{eq:qa-_loss}\\
	r&=\argmin_{r\in\Gamma}L_{\text{QA-}}(q+r,C;X), \label{eq:qa_opt}
\end{align}
and it can be solved with the PGD algorithm. Similarly, the
\textbf{QA+} loss function $L_{\text{QA+}}$ for {\it raising} the rank of $c$
is as follows:
\begin{equation}
L_{\text{QA+}}(q,C;X)=\sum_{c\in C}\sum_{x\in X}\big[d(q,c)-d(q,x)\big]_{+}.
\label{eq:qa+}
\end{equation}

Unlike \textbf{CA}, \textbf{QA} perturbs the \emph{query} image,
and hence may drastically change its semantics, resulting in abnormal retrieval results.
For instance, after perturbing a ``lamp'' query image, some unrelated candidates (\eg, ``shelf'', ``toaster'', \etc)
may appear in the top return list.
Thus, an ideal query attack should preserve the query
semantics, \ie, the candidates in $X \setminus C$ \footnote{The complement of the set $C$.}
should retain their original ranks if possible.
Thus, we propose the Semantics-Preserving Query Attack
(\textbf{SP-QA}) by adding the \textbf{SP} term to mitigate the semantic
changes $q$, \eg,
\begin{small}
\begin{equation}
	L_{\text{SP-QA-}}(q,C;X)=L_{\text{QA-}}(q,C;X) + \xi L_{\text{QA+}}(q,C_{\text{SP}};X), \label{eq:spqa}
\end{equation}
\end{small}
where $C_{\text{SP}} = \left\{ c \in X \setminus C | \text{Rank}_{X
\setminus C}(q,c) \leqslant G \right\}$, \ie, $C_{\text{SP}}$ contains the
top-$G$ most-relevant candidates corresponding to $q$, and the
$L_{\text{QA+}}(q,C_{\text{SP}};X)$ term helps preserve the query semantics by
retaining some $C_{\text{SP}}$ candidates in the retrieved ranking list.
Constant $G$ is a predefined integer; and constant $\xi$ is a hyper-parameter
for balancing the attack effect and semantics preservation.
Unless mentioned, in the following text \textbf{QA} means \textbf{SP-QA} by default.



\subsection{Robustness \& Defense}


Adversarial defense for classification has been extensively explored,
and many of them follows the adversarial training
mechanism~\cite{advtrain,advscale,madry}. In particular, the adversarial
counterparts of the original training samples are used to replace or
augment the training samples.
Until now, Madry defense~\cite{madry} is regarded
as the most effective~\cite{bilateral,obfuscated} adversarial training method.
However, when directly adapting such classification defense to improve
ranking robustness, we empirically discovered a primary challenge of
diverging training loss, possibly due to the generated
adversarial examples being too ``strong''. Moreover, such defense
mechanism needs to defend against distinct attacks individually.
Therefore, a \emph{generic} defense against all the proposed attacks is preferred.

Note that the underlying principle of adversarial ranking attack is to shift the
embeddings of candidates/queries to a proper place, and a
successful attack depends on a large shift distance as well as a correct shift direction.
A large shift distance is an indispensable objective for all the \textbf{CA+}, \textbf{CA-}, \textbf{QA+} and \textbf{QA-}
attacks. Predictably, a reduction in shift distance could improve
model robustness against all attacks simultaneously.

To this end, we propose a
``maximum-shift-distance'' attack that pushes an embedding vector as far from its original
position as possible (resembles Feature Adversary~\cite{featureadversary}
for classification),
$r = \argmax_{r\in\Gamma} d(c+r, c) \label{eq:es}$.
Then we use adversarial examples obtained from this attack to replace
original training samples for adversarial training, hence reduce the
shift distance incurred by adversarial perturbations.

A ranking model can be normally trained with the
defensive version of the triplet loss:
\begin{small}
\begin{align}
	L_{\text{d-t}}(q,c_p,c_n) = L_{\text{triplet}} \Big( & q + \argmax_{r\in\Gamma}d(q+r,q),\nonumber \\
	& c_p + \argmax_{r\in\Gamma}d(c_p+r,c_p),\nonumber \\
	& c_n + \argmax_{r\in\Gamma}d(c_n+r,c_n) \Big).
	\label{eq:defense}
\end{align}
\end{small}
Unlike the direct adaptation of Madry defense, the training loss does
converge in our experiments.



\section{Experiments}

\begin{table}[!t]
\centering
\resizebox{1.0\columnwidth}{!}{%
\setlength{\tabcolsep}{0.2em}

\begin{tabular}{c|cccc|cccc|cccc|cccc}
\toprule
\multirow{2}{*}{$\varepsilon$} & \multicolumn{4}{c|}{CA+} & \multicolumn{4}{c|}{CA-} & \multicolumn{4}{c|}{QA+} & \multicolumn{4}{c}{QA-}\tabularnewline
\cline{2-17} \cline{3-17} \cline{4-17} \cline{5-17} \cline{6-17} \cline{7-17} \cline{8-17} \cline{9-17} \cline{10-17} \cline{11-17} \cline{12-17} \cline{13-17} \cline{14-17} \cline{15-17} \cline{16-17} \cline{17-17}
 & \multicolumn{1}{c|}{$w=1$} & \multicolumn{1}{c|}{$2$} & \multicolumn{1}{c|}{$5$} & $10$ & \multicolumn{1}{c|}{$w=1$} & \multicolumn{1}{c|}{$2$} & \multicolumn{1}{c|}{$5$} & $10$ & \multicolumn{1}{c|}{$m=1$} & \multicolumn{1}{c|}{$2$} & \multicolumn{1}{c|}{$5$} & $10$ & \multicolumn{1}{c|}{$m=1$} & \multicolumn{1}{c|}{$2$} & \multicolumn{1}{c|}{$5$} & $10$\tabularnewline
\midrule
\rowcolor{black!10}\multicolumn{17}{c}{(CT) Cosine Distance, Triplet Loss (R@1=99.1\%)}\tabularnewline
0 & 50 & 50 & 50 & 50 & 2.1 & 2.1 & 2.1 & 2.1 & 50 & 50 & 50 & 50 & 0.5 & 0.5 & 0.5 & 0.5\tabularnewline
\hline
0.01 & 44.6 & 45.4 & 47.4 & 47.9 & 3.4 & 3.2 & 3.1 & 3.1 & 45.2 & 46.3 & 47.7 & 48.5 & 0.9 & 0.7 & 0.6 & 0.6\tabularnewline
0.03 & 33.4 & 37.3 & 41.9 & 43.9 & 6.3 & 5.9 & 5.7 & 5.6 & 35.6 & 39.2 & 43.4 & 45.8 & 1.9 & 1.4 & 1.1 & 1.1\tabularnewline
0.1 & 12.7 & 17.4 & 24.4 & 30.0 & 15.4 & 14.9 & 14.8 & 14.7 & 14.4 & 21.0 & 30.6 & 37.2 & 5.6 & 4.4 & 3.7 & 3.5\tabularnewline
0.3 & \textbf{2.1} & \textbf{9.1} & \textbf{13.0} & \textbf{17.9} & \textbf{93.9} & \textbf{93.2} & \textbf{93.0} & \textbf{92.9} & \textbf{6.3} & \textbf{11.2} & \textbf{22.5} & \textbf{32.1} & \textbf{8.6} & \textbf{6.6} & \textbf{5.3} & \textbf{4.8}\tabularnewline
\bottomrule
\end{tabular}
}
\caption{Adversarial ranking attack on vanilla model with MNIST.
The ``+'' attacks (\ie CA+ and QA+) raise the rank of chosen
candidates towards $0$ ($\%$); while the ``-'' attacks (\ie CA- and QA-) lower
the ranks of chosen candidates towards $100$ ($\%$).
Applying $\varepsilon=0.01,0.03,0.1,0.3$ QA+ attacks on the model,
the SP term keeps the ranks of $C_\text{SP}$
no larger than $3.6\%,5.7\%,7.7\%,7.7\%$, respectively, regardless of $m$.
With the QA- counterpart, the ranks of $C_\text{SP}$ are
kept no larger than $1.6\%,1.6\%,1.5\%,1.5\%$, respectively, regardless of $m$.
For all the numbers in the table, ``\%'' is omitted.
}
\label{tab:mnist-attack}
\end{table}



To validate the proposed attacks and defense, we use three commonly
used ranking datasets
including MNIST~\cite{mnist}, Fashion-MNIST~\cite{fashion}, and
Stanford Online Product
(SOP)~\cite{sop}.  We respectively train models on
these datasets with PyTorch~\cite{pytorch}, and conduct attacks 
\footnote{Specifically, we use PGD without random starts~\cite{madry}.}
on their corresponding test datasets (used as $X$).

\textbf{Evaluation Metric.}
Adversarial ranking attack aims to change the ranks of candidates.
For each candidate $c$, its \emph{normalized} rank is calculated as
$R(q,c) = \frac{\text{Rank}_X(q,c)}{|X|}\times 100\%$
where $c\in X$, and $|X|$ is the length of full ranking list.
Thus, $R(q,c)\in [0,1]$, and a top ranked $c$ will have a small $R(q,c)$.
The attack effectiveness can be measured by the magnitude of change in
$R(q,c)$.

\textbf{Performance of Attack.}
To measure the performance of a single CA attack, we average the rank of
candidate $c$ across every query $q \in Q$, \ie, $R_\text{CA}(c)=\sum_{q\in Q}{R}(q,c)/w$.
Similarly, the performance of a single QA attack can be measured by the average
rank across every candidate $c \in C$, \ie, $R_\text{QA}(q)=\sum_{c\in C}{R}(q,c)/m$.
For the overall performance of an attack,
we conduct $T$ times of independent attacks and report the mean of $R_\text{CA}(c)$
or $R_\text{QA}(q)$, accordingly.


\textbf{CA+ \& QA+.}
For CA+, the query set $Q$ is randomly sampled from $X$.
Likewise, for QA+, the candidate set $C$ is from $X$.
Without attack, both the $R_\text{CA}(c)$ and $R_\text{QA}(q)$ will approximate to $50\%$,
and the attacks should significantly \emph{decrease} the value.

\textbf{CA- \& QA-.}
In practice, the $Q$ for CA- and the $C$ for QA- cannot be randomly sampled,
because the two attacks are often to lower some top ranked candidates.
Thus, the two sets should be selected from the top ranked samples (top-$1\%$ in
our experiments) in $X$.
Formally, given the candidate $c$ for CA-,
we randomly sample the $w$ queries from
$\left\{ q \in X | R(c,q) \leqslant 1\% \right\}$ as $Q$.
Given the query $q$ for QA-,
$m$ candidates are randomly sampled from
$\left\{ c \in X | R(q,c) \leqslant 1\% \right\}$ as $C$.
Without attack, both the $R_\text{CA}(c)$ and $R_\text{QA}(q)$ will be close to $0\%$,
and the attacks should significantly \emph{increase} the value.


\textbf{Hyper-Parameters.}
We conduct CA with $w \in \{1,2,5,10\}$ queries, and QA with 
$m \in \{1,2,5,10\}$ candidates, respectively. In QA, we let $G=5$.
The SP balancing parameter $\xi$ is set to $1$ for QA+ , and $10^2$ for \mbox{QA-.}
In addition, We investigate attacks of different strength $\varepsilon$,
\ie~$0.01,0.03,0.1,0.3$ on MNIST and Fashion-MNIST following \cite{madry},
and $0.01,0.03,0.06$ on SOP following \cite{advscale}.
The PGD step size is empirically set to
$\alpha=\min(\max(\frac{\varepsilon}{10},\frac{1}{255}),0.01)$,
and the number of PGD iterations to $\eta=\min(\max(10, \frac{2\varepsilon}{\alpha}),30)$.
We perform $T=|X|$ times of attack to obtain the reported performance.


\textbf{Adversarial Defense.} Ranking models are trained
using Eq.~\eqref{eq:defense} with the strongest adversary
following the procedure of Madry defense~\cite{madry}.

\subsection{MNIST Dataset}

\begin{table}[!t]
\centering
\resizebox{1.0\columnwidth}{!}{%
\setlength{\tabcolsep}{0.2em}
\begin{tabular}{c|cccc|cccc|cccc|cccc}
\toprule
\multirow{2}{*}{$\varepsilon$} & \multicolumn{4}{c|}{CA+} & \multicolumn{4}{c|}{CA-} & \multicolumn{4}{c|}{QA+} & \multicolumn{4}{c}{QA-}\tabularnewline
\cline{2-17} \cline{3-17} \cline{4-17} \cline{5-17} \cline{6-17} \cline{7-17} \cline{8-17} \cline{9-17} \cline{10-17} \cline{11-17} \cline{12-17} \cline{13-17} \cline{14-17} \cline{15-17} \cline{16-17} \cline{17-17}
 & \multicolumn{1}{c|}{$w=1$} & \multicolumn{1}{c|}{$2$} & \multicolumn{1}{c|}{$5$} & $10$ & \multicolumn{1}{c|}{$w=1$} & \multicolumn{1}{c|}{$2$} & \multicolumn{1}{c|}{$5$} & $10$ & \multicolumn{1}{c|}{$m=1$} & \multicolumn{1}{c|}{$2$} & \multicolumn{1}{c|}{$5$} & $10$ & \multicolumn{1}{c|}{$m=1$} & \multicolumn{1}{c|}{$2$} & \multicolumn{1}{c|}{$5$} & $10$\tabularnewline
\midrule
\rowcolor{blue!10}\multicolumn{17}{c}{(CTD) Cosine Distance, Triplet Loss, Defensive (R@1=98.3\%)}\tabularnewline
0 & 50 & 50 & 50 & 50 & 2.0 & 2.0 & 2.0 & 2.0 & 50 & 50 & 50 & 50 & 0.5 & 0.5 & 0.5 & 0.5\tabularnewline
\hline
0.01 & 48.9 & 49.3 & 49.4 & 49.5 & 2.2 & 2.2 & 2.2 & 2.1 & 49.9 & 49.5 & 49.5 & 49.7 & 0.5 & 0.5 & 0.5 & 0.5\tabularnewline
0.03 & 47.4 & 48.4 & 48.6 & 48.9 & 2.5 & 2.5 & 2.4 & 2.4 & 48.0 & 48.5 & 49.2 & 49.5 & 0.6 & 0.6 & 0.5 & 0.5\tabularnewline
0.1 & 42.4 & 44.2 & 45.9 & 46.7 & 3.8 & 3.6 & 3.5 & 3.4 & 43.2 & 45.0 & 47.4 & 48.2 & 1.0 & 0.8 & 0.7 & 0.7\tabularnewline
0.3 & \textbf{30.7} & \textbf{34.5} & \textbf{38.7} & \textbf{40.7} & \textbf{7.0} & \textbf{6.7} & \textbf{6.5} & \textbf{6.5} & \textbf{33.2} & \textbf{37.2} & \textbf{42.3} & \textbf{45.1} & \textbf{2.4} & \textbf{1.9} & \textbf{1.6} & \textbf{1.5}\tabularnewline
\bottomrule
\end{tabular}
}
\caption{Adversarial ranking defense with MNIST.
Applying $\varepsilon=0.01,0.03,0.1,0.3$ QA+ attacks on model,
the ranks of candidates
in $C_\text{SP}$ are kept no larger than $0.5\%,0.5\%,0.5\%,0.5\%$, respectively, regardless of $m$.
With the QA- counterpart, the ranks of $C_\text{SP}$ are
kept less than $0.4\%,0.4\%,0.4\%,0.4\%$, respectively, regardless of $m$.
}
\label{tab:mnist-defense}
\end{table}


Following conventional settings with the MNIST~\cite{mnist} dataset,
we train a CNN ranking model comprising $2$ convolutional layers and $1$ fully-connected layer.
This CNN architecture (denoted as C2F1) is identical to the one used in~\cite{madry} except for
the removal of the last fully-connected layer.
Specifically, the ranking model is trained with cosine distance and triplet loss.
The retrieval performance of the
model is Recall@1=$99.1\%$ (R@$1$), as shown in Tab.~\ref{tab:mnist-attack} in grey highlight.


Attacking results against this vanilla 
model (\ie, the ranking model which is not enhanced with our defense method)
are presented in Tab.~\ref{tab:mnist-attack}.
For example, a strong \textbf{CA+} attack (\ie, $\varepsilon=0.3$) for $w=1$
can raise the rank $R_\text{CA}(c)$ from $50\%$ to $2.1\%$.
Likewise, the rank of $C$ can be raised to $9.1\%$, $13.0\%$, $17.9\%$
for $w=2,5,10$ chosen queries, respectively.


On the other hand, a strong \textbf{CA-} attack for $w=1$
can lower the rank $R_\text{CA}(c)$ from $2.1\%$ to $93.9\%$.
The results of strong \textbf{CA-} attacks for $w=2,5,10$
are similar to the $w=1$ case.

The results of \textbf{QA+} and \textbf{QA-} are also shown in Tab.~\ref{tab:mnist-attack}.
the rank changes with \textbf{QA} attacks
are less dramatic (but still significant) than \textbf{CA}.
This is due to the additional difficulty introduced by \textbf{SP} term in Eq.~\eqref{eq:spqa},
and the \textbf{QA} attack effectiveness is inversely correlated with $\xi$.
For instance, a strong \textbf{QA-} for $m=1$ can only lower the rank
$R_\text{QA}(q)$ from $0.5\%$ to $8.6\%$, but the attacking effect can be
further boosted by decreasing $\xi$.
More experimental results are presented in following discussion.
%
%
In brief, our proposed attacks against the vanilla ranking model is effective.

Next, we evaluate the performance of our defense method. Our defense should be
able to enhance the robustness of a ranking model, which can be measured by the
difference between the attack effectiveness with our defense and the attack effectiveness without our defense.
%
%
As a common phenomenon of adversarial training,
our defense mechanism leads to a slight retrieval performance degradation
for unperturbed input (highlighted in blue in Tab.~\ref{tab:mnist-defense}),
but the attacking effectiveness is clearly mitigated by our defense.
For instance, the same strong \textbf{CA+} attack for $w=1$ on the defensive model (\ie, the ranking model which is enhanced by our defense method) 
can only raise the rank $R_\text{CA}(c)$
from $50\%$ to $30.7\%$, compared to its vanilla counterpart raising to $2.1\%$.
Further analysis suggests that the weights in the first convolution layer of the defensive model are
closer to $0$ and have smaller variance than those of the vanilla model,
which may help resist adversarial perturbation from changing the layer
outputs into the local linear area of ReLU~\cite{fgsm}.

\begin{figure}[t!]
\includegraphics[width=1.0\columnwidth]{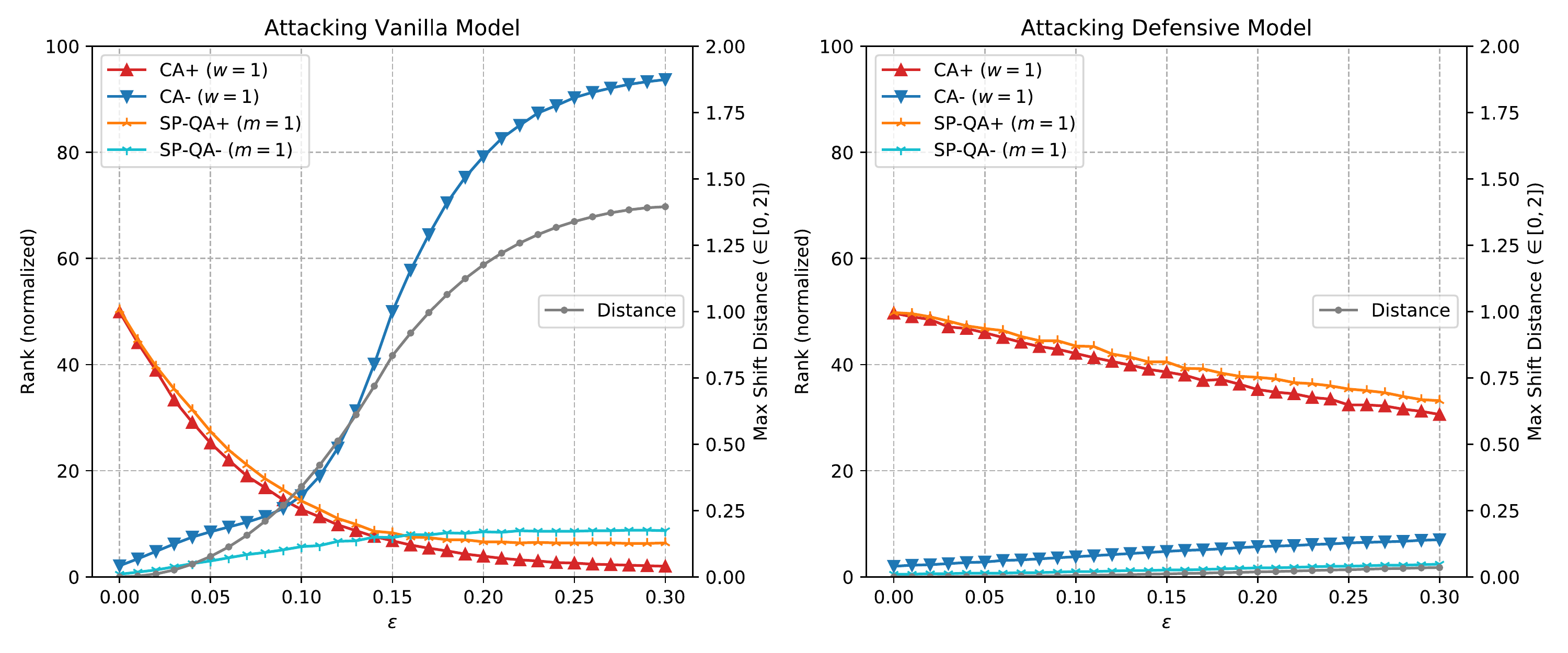}
\caption{Comparison of Attacks on vanilla and defensive models.
	Apart from the ranks of chosen candidates,
	We also measure the maximum shift distance of embedding vectors that
	adversarial perturbation could incur.
	}
\label{fig:mnist_curve}
\end{figure}

To visualize the effect of our attacks and defense, we track
the attacking effect with $\varepsilon$ varying from $0.0$ to $0.3$ on the
vanilla and defensive models, as shown in Fig.~\ref{fig:mnist_curve}.
It is noted that our defense could significantly suppress
the maximum embedding shift distance incurred by adversarial perturbation to nearly $0$,
but the defensive model is still not completely immune to attacks.
We speculate the
defensive model still has ``blind spots''~\cite{l-bfgs} in some local areas that could be
exploited by the attacks.

In summary, these results and further experiments 
suggest that:
(1) deep ranking
models are vulnerable to adversarial ranking attacks, no matter what loss
function or distance metric is selected;
(2) vanilla models trained with contrastive loss are more robust than those trained with triplet loss.
This is possibly
due to contrastive loss explicitly reducing the intra-class embedding variation.
Additionally, our defense method could consistently improve the robustness of all these models;
(3) different distance metrics have almost negligible contribution on robustness.
Specifically, Euclidean distance-based models are slightly more susceptible to weak
(\eg, $\varepsilon=0.03$) attacks;
(4) Euclidean distance-based models are harder to defend than
cosine distance-based ones.
%
Beyond these experiments, we also find that the margin hyper-parameter $\beta$
of triplet loss and the dimensionality of the embedding space have marginal influences on
model robustness.


\subsection{Fashion-MNIST Dataset}

\begin{table}[!t]
\centering
\resizebox{1.0\columnwidth}{!}{
\setlength{\tabcolsep}{0.2em}

\begin{tabular}{c|cccc|cccc|cccc|cccc}
\toprule
\multirow{2}{*}{$\varepsilon$} & \multicolumn{4}{c|}{CA+} & \multicolumn{4}{c|}{CA-} & \multicolumn{4}{c|}{QA+} & \multicolumn{4}{c}{QA-}\tabularnewline
\cline{2-17} \cline{3-17} \cline{4-17} \cline{5-17} \cline{6-17} \cline{7-17} \cline{8-17} \cline{9-17} \cline{10-17} \cline{11-17} \cline{12-17} \cline{13-17} \cline{14-17} \cline{15-17} \cline{16-17} \cline{17-17}
 & \multicolumn{1}{c|}{$w=1$} & \multicolumn{1}{c|}{$2$} & \multicolumn{1}{c|}{$5$} & $10$ & \multicolumn{1}{c|}{$w=1$} & \multicolumn{1}{c|}{$2$} & \multicolumn{1}{c|}{$5$} & $10$ & \multicolumn{1}{c|}{$m=1$} & \multicolumn{1}{c|}{$2$} & \multicolumn{1}{c|}{$5$} & $10$ & \multicolumn{1}{c|}{$m=1$} & \multicolumn{1}{c|}{$2$} & \multicolumn{1}{c|}{$5$} & $10$\tabularnewline
\midrule
\rowcolor{black!10}\multicolumn{17}{c}{(CT) Cosine Distance, Triplet Loss (R@1=88.8\%)}\tabularnewline
0 & 50 & 50 & 50 & 50 & 1.9 & 1.9 & 1.9 & 1.9 & 50 & 50 & 50 & 50 & 0.5 & 0.5 & 0.5 & 0.5\tabularnewline
\hline
0.01 & 36.6 & 39.9 & 43.2 & 44.8 & 5.6 & 5.1 & 4.9 & 4.8 & 39.4 & 42.0 & 45.3 & 47.1 & 2.1 & 1.6 & 1.2 & 1.1\tabularnewline
0.03 & 19.7 & 25.4 & 31.7 & 35.6 & 15.5 & 14.8 & 14.4 & 14.3 & 21.7 & 28.2 & 35.7 & 40.6 & 5.6 & 4.1 & 3.3 & 2.9\tabularnewline
0.1 & 3.7 & 10.5 & 17.3 & 22.7 & 87.2 & 86.7 & 86.3 & 86.3 & 7.1 & 12.4 & 23.6 & 32.5 & 10.9 & 8.3 & 6.7 & 6.0\tabularnewline
0.3 & \textbf{1.3} & \textbf{9.4} & \textbf{16.0} & \textbf{21.5} & \textbf{100.0} & \textbf{100.0} & \textbf{100.0} & \textbf{100.0} & \textbf{6.3} & \textbf{10.8} & \textbf{21.8} & \textbf{31.7} & \textbf{12.6} & \textbf{9.4} & \textbf{7.5} & \textbf{6.6}\tabularnewline
\hline
\rowcolor{blue!10}\multicolumn{17}{c}{(CTD) Cosine Distance, Triplet Loss, Defensive (R@1=79.6\%)}\tabularnewline
0 & 50 & 50 & 50 & 50 & 1.2 & 1.2 & 1.2 & 1.2 & 50 & 50 & 50 & 50 & 0.5 & 0.5 & 0.5 & 0.5\tabularnewline
\hline
0.01 & 48.9 & 48.9 & 49.3 & 49.3 & 1.4 & 1.4 & 1.4 & 1.4 & 49.4 & 49.9 & 49.9 & 50.0 & 0.5 & 0.5 & 0.5 & 0.5\tabularnewline
0.03 & 47.1 & 47.9 & 48.3 & 48.3 & 2.0 & 1.9 & 1.8 & 1.8 & 48.3 & 49.1 & 49.5 & 49.8 & 0.7 & 0.6 & 0.6 & 0.6\tabularnewline
0.1 & 42.4 & 43.5 & 44.5 & 44.8 & 4.6 & 4.2 & 4.0 & 3.9 & 45.4 & 47.2 & 48.7 & 49.2 & 1.4 & 1.2 & 1.1 & 1.1\tabularnewline
0.3 & \textbf{32.5} & \textbf{35.4} & \textbf{37.5} & \textbf{38.2} & \textbf{11.2} & \textbf{10.5} & \textbf{10.1} & \textbf{10.0} & \textbf{39.3} & \textbf{42.6} & \textbf{46.5} & \textbf{47.8} & \textbf{3.9} & \textbf{3.3} & \textbf{3.0} & \textbf{2.9}\tabularnewline
\bottomrule
\end{tabular}

}
\caption{Adversarial ranking attack and defense on Fashion-MNIST.
	The lowest ranks of $C_\text{SP}$
	are $3.0\%,5.2\%,7.8\%,8.3\%$ in QA+, and $1.9\%,1.9\%,1.9\%,1.8\%$ for QA+,
	respectively.}
\label{tab:fashion}
\end{table}


Fashion-MNIST~\cite{fashion} is an MNIST-like but more difficult dataset,
comprising $60,000$ training examples and $10,000$ test samples. The samples are
$28\times 28$ greyscale images covering $10$ different fashion
product classes, including ``T-shirt'' and ``dress'', \etc.  We train the vanilla
and defensive models based on the cosine distance and triplet loss and 
conduct attack experiments.


The attack and defense results are available in Tab.~\ref{tab:fashion}.
From the table, we note that our attacks could achieve better effect compared to
experiments on MNIST.
For example, in a strong \textbf{CA+} for $w=1$, the rank $R_\text{CA}(c)$ can be
raised to $1.3\%$.
On the other hand, despite the moderate improvement in
robustness, the defensive model performs worse in unperturbed
sample retrieval, as expected.
The performance degradation is more pronounced on this dataset compared to MNIST.
We speculate the differences are related to the increased dataset difficulty.

\subsection{Stanford Online Products Dataset}


Stanford Online Products (SOP) dataset~\cite{sop} contains $120$k images of
$23$k classes of real online products from eBay for metric learning.
We use the same dataset split as used in the original work~\cite{sop}.
We also train the same vanilla ranking model using the same
triplet ranking loss function with Euclidean distance, except that
the GoogLeNet~\cite{googlenet} is replaced with ResNet-18~\cite{resnet}.
The ResNet-18 achieves better retrieval performance.


Attack and defense results on SOP are present in Tab.~\ref{tab:sop}.
It is noted that our attacks are quite effective on this difficult large-scale dataset,
as merely $1\%$ perturbation ($\varepsilon=0.01$) to any candidate image could make
it ranked ahead or behind of nearly all the rest candidates (as shown by the CA+
and CA- results with $w=1$).
The QA on this dataset is significantly effective as well.
On the other hand, our defense method leads to
decreased retrieval performance, \emph{i.e.} R@1 from
$63.1\%$ to $46.4\%$, which is expected on such a difficult dataset.
Meanwhile, our defense could moderately improve the model
robustness against relatively weaker adversarial examples (\emph{e.g.}
$\varepsilon=0.01$), but improving model robustness on this dataset
is more difficult, compared to experiments on other datasets.

By comparing the results among all the three datasets, we find ranking
models trained on harder datasets more susceptible to adversarial attack,
and more difficult to defend.
Therefore, we speculate that models used in realistic applications could be
easier to attack, because they are usually trained on larger-scale and
more difficult datasets.


\begin{table}[!t]
\centering
\resizebox{1.00\columnwidth}{!}{
\setlength{\tabcolsep}{0.2em}

\begin{tabular}{c|cccc|cccc|cccc|cccc}
\toprule
\multirow{2}{*}{$\varepsilon$} & \multicolumn{4}{c|}{CA+} & \multicolumn{4}{c|}{CA-} & \multicolumn{4}{c|}{QA+} & \multicolumn{4}{c}{QA-}\tabularnewline
\cline{2-17} \cline{3-17} \cline{4-17} \cline{5-17} \cline{6-17} \cline{7-17} \cline{8-17} \cline{9-17} \cline{10-17} \cline{11-17} \cline{12-17} \cline{13-17} \cline{14-17} \cline{15-17} \cline{16-17} \cline{17-17}
 & \multicolumn{1}{c|}{$w=1$} & \multicolumn{1}{c|}{$2$} & \multicolumn{1}{c|}{$5$} & $10$ & \multicolumn{1}{c|}{$w=1$} & \multicolumn{1}{c|}{$2$} & \multicolumn{1}{c|}{$5$} & $10$ & \multicolumn{1}{c|}{$m=1$} & \multicolumn{1}{c|}{$2$} & \multicolumn{1}{c|}{$5$} & $10$ & \multicolumn{1}{c|}{$m=1$} & \multicolumn{1}{c|}{$2$} & \multicolumn{1}{c|}{$5$} & $10$\tabularnewline
\midrule
\rowcolor{black!10}\multicolumn{17}{c}{(ET) Euclidean Distance, Triplet Loss (R@1=63.1\%)}\tabularnewline
0 & 50 & 50 & 50 & 50 & 1.9 & 1.9 & 1.9 & 1.9 & 50 & 50 & 50 & 50 & 0.5 & 0.5 & 0.5 & 0.5\tabularnewline
\hline
0.01 & 0.0 & 0.8 & 2.0 & 2.6 & 99.7 & 99.6 & 99.4 & 99.3 & 4.8 & 7.0 & 16.3 & 25.8 & 54.9 & 40.2 & 27.1 & 21.9 \tabularnewline
0.03 & 0.0 & 0.3 & 1.0 & 1.5 & 100.0 & 100.0 & 100.0 & 100.0 & 1.6 & 3.3 & 10.0 & 19.2 & 68.1 & 52.4 & 36.6 & 30.1 \tabularnewline
0.06 & \textbf{0.0} & \textbf{0.2} & \textbf{1.0} & \textbf{1.5} & \textbf{100.0} & \textbf{100.0} & \textbf{100.0} & \textbf{100.0} & \textbf{1.1} & \textbf{2.7} & \textbf{8.8} & \textbf{17.6} & \textbf{73.8} & \textbf{57.9} & \textbf{40.3} & \textbf{32.4} \tabularnewline

\hline
\rowcolor{blue!10}\multicolumn{17}{c}{(ETD) Euclidean Distance, Triplet Loss, Defensive (R@1=46.4\%)}\tabularnewline
0 & 50 & 50 & 50 & 50 & 2.0 & 2.0 & 2.0 & 2.0 & 50 & 50 & 50 & 50 & 0.5 & 0.5 & 0.5 & 0.5\tabularnewline
\hline
0.01 & 7.5 & 12.2 & 16.5 & 18.0 & 66.4 & 62.6 & 59.3 & 57.8 & 16.1 & 24.8 & 36.1 & 41.4 & 26.7 & 18.1 & 12.2 & 10.2 \tabularnewline
0.03 & 0.7 & 4.5 & 8.7 & 10.4 & 91.7 & 90.2 & 89.1 & 88.4 & 7.9 & 14.5 & 27.2 & 35.6 & 43.4 & 31.7 & 21.9 & 18.1 \tabularnewline
0.06 & \textbf{0.1} & \textbf{3.8} & \textbf{7.9} & \textbf{9.7} & \textbf{97.3} & \textbf{96.8} & \textbf{96.4} & \textbf{96.2} & \textbf{6.9} & \textbf{12.5} & \textbf{24.3} & \textbf{33.4} & \textbf{51.4} & \textbf{39.0} & \textbf{28.0} & \textbf{23.5} \tabularnewline
\bottomrule
\end{tabular}
}

	\caption{Adversarial ranking attack and defense on SOP. With
	different $\varepsilon$, the worst ranks of $C_\text{SP}$ in
	QA+ are $0.2\%,0.7\%,2.0\%,3.3\%$, and those for QA-
	are $0.4\%,0.7\%,0.8\%,1.0\%$, respectively.
	}
\label{tab:sop}
\end{table}

\section{Discussions}

In this section, we study the transferability of our adversarial ranking
examples, and universal adversarial perturbation for ranking. Both of them
illustrate the possibility of practical black-box attack.
Additionally, we also perform parameter search on the balancing parameter $\xi$
for \textbf{QA}.


\subsection{Adversarial Example Transferability}

As demonstrated in previous experiments, deep ranking models can be
compromised by our white-box attacks.
In realistic scenarios, the white-box
attacks are not practical enough because the model to be attacked is often unknown (\ie, the architecture
and parameters are unknown).

On the other hand, adversarial examples for classification have been found
transferable~\cite{black-frame,mltransfer} (\ie model-agnostic) between different
models with different network architectures.
And the transferability has become the foundation of a class of existing black-box
attacks.
Specifically, for such a typical attack, adversarial examples
are generated from a replacement model~\cite{black-frame} using a white-box
attack, and are directly used to attack the black-box model.

Adversarial ranking attack could be more practical if the adversarial ranking
examples have the similar transferability.
Besides the C2F1 model,
we train two vanilla models on the MNIST dataset:
(1) LeNet~\cite{mnist}, which has lower model capacity compared to C2F1;
(2) ResNet-18~\cite{resnet} (denoted as Res18), which has a better network architecture and higher
model capacity.

\begin{table}[!t]
\centering
\resizebox{1.0\columnwidth}{!}{
\setlength{\tabcolsep}{0.2em}%
\begin{tabular}{c|ccc}
\toprule

\rowcolor{black!10}\multicolumn{4}{c}{CA+ Transfer (Black Box), $w=1$}\tabularnewline
\backslashbox{From}{To} & LeNet & C2F1 & Res18\tabularnewline
\hline
LeNet & \cellcolor{black!10}50$\rightarrow$16.6 &35.1  &34.3 \tabularnewline
C2F1 & 28.6 &\cellcolor{black!10}50$\rightarrow$2.1 &31.3\tabularnewline
Res18 & 24.4 &27.0 &\cellcolor{black!10}50$\rightarrow$2.2\tabularnewline

\midrule

\rowcolor{black!10}\multicolumn{4}{c}{CA- Transfer (Black Box), $w=1$}\tabularnewline
\backslashbox{From}{To} & LeNet & C2F1 & Res18\tabularnewline
\hline
LeNet & \cellcolor{black!10}2.5$\rightarrow$63.7 & 2.1$\rightarrow$10.0 & 2.1$\rightarrow$9.1\tabularnewline
C2F1 & 2.5$\rightarrow$9.1 & \cellcolor{black!10}2.1$\rightarrow$93.9 & 2.1$\rightarrow$9.3\tabularnewline
Res18 & 2.5$\rightarrow$9.9 & 2.1$\rightarrow$11.8 & \cellcolor{black!10}2.1$\rightarrow$66.7\tabularnewline

\bottomrule
\end{tabular}

\begin{tabular}{c|ccc}
\toprule

\rowcolor{black!10}\multicolumn{4}{c}{QA+ Transfer (Black Box), $m=1$}\tabularnewline
\backslashbox{From}{To} & LeNet & C2F1 & Res18\tabularnewline
\hline
LeNet & \cellcolor{black!10}50$\rightarrow$20.5 &43.0 &45.8\tabularnewline
C2F1 &  43.5  &\cellcolor{black!10}50$\rightarrow$6.3 &45.4 \tabularnewline
Res18 & 41.4  &40.4  &\cellcolor{black!10}50$\rightarrow$14.1 \tabularnewline

\midrule

\rowcolor{black!10}\multicolumn{4}{c}{QA- Transfer (Black Box), $m=1$}\tabularnewline
\backslashbox{From}{To} & LeNet & C2F1 & Res18\tabularnewline
\hline
LeNet & \cellcolor{black!10}0.5$\rightarrow$7.0 & 0.5$\rightarrow$1.6 & 0.5$\rightarrow$1.8 \tabularnewline
C2F1 &  0.5$\rightarrow$1.0 & \cellcolor{black!10}0.5$\rightarrow$8.6 & 0.5$\rightarrow$1.9 \tabularnewline
Res18 & 0.5$\rightarrow$0.8 & 0.5$\rightarrow$1.2 & \cellcolor{black!10}0.5$\rightarrow$6.9 \tabularnewline

\bottomrule
\end{tabular}
}

\caption{Transferability of adversarial ranking examples.
Adversarial examples are generated from one model and directly used on another.
We report the rank of the same $c$ with respect to the same $q$ across
different models to illustrate the transfer attack effectiveness.
Transferring adversarial examples to a model itself (the diagonal lines) is equivalent to white-box attack.
}
\label{tab:transfer}
\end{table}


The results are present in Tab.~\ref{tab:transfer}.
For example, in the \textbf{CA+} transfer attack, we
generate adversarial candidates from the C2F1 model and directly use them to attack
the Res18 model (row 2, column 3, top-left table), and the ranks of the adversarial candidates
with respect to the same query is still raised to
$31.3\%$.
We also find the \textbf{CA-} transfer attack is effective, where the
ranks of our adversarial candidates are
lowered, \emph{e.g.} from $2.1\%$ to $9.3\%$ (row 2, column 3, bottom-left table).
Similar results can be observed on the \textbf{QA}
transfer experiments, and they show weaker effect due to the SP term.

From the results, we find that: (1) CNN with better architecture
and higher model capacity (\ie, Res18) is less susceptible to
adversarial ranking attack.
This conclusion is consistent with Madry's~\cite{madry}, which
claims that higher model capacity could help improve model robustness; 
(2) adversarial examples generated from the Res18 have the most significant effectiveness in transfer attack;
(3) CNN of low model capacity (\ie,
LeNet), performs moderately in terms of both adversarial example transferability
and model robustness. We speculate its robustness stems
from a forced regularization effect due low model capacity.
Beyond these, adversarial ranking examples are
transferable disregarding the difference in loss function or distance metric.

Apart from transferability across different architectures, we also investigated
the transferability between the C2F1 models with different network parameters.
Results suggest similar transferability between these models.
Notably, when transferring adversarial examples to a defensive C2F1 model, the attacking effect
is significantly mitigated. The result further demonstrates the effectiveness of
our defense. 

\subsection{Universal Perturbation for Ranking}

Recently, universal (\ie image-agnostic) adversarial
perturbation~\cite{universal} for classification has been found possible, where a single
perturbation may lead to misclassification when added to any image.
Thus, we
also investigate the existence of universal adversarial perturbation for ranking.

To this end, we follow~\cite{universal} and formulate the image-agnostic CA+ (\emph{abbr.} \textbf{I-CA+}).
Given a set of candidates
$C=\{c_1,c_2,\ldots,c_m\}$ and a set of queries $Q=\{q_1,q_2,\ldots,q_w\}$,
\textbf{I-CA+} is to find a \emph{single} universal adversarial perturbation $r$, so that
the rank of every perturbed candidate $\tilde{c}=c+r ~ (c\in C)$ with respect to $Q$
can be raised.
The corresponding optimization problem of \textbf{I-CA+} is:
\begin{equation}
	r=\argmin_{r\in\Gamma} \sum_{c\in C} L_\text{CA+}(c+r,Q;X).
\end{equation}
When applied with such
universal perturbation, the rank of any candidate \emph{w.r.t} $Q$ is expected to be
raised.
The objective functions of \textbf{I-CA-}, \textbf{I-QA+} and \textbf{I-QA-} can be obtained in similar way.
Note, unlike \cite{universalret} which aims to find universal perturbation that can make image retrieval system return
irrelevant results, our universal perturbations have distinct purposes.

We conduct experiment on the MNIST dataset.
For \textbf{I-CA+} attack, we randomly sample $5\%$ of $X$ for generating the universal perturbation.
Following~\cite{universal}, another non-overlapping $5\%$ examples are randomly sampled from $X$ to test whether
the generated perturbation is generalizable on ``unseen'' (\ie, not used for generating the perturbation)
images.
Experiments for the other image-agnostic attacks are conducted similarly.
Note, we only report the \textbf{I-CA-} and \textbf{I-QA-} effectiveness
on the $1\%$ top ranked samples, similar to \textbf{CA-} and \textbf{QA-}.


\begin{table}[t!]
\centering
\resizebox{0.85\columnwidth}{!}{
\setlength{\tabcolsep}{0.4em}
\begin{tabular}{cc|cc}
	\toprule
	\rowcolor{black!10}CA+ & CA- & QA+ & QA-\tabularnewline
 50 $\rightarrow$ 2.1 & 2.1 $\rightarrow$ 93.9 & 50 $\rightarrow$ 0.2 & 0.5 $\rightarrow$ 94.1 \tabularnewline
\rowcolor{black!10}I-CA+ & I-CA- & I-QA+ & I-QA-\tabularnewline
 50 $\rightarrow$ 18.1 & 0.6 $\rightarrow$ 9.5 & 50 $\rightarrow$ 20.5 & 2.1 $\rightarrow$ 7.6\tabularnewline
 \midrule
\rowcolor{black!10}I-CA+ (unseen) & I-CA- (unseen) & I-QA+ (unseen) & I-QA- (unseen)\tabularnewline
 50 $\rightarrow$ 18.5& 0.7 $\rightarrow$ 9.4 & 50 $\rightarrow$ 21.0 & 2.2 $\rightarrow$ 7.4\tabularnewline
\bottomrule
\end{tabular}}

\caption{Universal Adversarial Perturbation for Ranking on MNIST. Each pair of results
	presents the original rank of chosen candidates and that
	after adding adversarial perturbation. Both
	$w$, $m$ are set to $1$. Parameter $\xi$ is set to $0$ to reduce attack difficulty.}

\label{tab:iap-mnist}
\end{table}

As shown in Tab.~\ref{tab:iap-mnist},
our \textbf{I-CA} can raise the ranks of $C$ to $18.1\%$, or lower them to $9.5\%$.
When added to ``unseen`` candidate images, our universal perturbation
could retain nearly the same effectiveness.
This may due to low intra-class variance of the MNIST dataset.
%

\subsection{Semantics Preserving for QA}

As discussed previously, the Query Attack (\textbf{QA}) may drastically change the
semantics of the query $q$. To alleviate this problem, the Semantics-Preserving (SP)
term is added to the naive \textbf{QA} to help preserve the query semantics.
Predictably, it is more difficult to perform \textbf{QA} with a large $\xi$,
as the ranks of $C_\text{SP}$ are almost
not allowed to be changed.

To investigate the actual influence of the balancing parameter $\xi$, we provide
parameter search on it with MNIST dataset. In particular,
We set $\xi$ to $0,10^0,10^2,10^4$, and compare their results.
Note that when $\xi=0$, the QA becomes \emph{naive} QA as the SP term is
eliminated. With a strong SP constant, \eg $\xi=10^4$, the semantics
of the chosen query is almost not allowed to be changed, hence result in extreme difficulty of attack.

As shown in Tab.~\ref{tab:xi-ablation-lite},
setting $\xi$ to $0$ could greatly boost the attacking effect, but
consequently the ranks of $C_\text{SP}$ will be drastically changed.
In contrast, when $\xi$ is set to the excessive value $10^4$ for a perfectly stealth QA,
the attack can still raise the rank of chosen candidate from $50\%$ to $37.9\%$ in QA+ with $m=1$,
or lower the rank of chosen candidate from $0.5\%$ to $1.9\%$ in QA- with $m=1$.
During these attacks, the ranks of $C_\text{SP}$ are kept within $0.1$ despite
of the extreme difficulty. It means the query semantics can be preserved.
In practice, we empirically set the parameter $\xi$ as $1$ for QA+, or as $10^2$ for QA-
for the balance between attack effectiveness and preserving query semantics.

\begin{table}[t!]
\centering
\resizebox{1.0\columnwidth}{!}{
\setlength{\tabcolsep}{0.2em}%
\begin{tabular}{c|cccc|cccc}
\toprule
\multirow{2}{*}{$\xi$} & \multicolumn{4}{c|}{QA+} & \multicolumn{4}{c}{QA-}\tabularnewline
\cline{2-9} \cline{3-9} \cline{4-9} \cline{5-9} \cline{6-9} \cline{7-9} \cline{8-9} \cline{9-9}
 & \multicolumn{1}{c|}{m=1} & \multicolumn{1}{c|}{2} & \multicolumn{1}{c|}{5} & 10 & \multicolumn{1}{c|}{m=1} & \multicolumn{1}{c|}{2} & \multicolumn{1}{c|}{5} & 10\tabularnewline
 \midrule

\rowcolor{black!10}\multicolumn{9}{c}{(CT) Cosine distance, Triplet loss}\tabularnewline
0 & 0.2,~33.6 & 6.3,~23.7 & 18.5,~26.5 & 29.6,~25.7 & 94.1,~89.4 & 93.2,~90.3 & 92.6,~90.9 & 92.3,~91.2\tabularnewline
$10^0$ & 6.3,~3.6 & 11.2,~5.7 & 22.5,~7.7 & 32.1,~7.7 & 55.5,~35.6 & 52.4,~37.6 & 50.2,~39.3 & 49.4,~40.0\tabularnewline
$10^2$ & 14.1,~0.6 & 20.8,~0.7 & 31.2,~0.7 & 38.1,~0.7 & 8.6,~1.6 & 6.6,~1.6 & 5.3,~1.5 & 4.8,~1.5\tabularnewline
$10^4$ & 37.9,~0.1 & 42.6,~0.1 & 46.3,~0.1 & 47.8,~0.1 & 1.9,~0.1 & 1.4,~0.1 & 1.2,~0.1 & 1.1,~0.1\tabularnewline
\bottomrule
\end{tabular}
	}

\caption{Parameter search on Semantics-Preserving balancing parameter $\xi$ with MNIST.
	We report two mean ranks in each cell: one for the chosen candidates $C$;
	another for $C_\text{SP}$ used for SP.
	}
\label{tab:xi-ablation-lite}
\end{table}

\section{Conclusion}

Deep ranking models are vulnerable to adversarial perturbations that could
intentionally change the ranking result. In this paper, we define
and implement \emph{adversarial ranking attack} that can compromise
deep ranking models. We also propose an \emph{adversarial ranking
defense} that can significantly suppress embedding shift distance and moderately
improve the ranking model robustness. Moreover, the transferability of our adversarial
examples and the existence of universal
adversarial perturbations for ranking attack illustrate the possibility of practical black-box attack
and potential risk of realistic ranking applications.
In the potential of future work, we may explore
better ranking loss functions, defenses, and black-box attacks.

\section*{Acknowledgements}

{\small
This work was supported partly by National Key R\&D Program of China Grant
2018AAA0101400, NSFC Grants 61629301, 61773312, 61976171, and 61672402, China
Postdoctoral Science Foundation Grant 2019M653642, Young Elite Scientists
Sponsorship Program by CAST Grant 2018QNRC001, and Natural Science Foundation
of Shaanxi Grant 2020JQ-069.
}

{\small
\bibliographystyle{ieee_fullname}
\bibliography{egbib}
}

\newpage
\appendix

\section{Adversarial Example Visualization}

Some adversarial ranking examples are presented in this section.
Every figure contains three rows of pictures.
\textbf{The first row}
shows \textcolor{violet}{$c$, $r$, $\tilde{c}=c+r$} for CA,
or \textcolor{violet}{$q$, $r$, $\tilde{q}=q+r$} for QA.
\textbf{The second row} shows the query and the original
retrieval results, as well as the chosen candidate \textcolor{violet}{$c$}
and its immediately adjacent candidates.
\textbf{The third row} shows the effects of the attack on the ranking
list, \ie, either the chosen candidate \textcolor{violet}{$c$} is replaced
with \textcolor{violet}{$\tilde{c}$} for CA, or the
query \textcolor{violet}{$q$} is replaced with \textcolor{violet}{$\tilde{q}$} for QA.
\textbf{The digits} above every picture is the value of $\text{Rank}_X(q,\cdot)-1$
\footnote{In mathematical context the rank counts from $1$, but in implementation
it counts from $0$ instead. Hence the offset.}.

Pictures with a `` \textcolor{yellow}{\contour{cyan}{$\bigstar$}}
'' mark on the top-left corner are
adversarial examples.
The ``\textcolor{red}{$\leftarrow$}'' indicates the
chosen candidate whose rank will be \textcolor{red}{raised}.
The ``\textcolor{blue}{$\rightarrow$}'' indicates the
chosen candidate whose rank will be \textcolor{blue}{lowered}.

Full visualization results including those on SOP (removed from the
Arxiv version due to file size limit)
can be found at \url{https://github.com/cdluminate/advrank-pub}. 

\subsection{MNIST Dataset}

\begin{figure}[h]
\includegraphics[width=1.0\columnwidth]{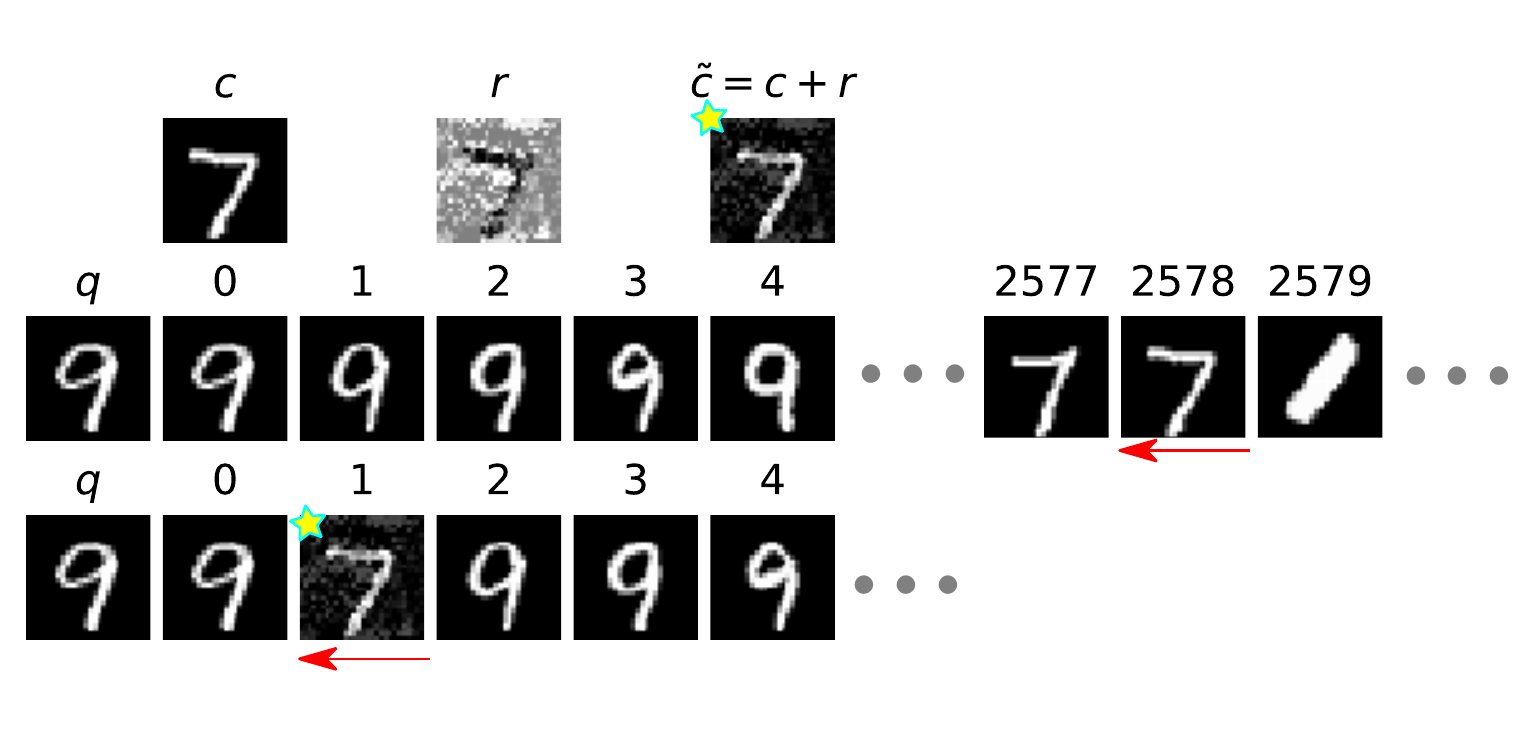}

\caption{CA+ on MNIST. Example 1. For query ``9'', the candidate ``7'' is ranked at
	the $2576$-th position in the original ranking list (row 2). After
	adversarial ranking attack, the rank of the perturbed candidate ``7'' is
	raised to the $1$-st position (row 3). The original candidate ``7'', the
	learned perturbation $r$, and the perturbed candidate ``7'' are illustrated
	in the row $1$.}

\label{fig:mcp1}
\end{figure}

\begin{figure}[h]
\includegraphics[width=1.0\columnwidth]{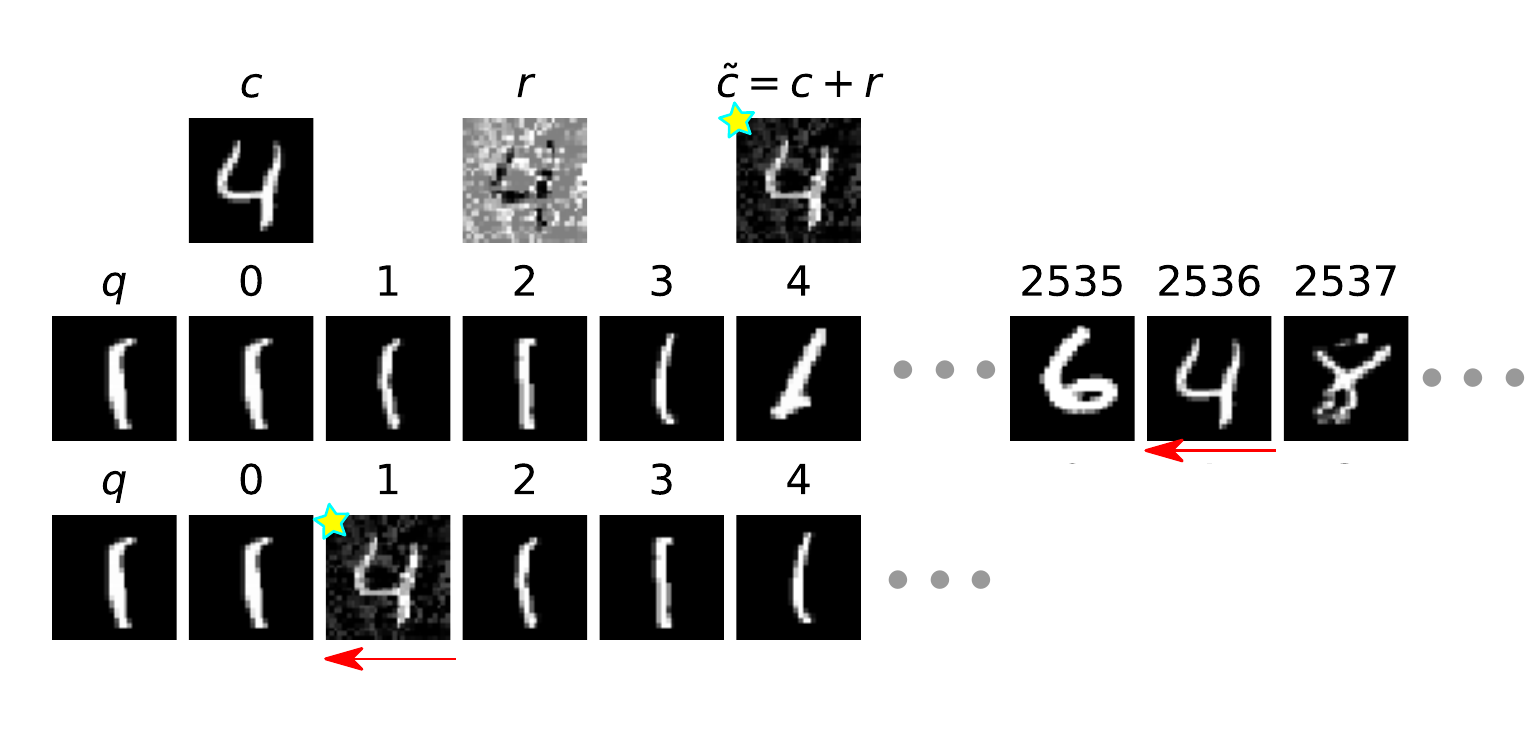}
\caption{CA+ on MNIST. Example 2.}
\label{fig:mcp2}
\end{figure}

\begin{figure}[h]
\includegraphics[width=1.0\columnwidth]{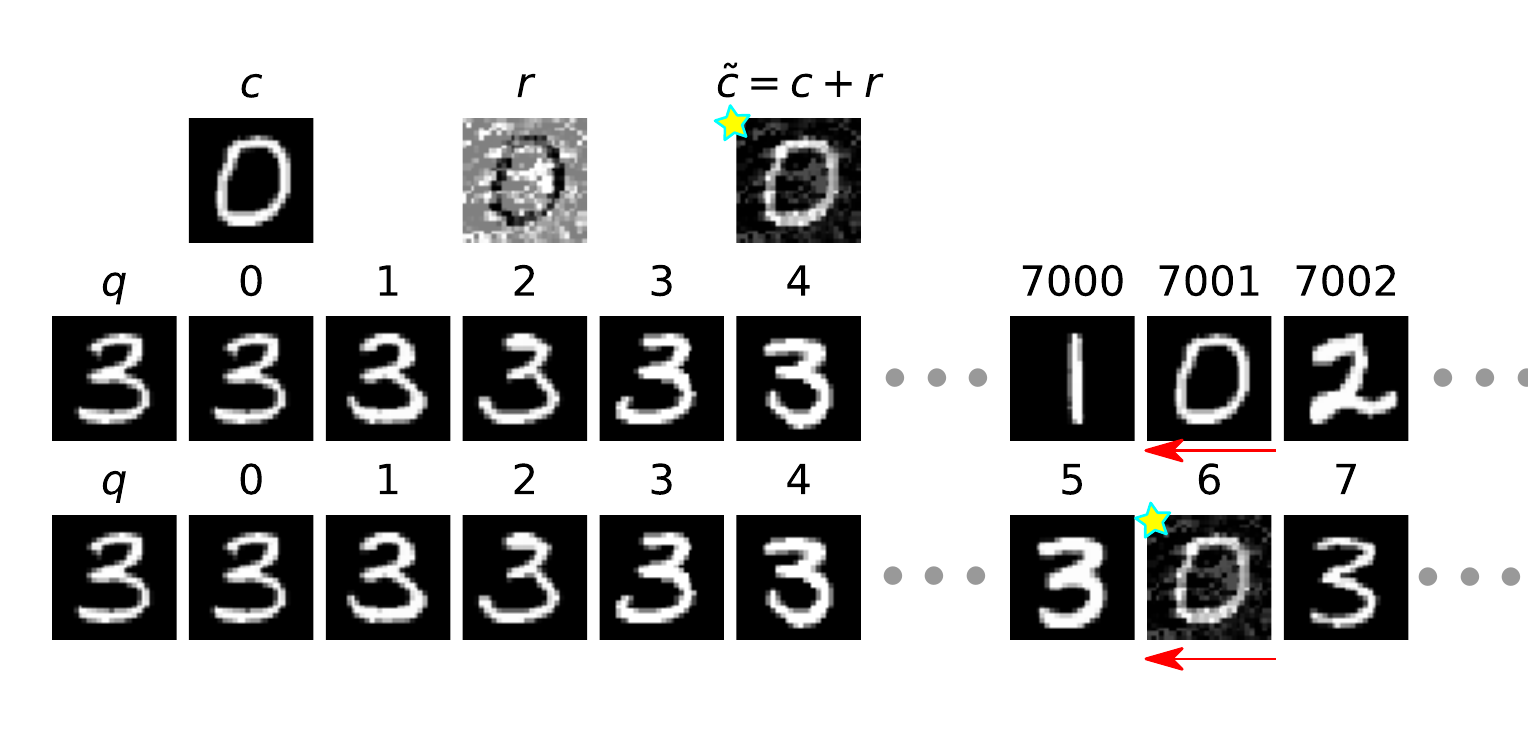}
\caption{CA+ on MNIST. Example 3.}
\label{fig:mcp3}
\end{figure}

\begin{figure}[h]
\includegraphics[width=1.0\columnwidth]{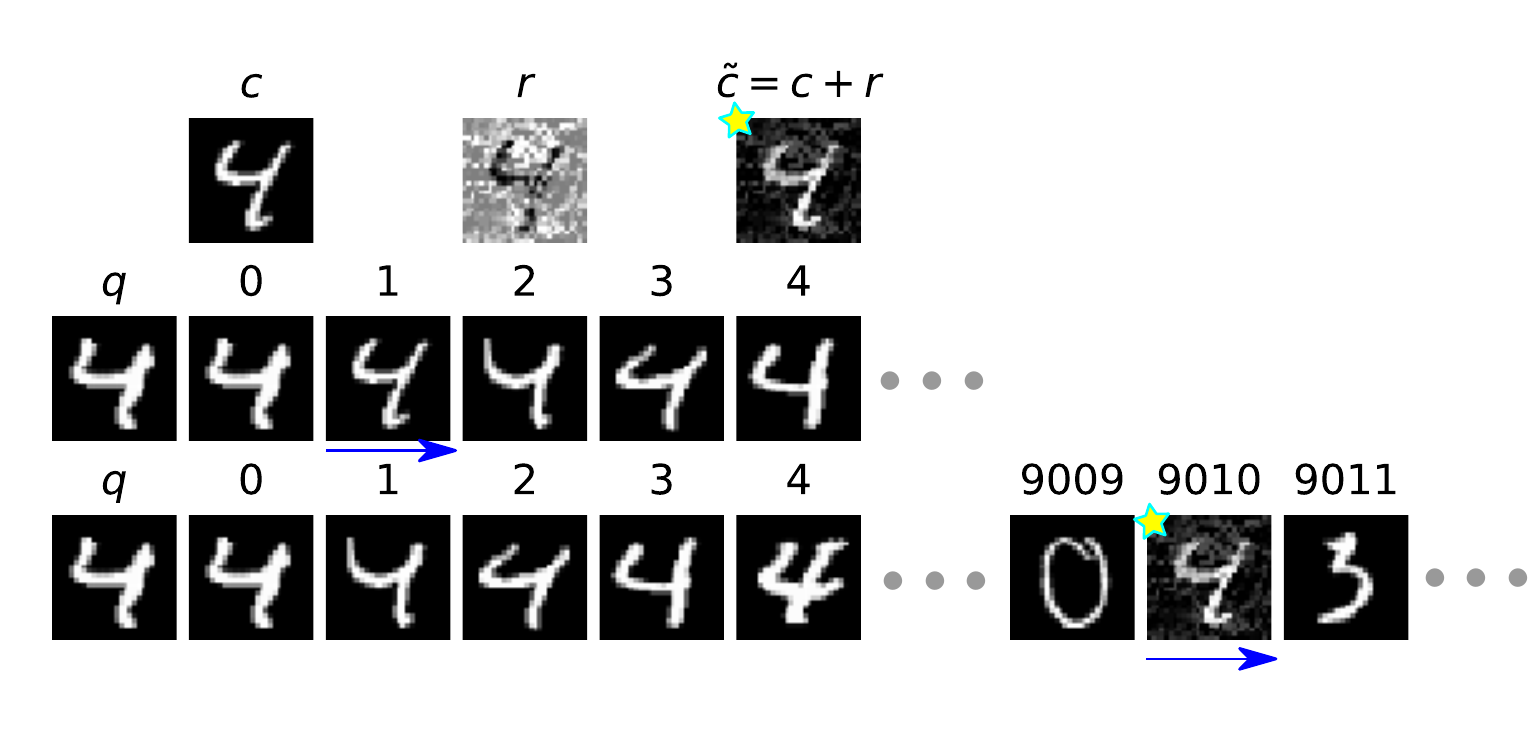}
\caption{CA- on MNIST. Example 1.}
\label{fig:mcm1}
\end{figure}

\begin{figure}[h]
\includegraphics[width=1.0\columnwidth]{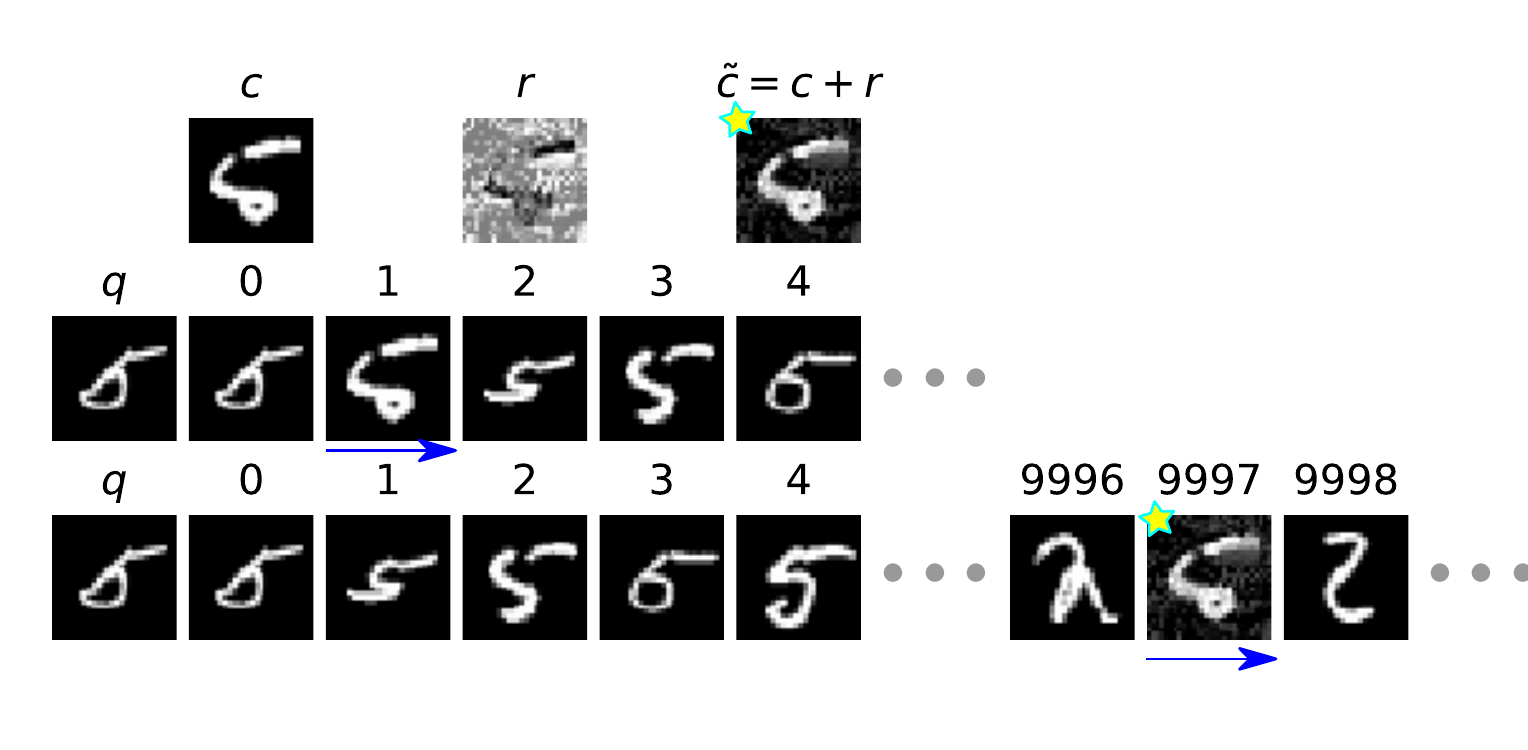}
\caption{CA- on MNIST. Example 2.}
\label{fig:mcm2}
\end{figure}

\begin{figure}[h]
\includegraphics[width=1.0\columnwidth]{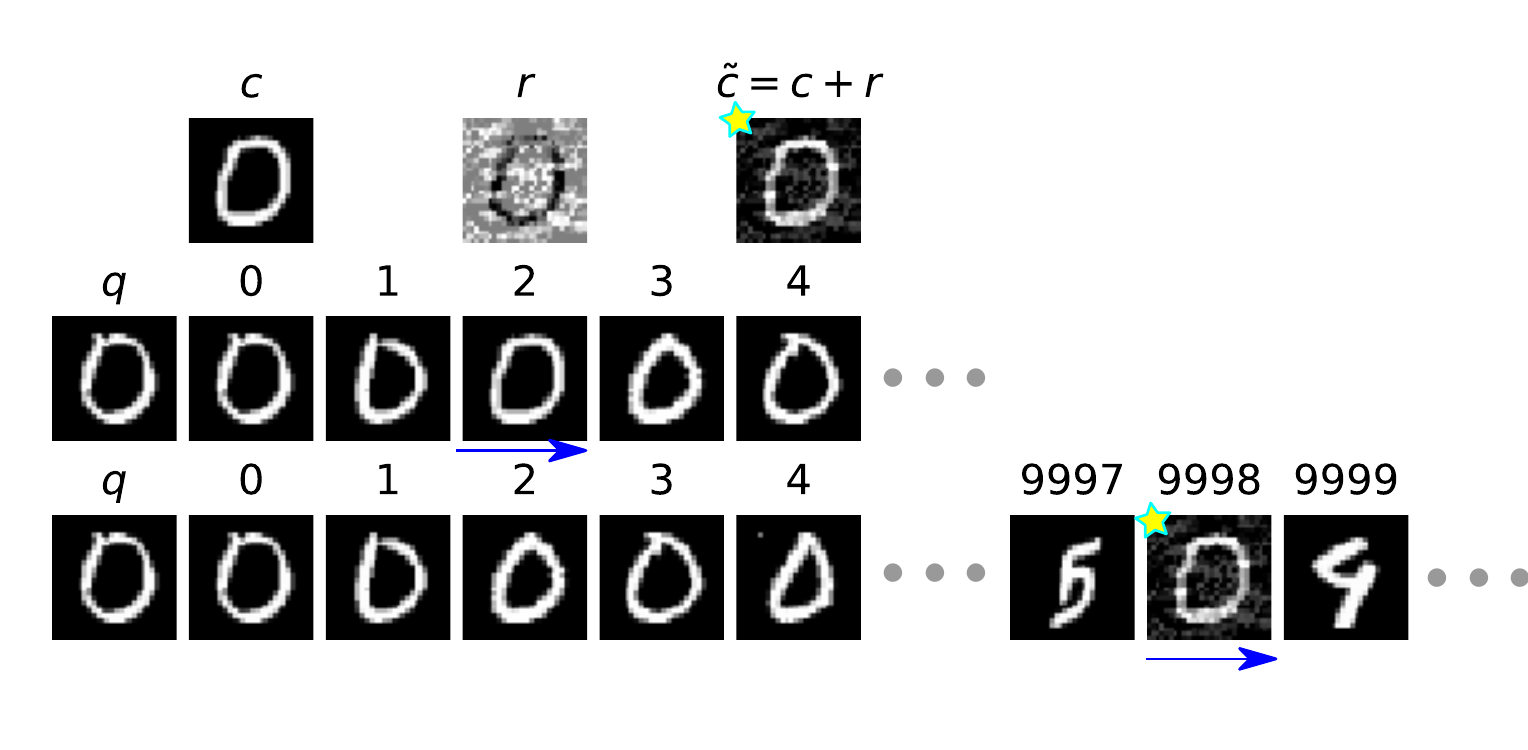}
\caption{CA- on MNIST. Example 3.}
\label{fig:mcm3}
\end{figure}

\textbf{CA+}. See Fig.~\ref{fig:mcp1},\ref{fig:mcp2},\ref{fig:mcp3}.
\textbf{CA-}. See Fig.~\ref{fig:mcm1},\ref{fig:mcm2},\ref{fig:mcm3}.

\begin{figure}[h]
\includegraphics[width=1.0\columnwidth]{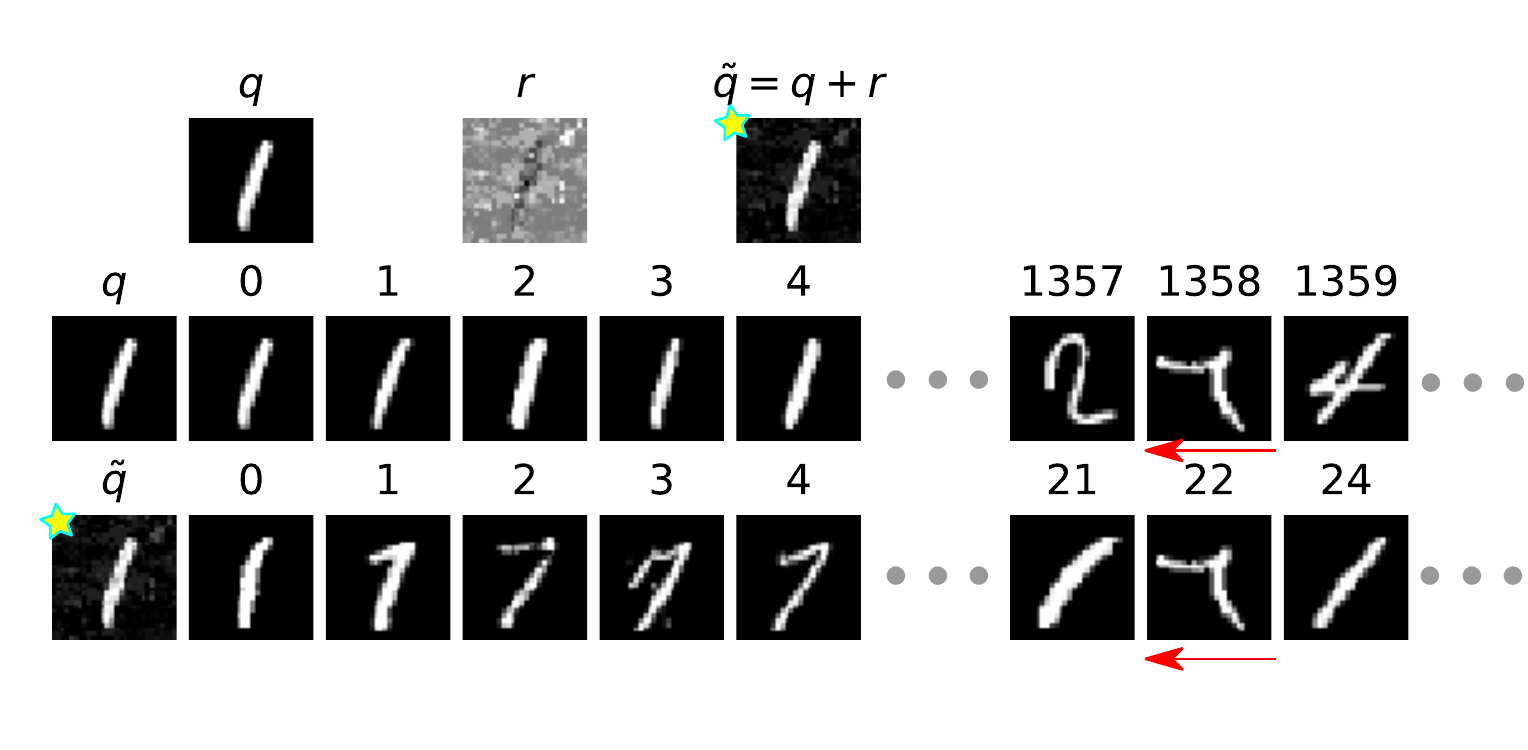}
\caption{QA+ on MNIST. Example 1.}
\label{fig:mqp1}
\end{figure}

\begin{figure}[h]
\includegraphics[width=1.0\columnwidth]{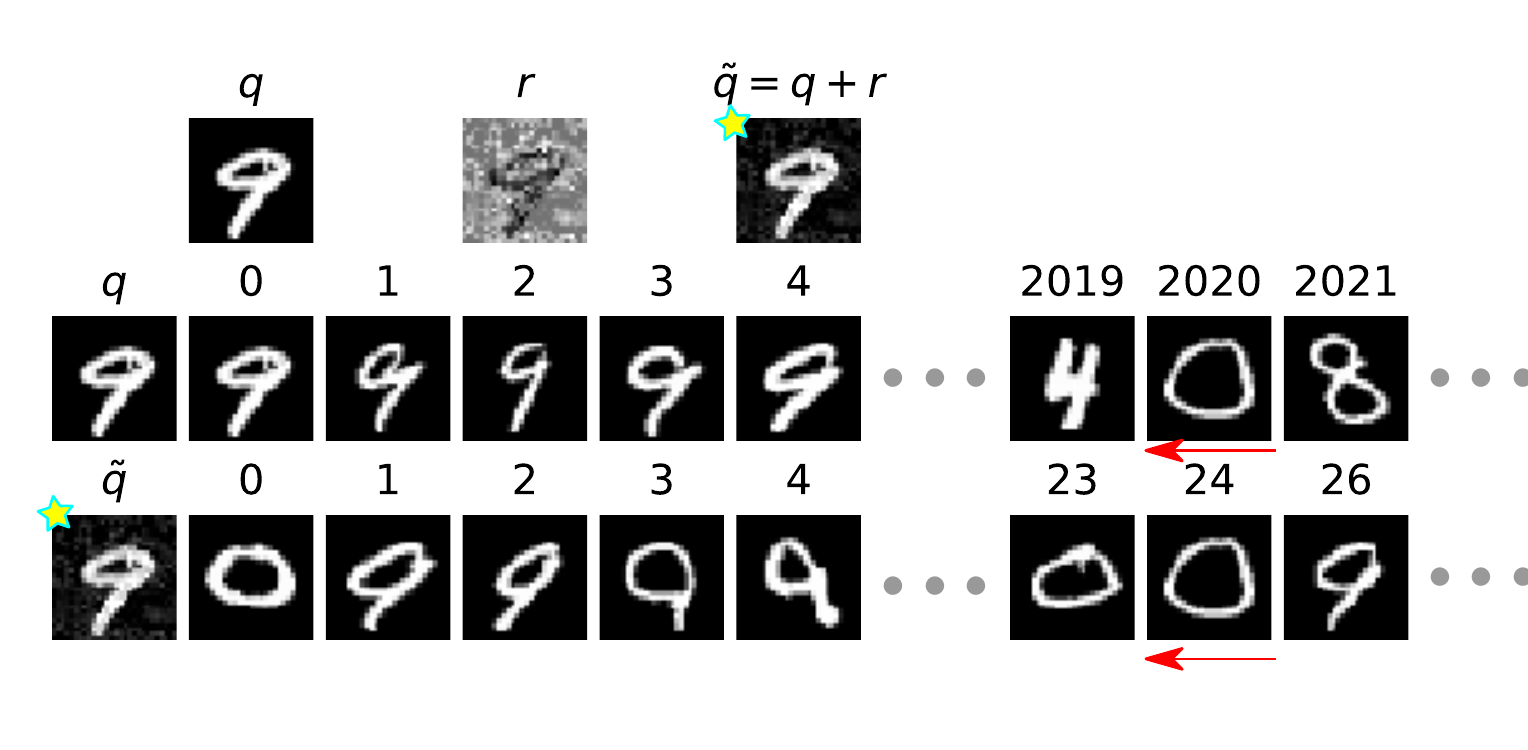}
\caption{QA+ on MNIST. Example 2.}
\label{fig:mqp2}
\end{figure}

\begin{figure}[h]
\includegraphics[width=1.0\columnwidth]{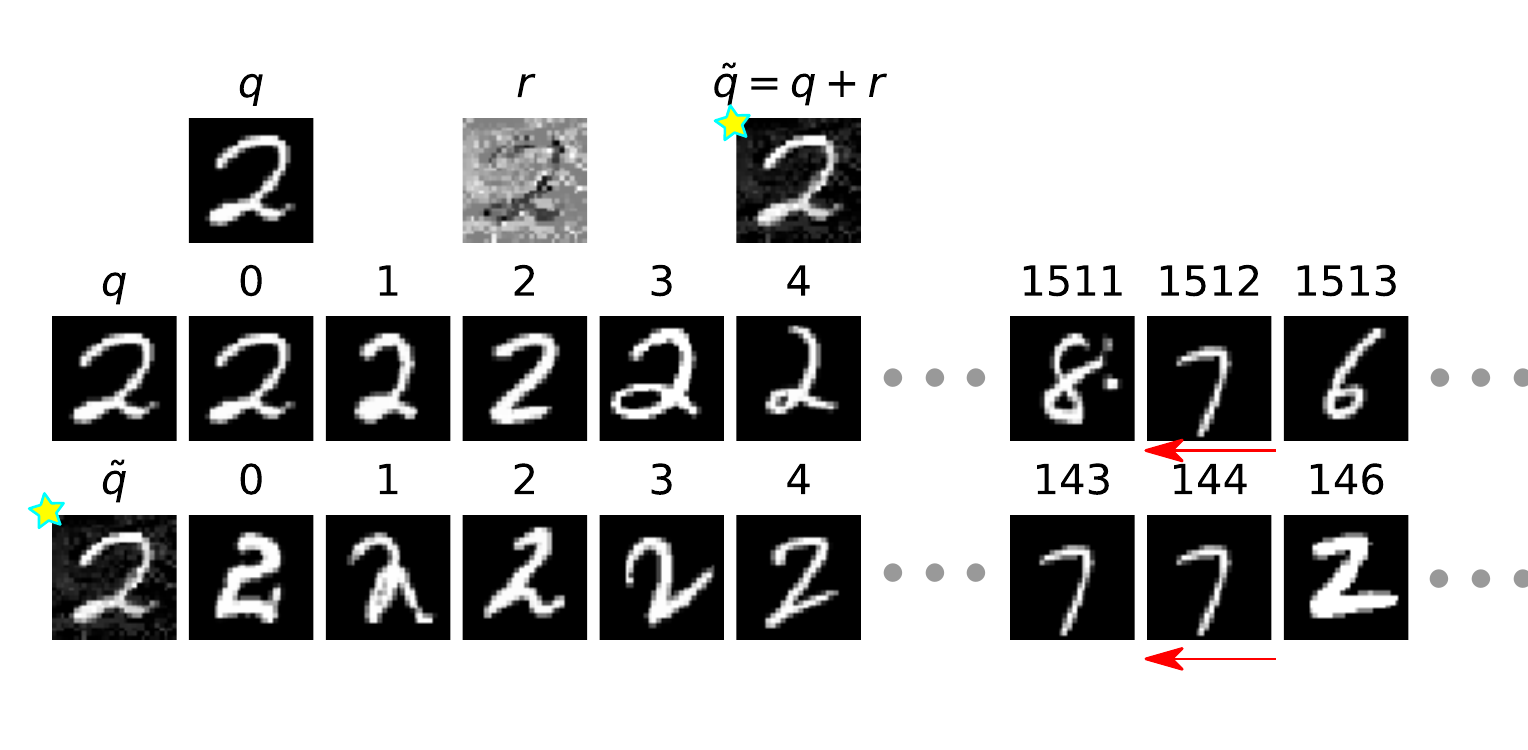}
\caption{QA+ on MNIST. Example 3.}
\label{fig:mqp3}
\end{figure}

\begin{figure}[h]
\includegraphics[width=1.0\columnwidth]{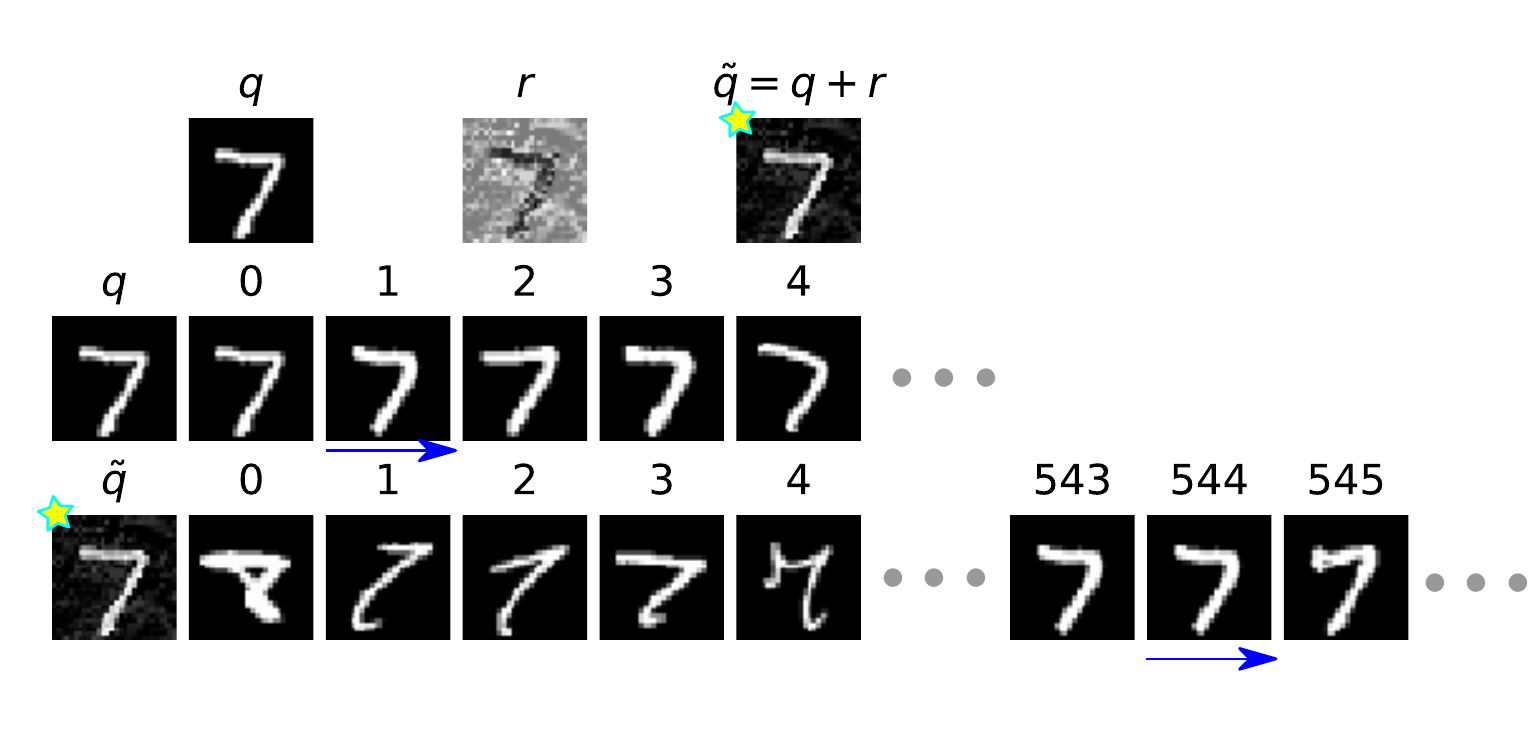}
\caption{QA- on MNIST. Example 1.}
\label{fig:mqm1}
\end{figure}

\begin{figure}[h]
\includegraphics[width=1.0\columnwidth]{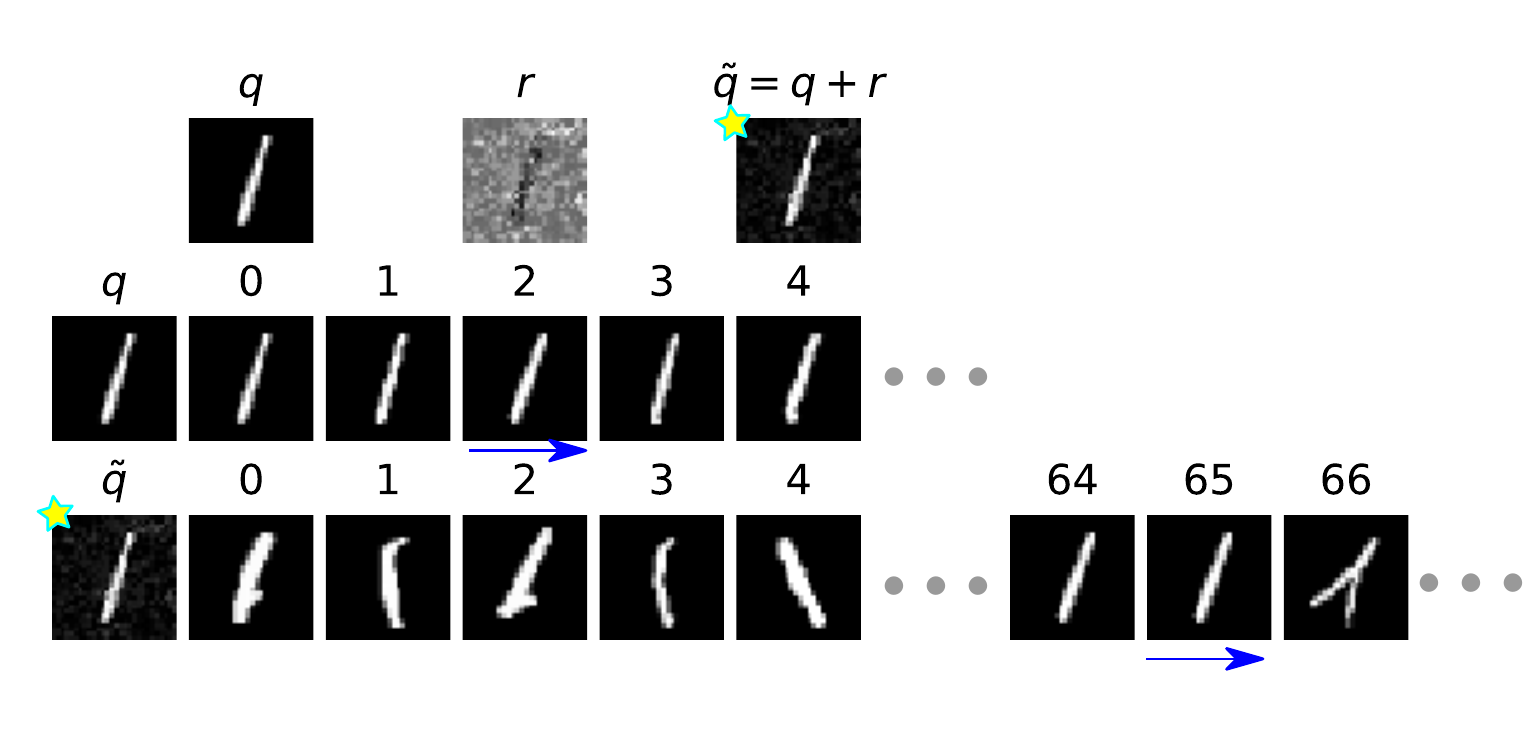}
\caption{QA- on MNIST. Example 2.}
\label{fig:mqm2}
\end{figure}

\begin{figure}[h]
\includegraphics[width=1.0\columnwidth]{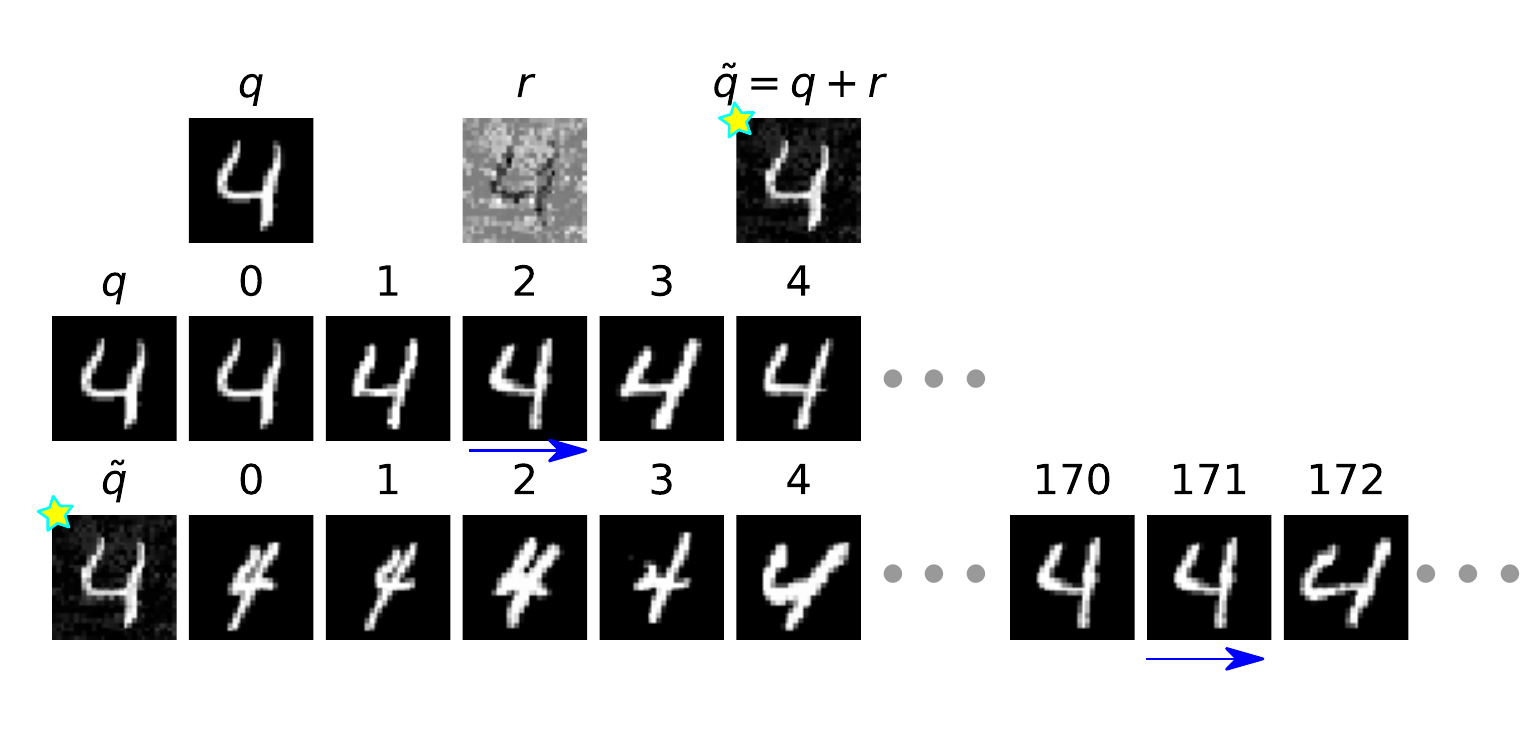}
\caption{QA- on MNIST. Example 3.}
\label{fig:mqm3}
\end{figure}

\textbf{QA+}. See Fig.~\ref{fig:mqp1},\ref{fig:mqp2},\ref{fig:mqp3}.
\textbf{QA-}. See Fig.~\ref{fig:mqm1},\ref{fig:mqm2},\ref{fig:mqm3}.





\subsection{Fashion-MNIST}

\begin{figure}[h]
\includegraphics[width=1.0\columnwidth]{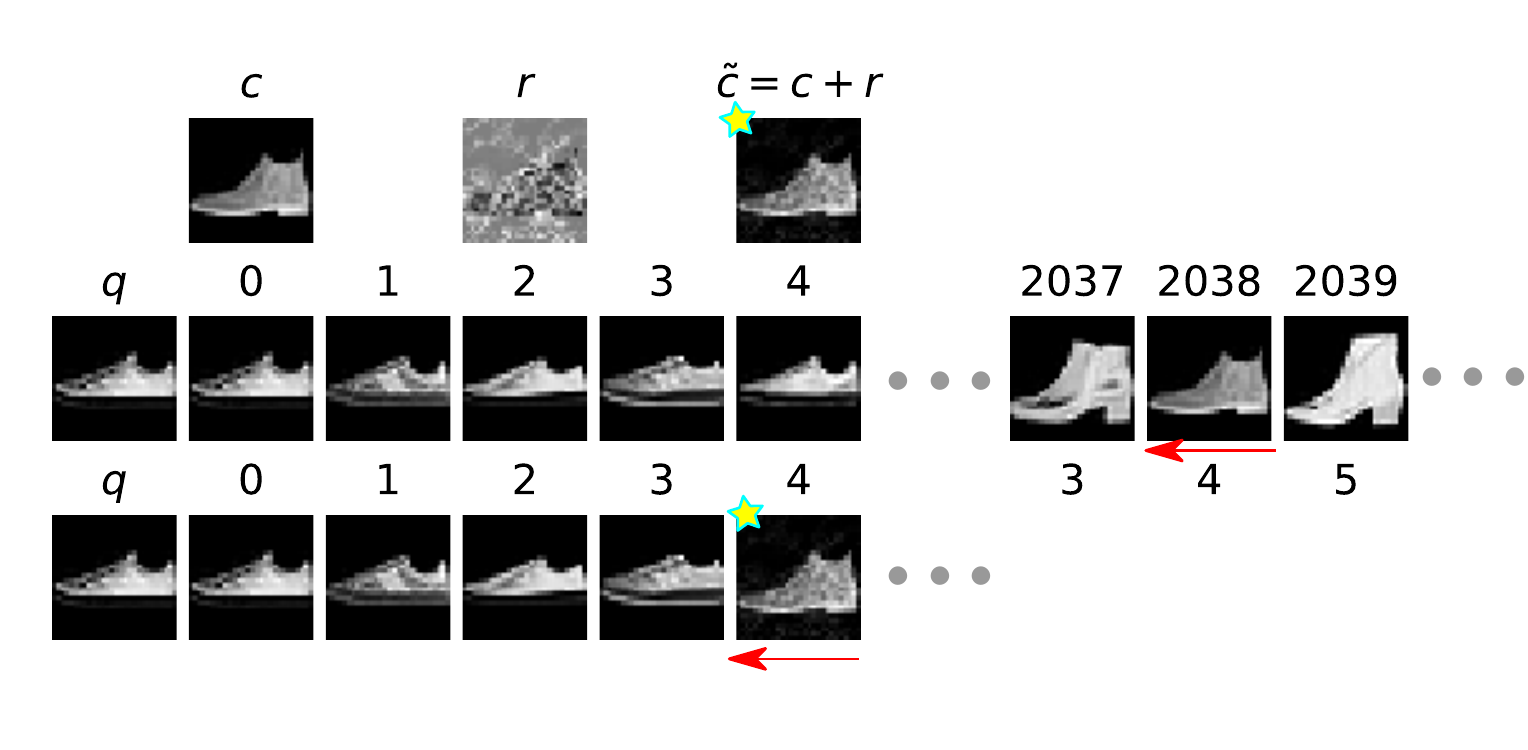}
\caption{CA+ on Fashion-MNIST. Example 1.}
\label{fig:fcp1}
\end{figure}

\begin{figure}[h]
\includegraphics[width=1.0\columnwidth]{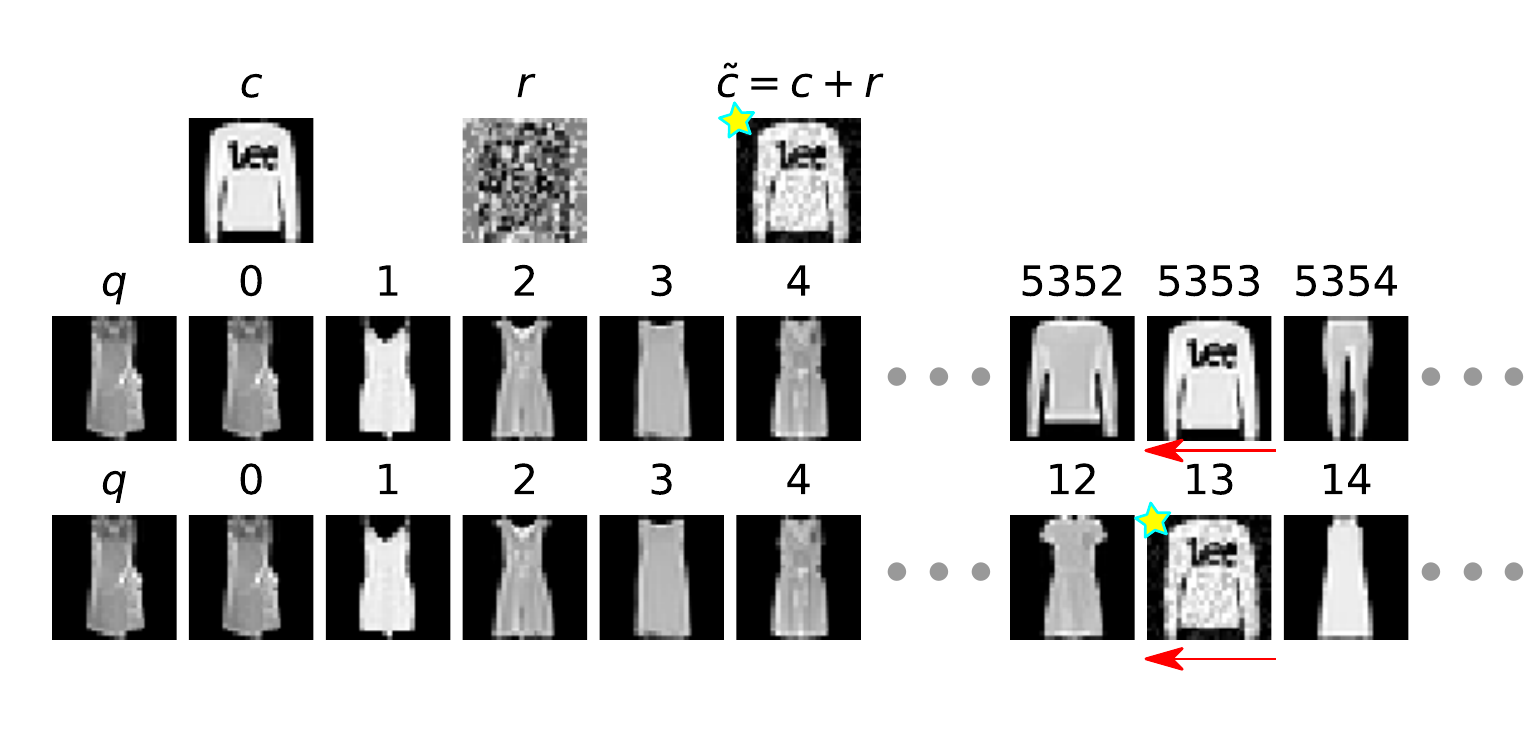}
\caption{CA+ on Fashion-MNIST. Example 2.}
\label{fig:fcp2}
\end{figure}

\begin{figure}[h]
\includegraphics[width=1.0\columnwidth]{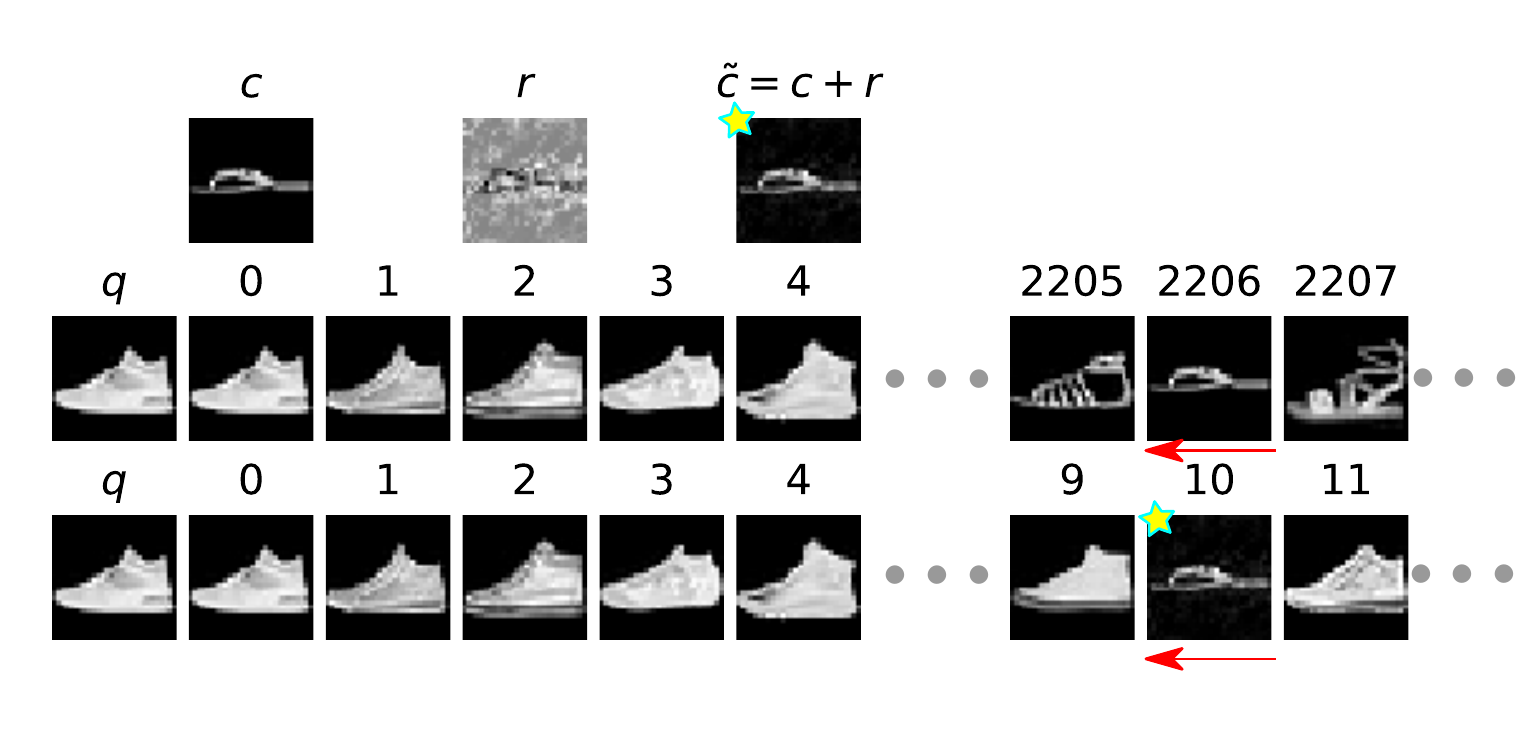}
\caption{CA+ on Fashion-MNIST. Example 3.}
\label{fig:fcp3}
\end{figure}

\begin{figure}[h]
\includegraphics[width=1.0\columnwidth]{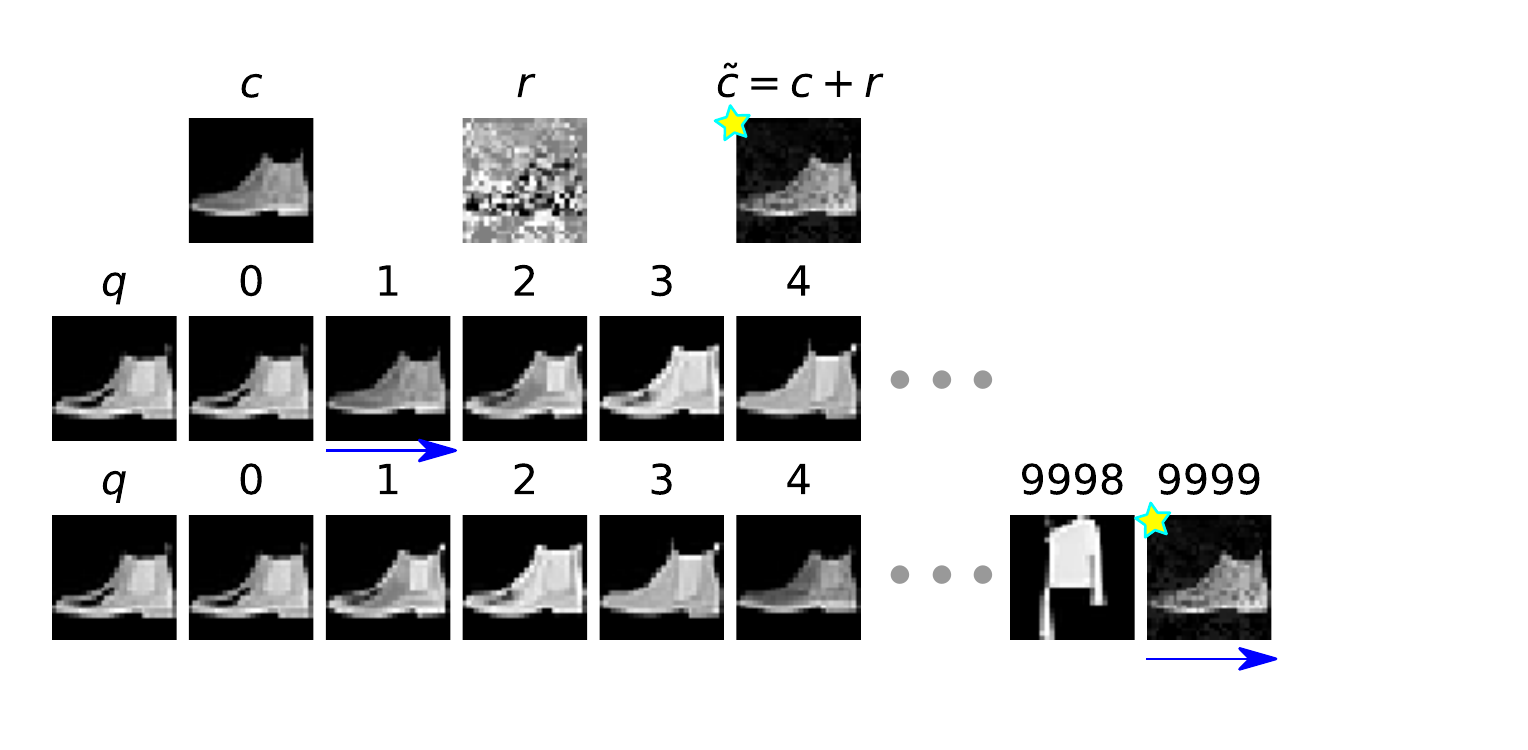}
\caption{CA- on Fashion-MNIST. Example 1.}
\label{fig:fcm1}
\end{figure}

\begin{figure}[h]
\includegraphics[width=1.0\columnwidth]{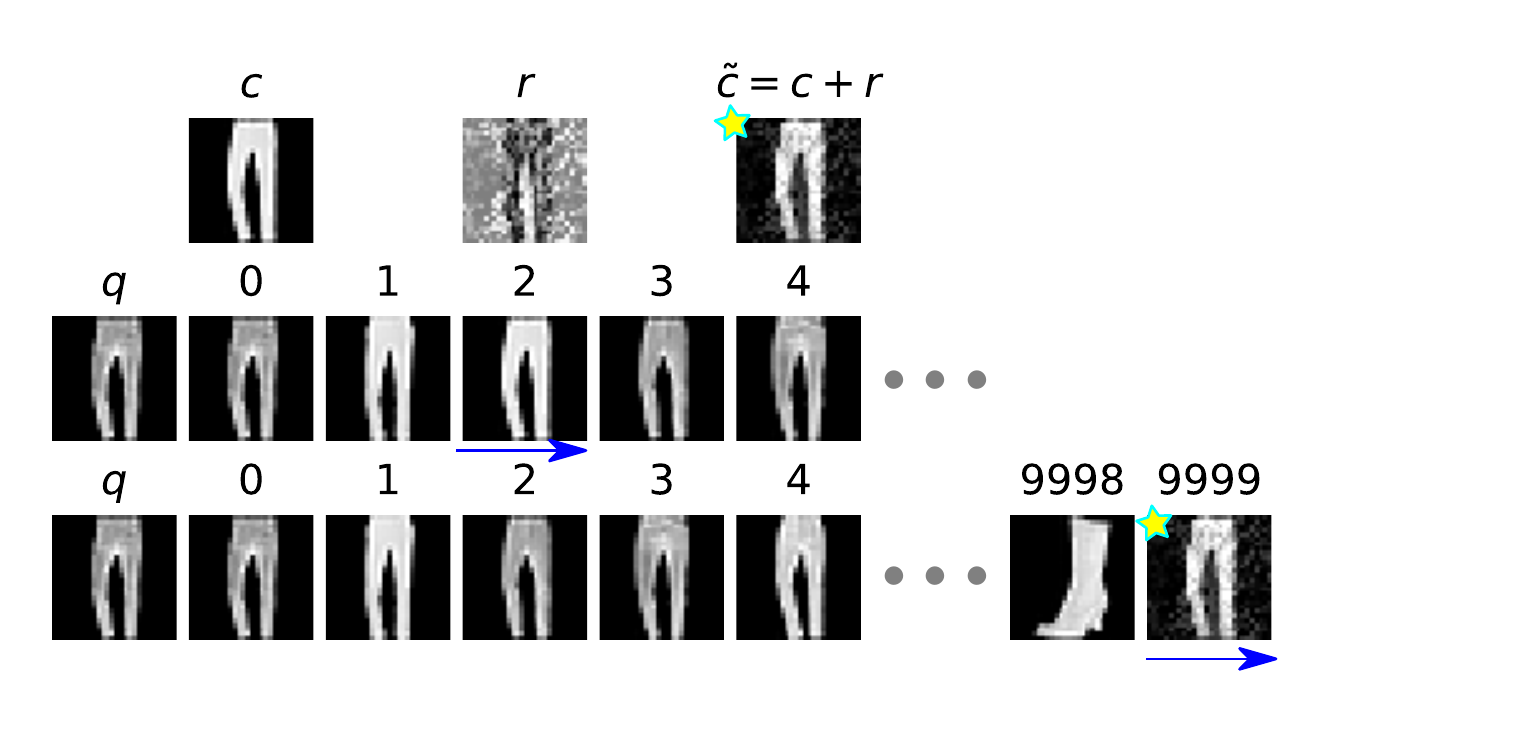}
\caption{CA- on Fashion-MNIST. Example 2.}
\label{fig:fcm2}
\end{figure}

\begin{figure}[h]
\includegraphics[width=1.0\columnwidth]{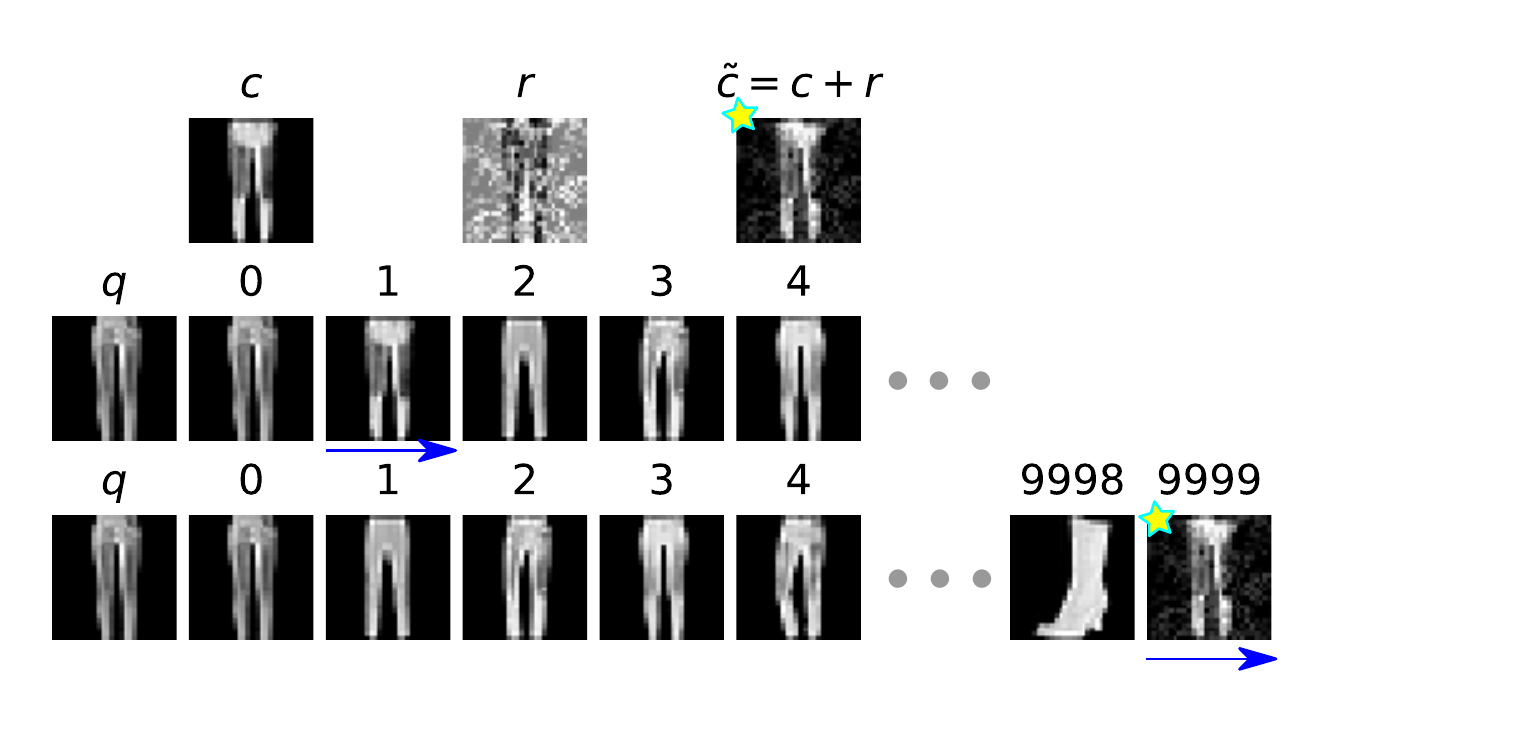}
\caption{CA- on Fashion-MNIST. Example 3.}
\label{fig:fcm3}
\end{figure}

\textbf{CA+}. See Fig.~\ref{fig:fcp1},\ref{fig:fcp2},\ref{fig:fcp3}.
\textbf{CA-}. See Fig.~\ref{fig:fcm1},\ref{fig:fcm2},\ref{fig:fcm3}.

\begin{figure}[h]
\includegraphics[width=1.0\columnwidth]{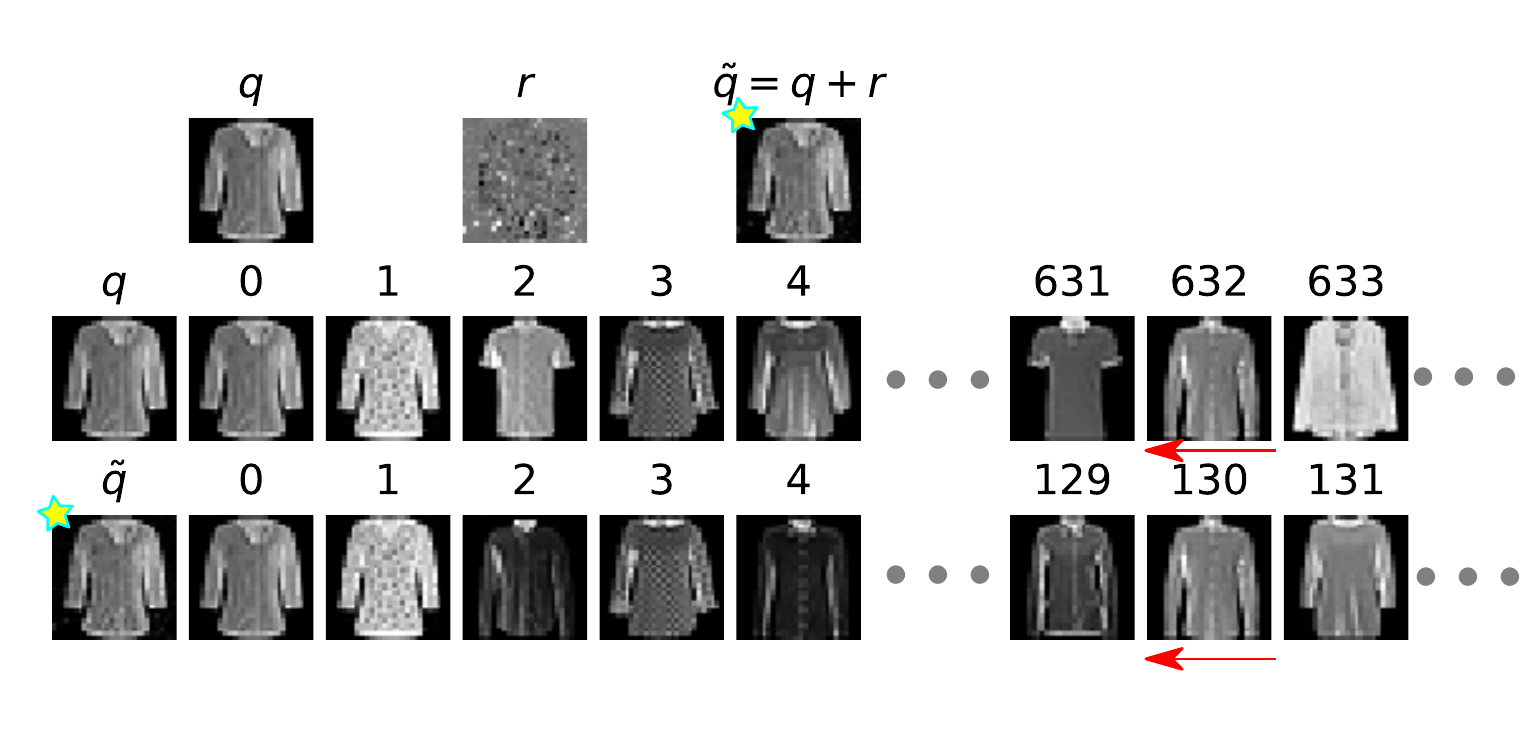}
\caption{QA+ on Fashion-MNIST. Example 1.}
\label{fig:fqp1}
\end{figure}

\begin{figure}[h]
\includegraphics[width=1.0\columnwidth]{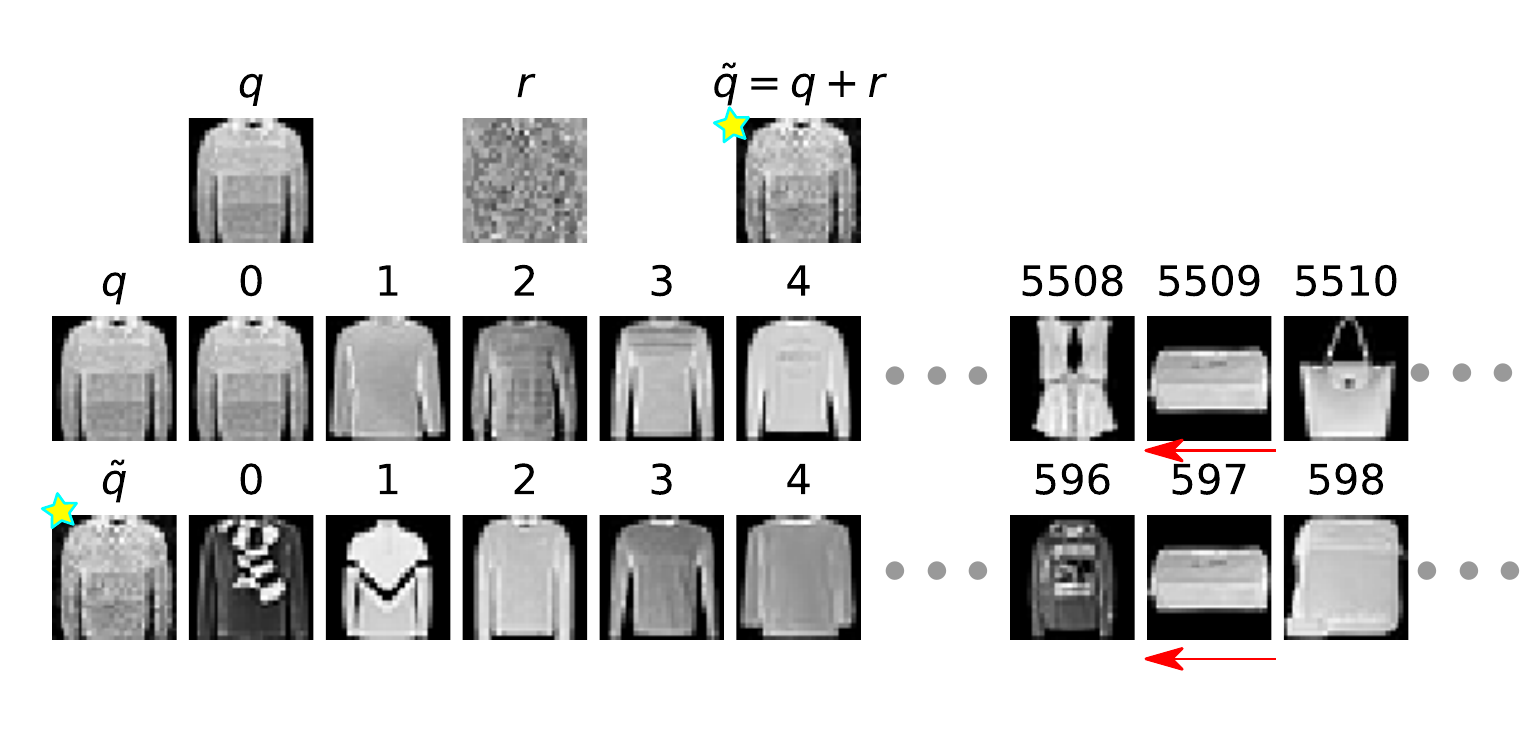}
\caption{QA+ on Fashion-MNIST. Example 2.}
\label{fig:fqp2}
\end{figure}

\begin{figure}[h]
\includegraphics[width=1.0\columnwidth]{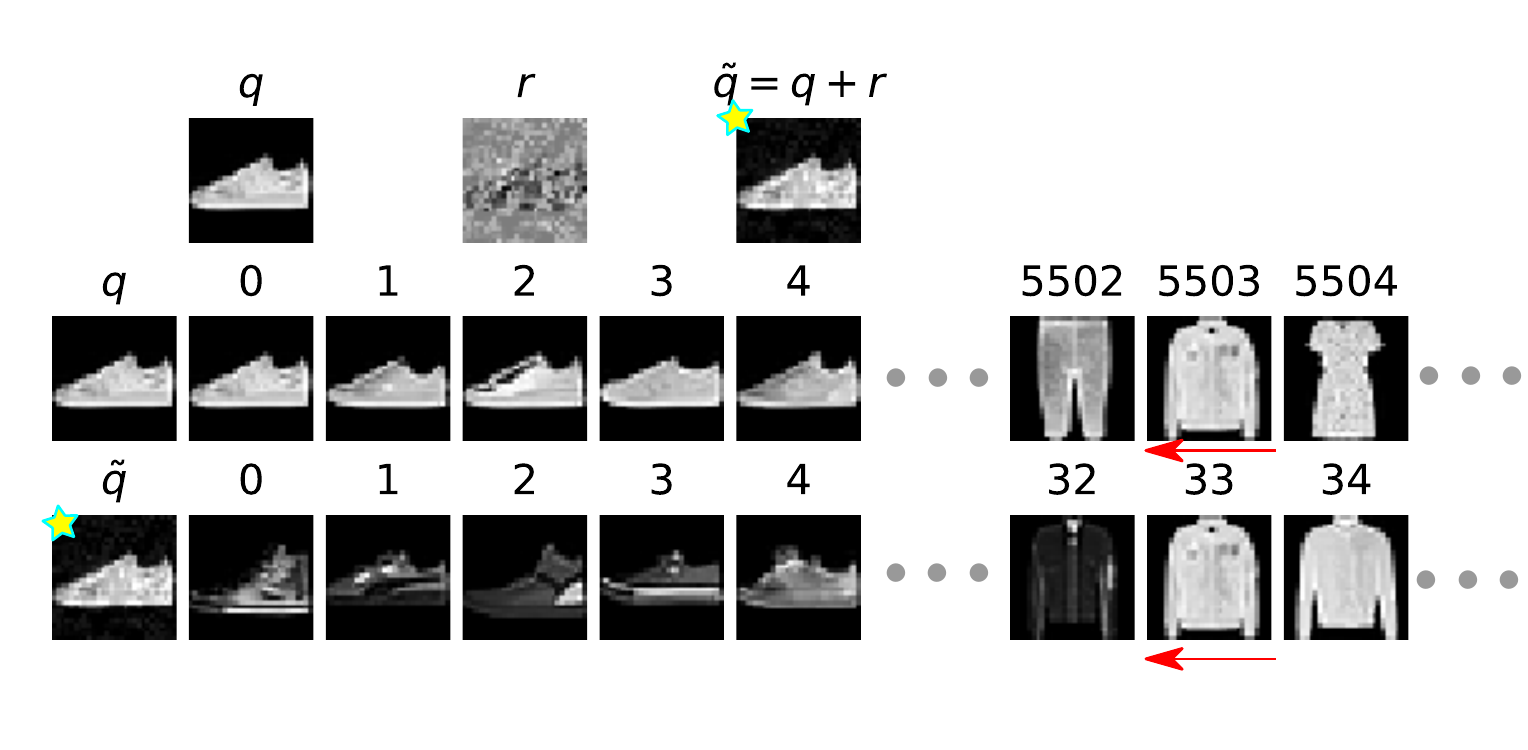}
\caption{QA+ on Fashion-MNIST. Example 3.}
\label{fig:fqp3}
\end{figure}

\begin{figure}[h]
\includegraphics[width=1.0\columnwidth]{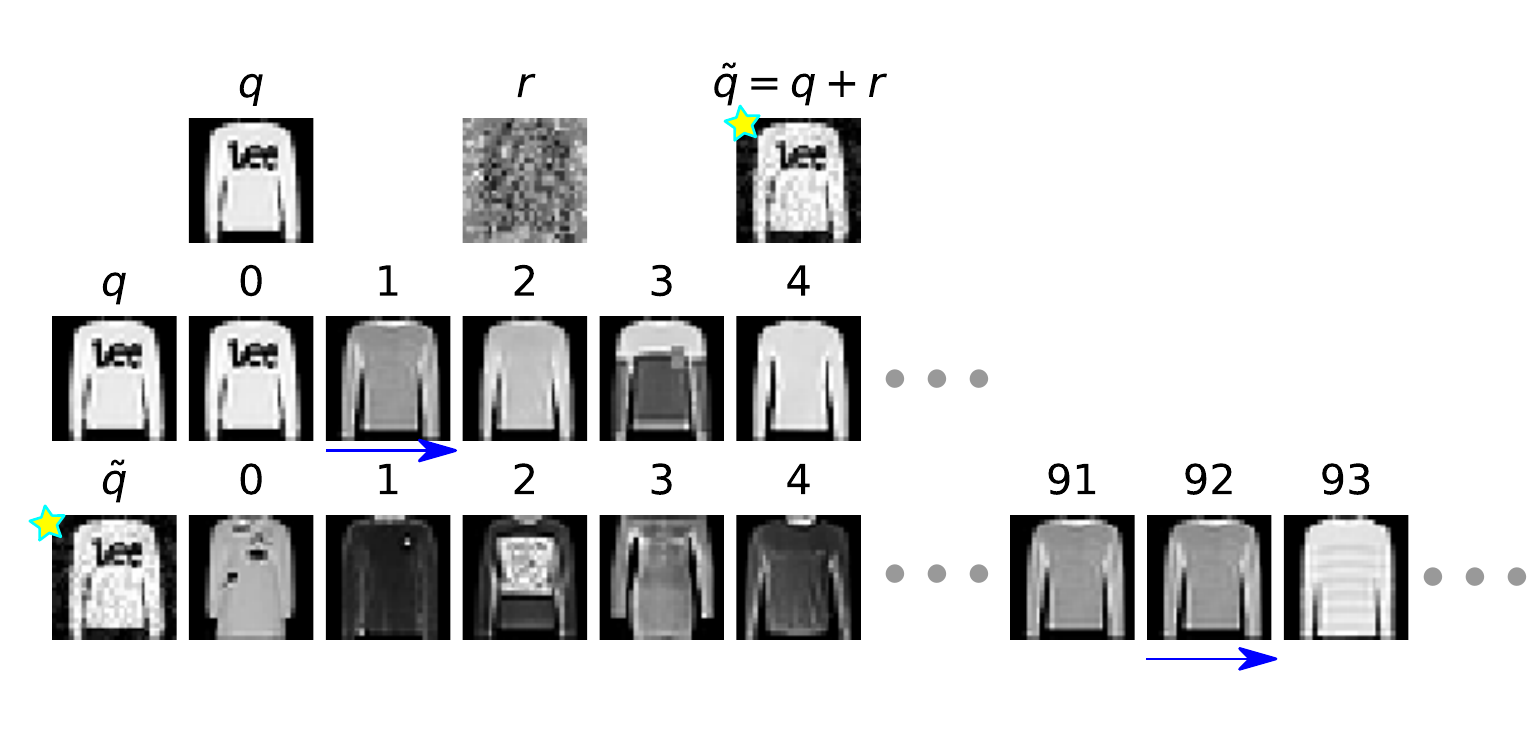}
\caption{QA- on Fashion-MNIST. Example 1.}
\label{fig:fqm1}
\end{figure}

\begin{figure}[h]
\includegraphics[width=1.0\columnwidth]{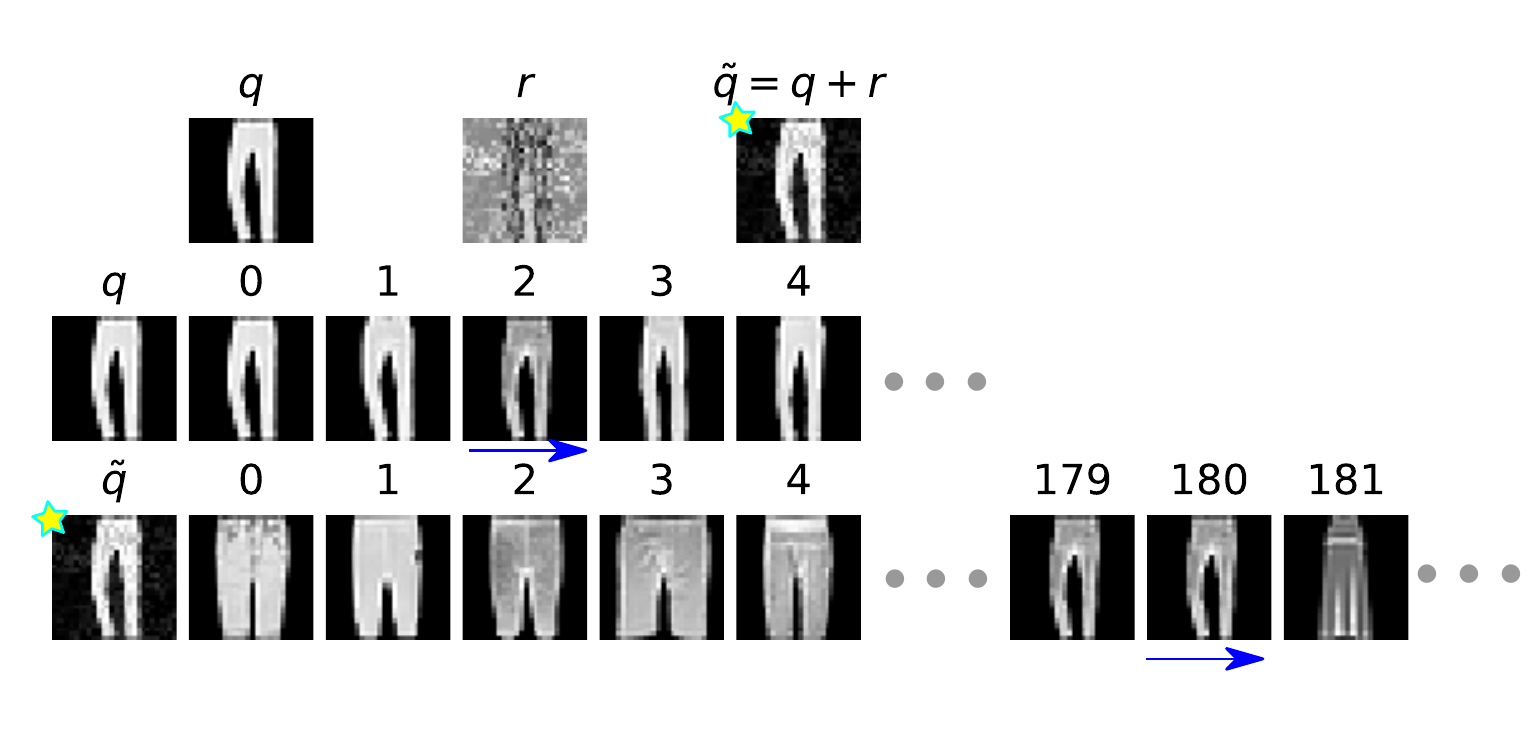}
\caption{QA- on Fashion-MNIST. Example 2.}
\label{fig:fqm2}
\end{figure}

\begin{figure}[h]
\includegraphics[width=1.0\columnwidth]{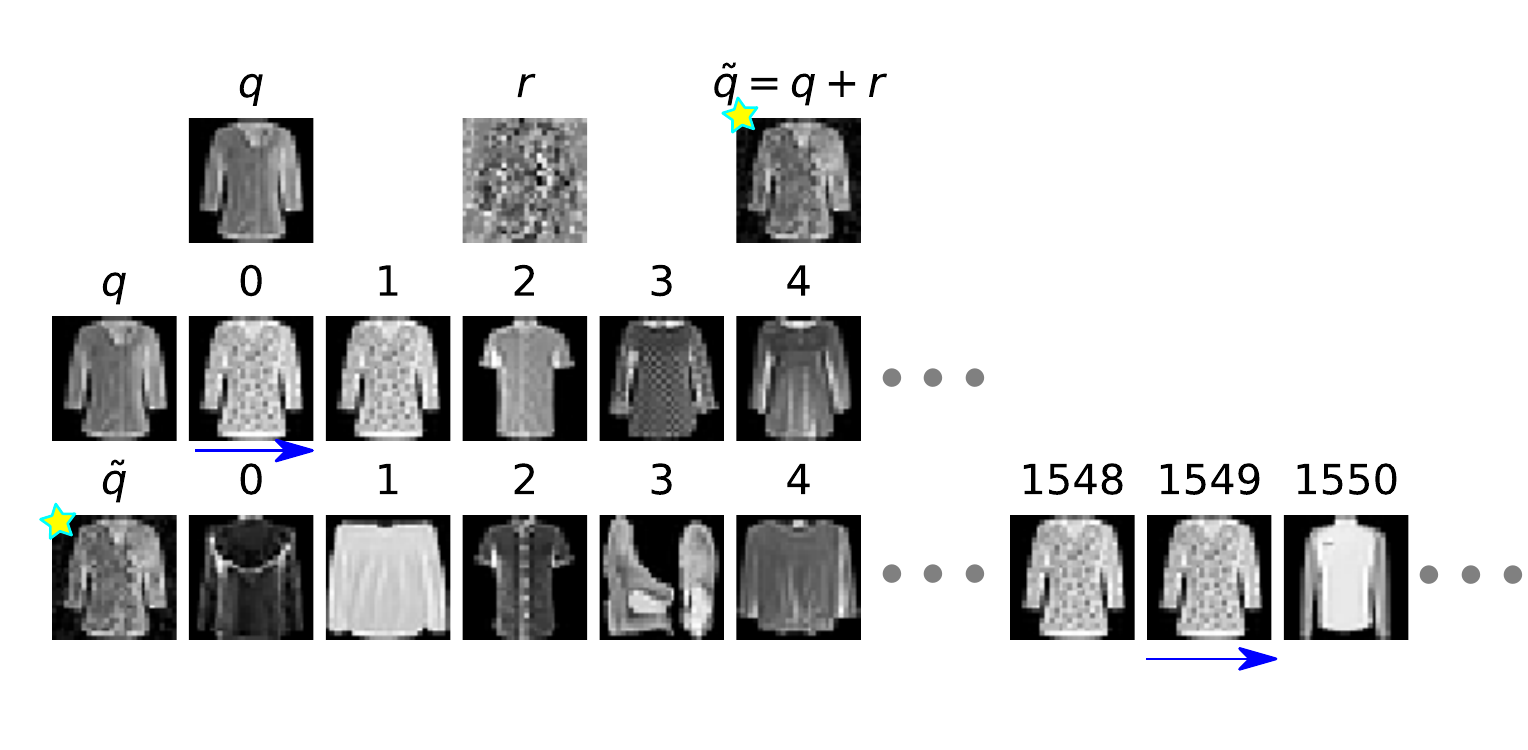}
\caption{QA- on Fashion-MNIST. Example 3.}
\label{fig:fqm3}
\end{figure}

\textbf{QA+}. See Fig.~\ref{fig:fqp1},\ref{fig:fqp2},\ref{fig:fqp3}.
\textbf{QA-}. See Fig.~\ref{fig:fqm1},\ref{fig:fqm2},\ref{fig:fqm3}.





\subsection{Semantic-Preserving for QA}

\begin{figure}[h]
\includegraphics[width=1.0\columnwidth]{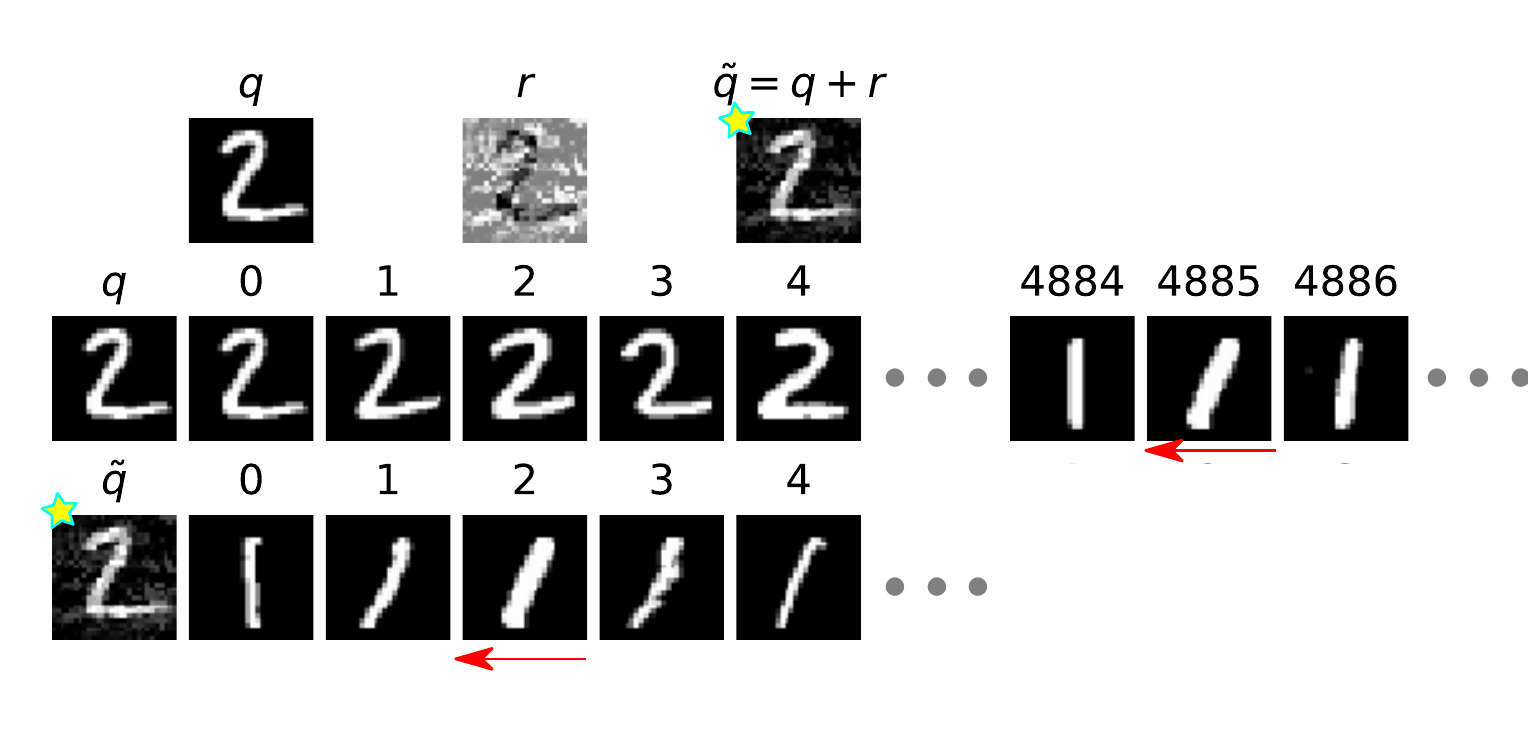}
\caption{QA+ without preserving query semantics.}
\label{fig:nosp}
\end{figure}

Conducting Query Attack without preserving the query semantics will often lead
to irrelevant retrieval results at the top of the ranking list,
as shown in Fig.~\ref{fig:nosp},
which raises red flags and possibly reveals the attack.
Therefore,
the value of the Semantics-Preserving term in QA is substantial, as it keeps the
retrieval results as ``normal'' as possible while achieving the attacking goal.

\clearpage

\section{Complete Results of Attack \& Defense}

Some experiments and details on MNIST, Fashion-MNIST and SOP datasets are
omitted in the manuscript due to limited space.
In this section, we present the complete experimental results on these datasets,
including the average rank of $C_\text{SP}$ during QA. Besides, we also conduct
attacking experiments on ranking models trained with different combinations
of loss functions and distance metrics. For brevity, we denote the (Cosine distance,
Triplet loss) setting as (CT), the (Euclidean distance, Contrastive loss) as (EC).
Models trained with our defense method will have a trailing ``D'' in notation,
\eg the (CT) model with our defense will be denoted as (CTD).

See Tab.~\ref{tab:mnist-complete} for complete results on MNIST.

See Tab.~\ref{tab:fashion-complete} for complete results on Fashion-MNIST.

See Tab.~\ref{tab:sop-complete} for results on SOP.

\begin{table}[!h]
\resizebox{1.0\columnwidth}{!}{
\setlength{\tabcolsep}{0.2em}
\centering
\begin{tabular}{c|cccc|cccc|cccc|cccc}
\toprule
\multirow{2}{*}{$\varepsilon$} & \multicolumn{4}{c|}{CA+} & \multicolumn{4}{c|}{CA-} & \multicolumn{4}{c|}{SP-QA+} & \multicolumn{4}{c}{SP-QA-}\tabularnewline
\cline{2-17} \cline{3-17} \cline{4-17} \cline{5-17} \cline{6-17} \cline{7-17} \cline{8-17} \cline{9-17} \cline{10-17} \cline{11-17} \cline{12-17} \cline{13-17} \cline{14-17} \cline{15-17} \cline{16-17} \cline{17-17}
 & \multicolumn{1}{c|}{$w=1$} & \multicolumn{1}{c|}{$2$} & \multicolumn{1}{c|}{$5$} & $10$ & \multicolumn{1}{c|}{$w=1$} & \multicolumn{1}{c|}{$2$} & \multicolumn{1}{c|}{$5$} & $10$ & \multicolumn{1}{c|}{$m=1$} & \multicolumn{1}{c|}{$2$} & \multicolumn{1}{c|}{$5$} & $10$ & \multicolumn{1}{c|}{$m=1$} & \multicolumn{1}{c|}{$2$} & \multicolumn{1}{c|}{$5$} & $10$\tabularnewline
 \midrule
\rowcolor{black!10}\multicolumn{17}{c}{(ET) Euclidean Distance, Triplet Loss (R@1=63.1\%)}\tabularnewline
0 & 50 & 50 & 50 & 50 & 1.9 & 1.9 & 1.9 & 1.9 & 50 & 50 & 50 & 50 & 0.5 & 0.5 & 0.5 & 0.5\tabularnewline
\hline
0.01 & 0.0 & 0.8 & 2.0 & 2.6 & 99.7 & 99.6 & 99.4 & 99.3 & 4.8,~0.2 & 7.0,~0.7 & 16.3,~1.9 & 25.8,~2.3 & 54.9,~0.3 & 40.2,~0.6 & 27.1,~0.8 & 21.9,~0.9\tabularnewline
0.03 & 0.0 & 0.3 & 1.0 & 1.5 & 100.0 & 100.0 & 100.0 & 100.0 & 1.6,~0.1 & 3.3,~0.5 & 10.0,~2.0 & 19.2,~2.7 & 68.1,~0.3 & 52.4,~0.6 & 36.6,~0.8 & 30.1,~1.0\tabularnewline
0.06 & \textbf{0.0} & \textbf{0.2} & \textbf{1.0} & \textbf{1.5} & \textbf{100.0} & \textbf{100.0} & \textbf{100.0} & \textbf{100.0} & \textbf{1.1},~0.2 & \textbf{2.7},~0.6 & \textbf{8.8},~1.9 & \textbf{17.6},~3.3 & \textbf{73.8},~0.4 & \textbf{57.9},~0.7 & \textbf{40.3},~0.8 & \textbf{32.4},~1.0\tabularnewline
\hline
\rowcolor{blue!10}\multicolumn{17}{c}{(ETD) Euclidean Distance, Triplet Loss, Defensive (R@1=46.4\%)}\tabularnewline
0 & 50 & 50 & 50 & 50 & 2.0 & 2.0 & 2.0 & 2.0 & 50 & 50 & 50 & 50 & 0.5 & 0.5 & 0.5 & 0.5\tabularnewline
\hline
0.01 & 7.5 & 12.2 & 16.5 & 18.0 & 66.4 & 62.6 & 59.3 & 57.8 & 16.1,~2.1 & 24.8,~2.7 & 36.1,~2.7 & 41.4,~2.5 & 26.7,~0.6 & 18.1,~0.7 & 12.2,~0.7 & 10.2,~0.7 \tabularnewline
0.03 & 0.7 & 4.5 & 8.7 & 10.4 & 91.7 & 90.2 & 89.1 & 88.4 & 7.9,~2.9 & 14.5,~4.2 & 27.2,~5.4 & 35.6,~5.3 & 43.4,~1.3 & 31.7,~1.5 & 21.9,~1.7 & 18.1,~1.8 \tabularnewline
0.06 & \textbf{0.1} & \textbf{3.8} & \textbf{7.9} & \textbf{9.7} & \textbf{97.3} & \textbf{96.8} & \textbf{96.4} & \textbf{96.2} & \textbf{6.9},~4.3 & \textbf{12.5},~5.8 & \textbf{24.3},~7.0 & \textbf{33.4},~6.9 & \textbf{51.4},~2.8 & \textbf{39.0},~3.2 & \textbf{28.0},~3.5 & \textbf{23.5},~3.6 \tabularnewline
\bottomrule
\end{tabular}

}

\caption{Experiments on SOP dataset.}
\label{tab:sop-complete}
\end{table}

\begin{table}[!ht]
\centering
\resizebox{1.0\columnwidth}{!}{%
\setlength{\tabcolsep}{0.2em}

\begin{tabular}{c|cccc|cccc|cccc|cccc}
\toprule
\multirow{2}{*}{$\varepsilon$} & \multicolumn{4}{c|}{CA+} & \multicolumn{4}{c|}{CA-} & \multicolumn{4}{c|}{SP-QA+} & \multicolumn{4}{c}{SP-QA-}\tabularnewline
\cline{2-17} \cline{3-17} \cline{4-17} \cline{5-17} \cline{6-17} \cline{7-17} \cline{8-17} \cline{9-17} \cline{10-17} \cline{11-17} \cline{12-17} \cline{13-17} \cline{14-17} \cline{15-17} \cline{16-17} \cline{17-17}
 & \multicolumn{1}{c|}{$w=1$} & \multicolumn{1}{c|}{$2$} & \multicolumn{1}{c|}{$5$} & $10$ & \multicolumn{1}{c|}{$w=1$} & \multicolumn{1}{c|}{$2$} & \multicolumn{1}{c|}{$5$} & $10$ & \multicolumn{1}{c|}{$m=1$} & \multicolumn{1}{c|}{$2$} & \multicolumn{1}{c|}{$5$} & $10$ & \multicolumn{1}{c|}{$m=1$} & \multicolumn{1}{c|}{$2$} & \multicolumn{1}{c|}{$5$} & $10$\tabularnewline
\midrule
\rowcolor{black!10}\multicolumn{17}{c}{(CC) Cosine Distance, Contrastive Loss (R@1=98.6\%)}\tabularnewline
0 & 50 & 50 & 50 & 50 & 2.2 & 2.2 & 2.2 & 2.2 & 50 & 50 & 50 & 50 & 0.5 & 0.5 & 0.5 & 0.5\tabularnewline
\hline
0.01 & 42.2 & 43.3 & 44.6 & 44.7 & 5.6 & 5.3 & 5.2 & 5.1 & 45.0,~0.5 & 46.9,~0.5 & 48.4,~0.5 & 49.1,~0.5 & 1.2,~0.5 & 1.2,~0.5 & 1.1,~0.5 & 1.1,~0.5\tabularnewline
0.03 & 31.3 & 33.5 & 35.4 & 36.1 & 9.3 & 9.2 & 9.1 & 9.1 & 37.6,~2.3 & 41.7,~2.5 & 45.9,~2.5 & 47.7,~2.5 & 2.8,~1.8 & 2.7,~1.8 & 2.7,~1.8 & 2.6,~1.8\tabularnewline
0.1 & 9.7 & 12.5 & 14.6 & 15.6 & 52.1 & 52.2 & 52.4 & 52.5 & 18.5,~8.2 & 27.1,~9.7 & 38.0,~9.9 & 43.4,~9.9 & 5.6,~4.1 & 5.4,~4.1 & 5.4,~4.1 & 5.4,~4.1\tabularnewline
0.3 & \textbf{5.8} & \textbf{10.0} & \textbf{11.9} & \textbf{12.9} & \textbf{99.0} & \textbf{99.1} & \textbf{99.0} & \textbf{99.1} & \textbf{14.3},~9.7 & \textbf{23.3},~12.5 & \textbf{36.3},~13.8 & \textbf{42.6},~13.9 & \textbf{6.1},~4.4 & \textbf{5.8},~4.4 & \textbf{5.7},~4.4 & \textbf{5.7},~4.4\tabularnewline
\hline
\rowcolor{black!10}\multicolumn{17}{c}{(CT) Cosine Distance, Triplet Loss (R@1=99.1\%)}\tabularnewline
0 & 50 & 50 & 50 & 50 & 2.1 & 2.1 & 2.1 & 2.1 & 50 & 50 & 50 & 50 & 0.5 & 0.5 & 0.5 & 0.5\tabularnewline
\hline
0.01 & 44.6 & 45.4 & 47.4 & 47.9 & 3.4 & 3.2 & 3.1 & 3.1 & 45.2,~0.0 & 46.3,~0.0 & 47.7,~0.0 & 48.5,~0.0 & 0.9,~0.0 & 0.7,~0.0 & 0.6,~0.0 & 0.6,~0.1\tabularnewline
0.03 & 33.4 & 37.3 & 41.9 & 43.9 & 6.3 & 5.9 & 5.7 & 5.6 & 35.6,~0.3 & 39.2,~0.3 & 43.4,~0.3 & 45.8,~0.3 & 1.9,~0.2 & 1.4,~0.2 & 1.1,~0.2 & 1.1,~0.2\tabularnewline
0.1 & 12.7 & 17.4 & 24.4 & 30.0 & 15.4 & 14.9 & 14.8 & 14.7 & 14.4,~2.2 & 21.0,~2.3 & 30.6,~2.3 & 37.2,~2.3 & 5.6,~1.2 & 4.4,~1.2 & 3.7,~1.2 & 3.5,~1.2\tabularnewline
0.3 & \textbf{2.1} & \textbf{9.1} & \textbf{13.0} & \textbf{17.9} & \textbf{93.9} & \textbf{93.2} & \textbf{93.0} & \textbf{92.9} & \textbf{6.3},~3.6 & \textbf{11.2},~5.7 & \textbf{22.5},~7.7 & \textbf{32.1},~7.7 & \textbf{8.6},~1.6 & \textbf{6.6},~1.6 & \textbf{5.3},~1.5 & \textbf{4.8},~1.5\tabularnewline
\hline
\rowcolor{black!10}\multicolumn{17}{c}{(EC) Euclidean Distance, Contrastive Loss (R@1=99.0\%)}\tabularnewline
0 & 50 & 50 & 50 & 50 & 1.8 & 1.8 & 1.8 & 1.8 & 50 & 50 & 50 & 50 & 0.5 & 0.5 & 0.5 & 0.5\tabularnewline
\hline
0.01 & 31.9 & 33.5 & 34.6 & 34.9 & 8.2 & 8.0 & 7.9 & 7.8 & 41.3,~2.0 & 44.0,~2.2 & 47.2,~2.4 & 48.4,~2.5 & 2.6,~1.3 & 2.4,~1.3 & 2.2,~1.3 & 2.2,~1.3\tabularnewline
0.03 & 15.8 & 17.4 & 18.7 & 19.1 & 10.7 & 10.6 & 10.5 & 10.5 & 27.2,~4.8 & 34.6,~5.4 & 42.0,~5.6 & 45.6,~5.8 & 4.1,~2.5 & 3.9,~2.5 & 3.7,~2.5 & 3.7,~2.5\tabularnewline
0.1 & 6.7 & 10.0 & 12.1 & 12.7 & 82.1 & 81.6 & 81.7 & 82.0 & 15.7,~9.8 & 25.4,~12.0 & 37.7,~13.1 & 43.2,~13.1 & 5.6,~3.2 & 5.3,~3.2 & 4.9,~3.2 & 4.9,~3.2\tabularnewline
0.3 & \textbf{4.8} & \textbf{9.9} & \textbf{12.1} & \textbf{12.7} & \textbf{99.8} & \textbf{99.8} & \textbf{99.8} & \textbf{99.8} & \textbf{14.4},~9.4 & \textbf{23.9},~12.5 & \textbf{36.7},~13.6 & \textbf{42.6},~13.8 & \textbf{5.9},~3.2 & \textbf{5.4},~3.2 & \textbf{5.0},~3.2 & \textbf{4.9},~3.1\tabularnewline
\hline
\rowcolor{black!10}\multicolumn{17}{c}{(ET) Euclidean Distance, Triplet Loss (R@1=99.2\%)}\tabularnewline
0 & 50 & 50 & 50 & 50 & 1.6 & 1.6 & 1.6 & 1.6 & 50 & 50 & 50 & 50 & 0.5 & 0.5 & 0.5 & 0.5\tabularnewline
\hline
0.01 & 39.0 & 40.4 & 40.6 & 40.8 & 3.2 & 3.0 & 2.8 & 2.8 & 45.5,~0.1 & 47.4,~0.1 & 48.1,~0.1 & 49.0,~0.1 & 1.0,~0.1 & 0.9,~0.1 & 0.8,~0.1 & 0.8,~0.1\tabularnewline
0.03 & 23.5 & 25.6 & 26.5 & 27.0 & 7.0 & 6.6 & 6.3 & 6.2 & 36.4,~0.6 & 40.9,~0.7 & 45.1,~0.8 & 47.1,~0.9 & 2.7,~0.7 & 2.3,~0.7 & 2.0,~0.7 & 2.0,~0.8\tabularnewline
0.1 & 8.1 & 10.6 & 12.1 & 12.7 & 13.7 & 13.3 & 13.0 & 12.9 & 14.1,~4.8 & 23.4,~5.7 & 34.6,~6.2 & 40.8,~6.5 & 6.2,~1.7 & 5.1,~1.7 & 4.4,~1.7 & 4.2,~1.7\tabularnewline
0.3 & \textbf{1.8} & \textbf{8.5} & \textbf{10.9} & \textbf{11.7} & \textbf{77.4} & \textbf{75.7} & \textbf{75.0} & \textbf{74.7} & \textbf{7.5},~5.2 & \textbf{15.0},~7.4 & \textbf{28.1},~9.0 & \textbf{36.3},~8.9 & \textbf{7.4},~1.7 & \textbf{5.8},~1.7 & \textbf{4.9},~1.7 & \textbf{4.6},~1.7\tabularnewline
\midrule
\rowcolor{blue!10}\multicolumn{17}{c}{(CCD) Cosine Distance, Contrastive Loss, Defensive (R@1=97.5\%)}\tabularnewline
0 & 50 & 50 & 50 & 50 & 2.2 & 2.2 & 2.2 & 2.2 & 50 & 50 & 50 & 50 & 0.5 & 0.5 & 0.5 & 0.5\tabularnewline
\hline
0.01 & 49.3 & 49.0 & 49.3 & 49.3 & 2.4 & 2.4 & 2.3 & 2.3 & 49.7,~0.0 & 49.6,~0.0 & 49.8,~0.0 & 49.9,~0.0 & 0.5,~0.0 & 0.5,~0.0 & 0.5,~0.0 & 0.5,~0.0\tabularnewline
0.03 & 47.0 & 47.9 & 48.1 & 48.0 & 2.8 & 2.7 & 2.7 & 2.7 & 48.3,~0.0 & 49.1,~0.0 & 49.1,~0.0 & 49.4,~0.0 & 0.6,~0.0 & 0.5,~0.0 & 0.5,~0.0 & 0.5,~0.0\tabularnewline
0.1 & 42.3 & 43.3 & 44.0 & 44.4 & 4.2 & 4.0 & 3.9 & 3.8 & 44.9~,0.1 & 46.7~,0.1 & 48.1~,0.1 & 48.8~,0.1 & 0.8~,0.1 & 0.7~,0.1 & 0.7~,0.1 & 0.7~,0.1\tabularnewline
0.3 & \textbf{32.0} & \textbf{34.2} & \textbf{36.1} & \textbf{36.7} & \textbf{7.0} & \textbf{7.0} & \textbf{6.5} & \textbf{6.4} & \textbf{37.4},~0.6 & \textbf{41.1},~0.6 & \textbf{44.9},~0.5 & \textbf{47.2},~0.5 & \textbf{1.9},~0.5 & \textbf{1.6},~0.5 & \textbf{1.5},~0.5 & \textbf{1.5},~0.5\tabularnewline
\hline
\rowcolor{blue!10}\multicolumn{17}{c}{(CTD) Cosine Distance, Triplet Loss, Defensive (R@1=98.3\%)}\tabularnewline
0 & 50 & 50 & 50 & 50 & 2.0 & 2.0 & 2.0 & 2.0 & 50 & 50 & 50 & 50 & 0.5 & 0.5 & 0.5 & 0.5\tabularnewline
\hline
0.01 & 48.9 & 49.3 & 49.4 & 49.5 & 2.2 & 2.2 & 2.2 & 2.1 & 49.9,~0.0 & 49.5,~0.0 & 49.5,~0.0 & 49.7,~0.0 & 0.5,~0.0 & 0.5,~0.0 & 0.5,~0.0 & 0.5,~0.0\tabularnewline
0.03 & 47.4 & 48.4 & 48.6 & 48.9 & 2.5 & 2.5 & 2.4 & 2.4 & 48.0,~0.0 & 48.5,~0.0 & 49.2,~0.0 & 49.5,~0.0 & 0.6,~0.0 & 0.6,~0.0 & 0.5,~0.0 & 0.5,~0.0\tabularnewline
0.1 & 42.4 & 44.2 & 45.9 & 46.7 & 3.8 & 3.6 & 3.5 & 3.4 & 43.2,~0.1 & 45.0,~0.1 & 47.4,~0.1 & 48.2,~0.1 & 1.0,~0.1 & 0.8,~0.1 & 0.7,~0.1 & 0.7,~0.1\tabularnewline
0.3 & \textbf{30.7} & \textbf{34.5} & \textbf{38.7} & \textbf{40.7} & \textbf{7.0} & \textbf{6.7} & \textbf{6.5} & \textbf{6.5} & \textbf{33.2},~0.5 & \textbf{37.2},~0.5 & \textbf{42.3},~0.5 & \textbf{45.1},~0.5 & \textbf{2.4},~0.4 & \textbf{1.9},~0.4 & \textbf{1.6},~0.4 & \textbf{1.5},~0.4\tabularnewline
\hline
\rowcolor{blue!10}\multicolumn{17}{c}{(ECD) Euclidean Distance, Contrastive Loss, Defensive (R@1=97.9\%)}\tabularnewline
0 & 50 & 50 & 50 & 50 & 1.3 & 1.3 & 1.3 & 1.3 & 50 & 50 & 50 & 50 & 0.5 & 0.5 & 0.5 & 0.5\tabularnewline
\hline
0.01 & 47.4 & 47.6 & 48.0 & 47.8 & 2.3 & 2.2 & 2.1 & 2.1 & 48.7,~0.1 & 49.1,~0.1 & 49.4,~0.1 & 49.7,~0.1 & 0.9,~0.1 & 0.8,~0.1 & 0.7,~0.1 & 0.7,~0.1\tabularnewline
0.03 & 42.7 & 43.6 & 44.0 & 44.2 & 4.5 & 4.2 & 4.0 & 4.0 & 46.3,~0.6 & 48.1,~0.6 & 48.8,~0.6 & 49.2,~0.6 & 1.8,~0.6 & 1.6,~0.6 & 1.5,~0.6 & 1.4,~0.6\tabularnewline
0.1 & 31.7 & 33.7 & 34.9 & 35.3 & 10.0 & 9.6 & 9.4 & 9.3 & 39.2,~2.8 & 43.2,~2.9 & 46.6,~2.9 & 47.9,~2.9 & 3.3,~1.3 & 2.9,~1.3 & 2.7,~1.3 & 2.6,~1.3\tabularnewline
0.3 & \textbf{19.6} & \textbf{23.0} & \textbf{25.4} & \textbf{26.3} & \textbf{35.6} & \textbf{35.2} & \textbf{35.7} & \textbf{36.0} & \textbf{27.3},~7.1 & \textbf{34.4},~7.4 & \textbf{42.2},~7.1 & \textbf{45.4},~6.9 & \textbf{4.5},~1.8 & \textbf{3.9},~1.8 & \textbf{3.6},~1.8 & \textbf{3.4},~1.8\tabularnewline
\hline
\rowcolor{blue!10}\multicolumn{17}{c}{(ETD) Euclidean Distance, Triplet Loss, Defensive (R@1=99.0\%)}\tabularnewline
0 & 50 & 50 & 50 & 50 & 1.4 & 1.4 & 1.4 & 1.4 & 50 & 50 & 50 & 50 & 0.5 & 0.5 & 0.5 & 0.5\tabularnewline
\hline
0.01 & 47.5 & 48.2 & 48.1 & 48.2 & 1.7 & 1.7 & 1.6 & 1.6 & 48.5,~0.0 & 48.8,~0.0 & 49.4,~0.0 & 49.7,~0.0 & 0.6,~0.0 & 0.6,~0.0 & 0.5,~0.0 & 0.5,~0.0\tabularnewline
0.03 & 43.4 & 43.8 & 44.1 & 44.5 & 2.4 & 2.3 & 2.2 & 2.2 & 46.6,~0.0 & 47.8,~0.0 & 48.9,~0.1 & 49.3,~0.1 & 0.8,~0.1 & 0.7,~0.1 & 0.6,~0.1 & 0.6,~0.1\tabularnewline
0.1 & 29.8 & 31.3 & 32.6 & 33.0 & 5.7 & 5.2 & 5.0 & 4.9 & 39.6,~0.3 & 42.9,~0.3 & 45.9,~0.4 & 47.9,~0.4 & 2.1,~0.4 & 1.7,~0.4 & 1.5,~0.4 & 1.4,~0.4\tabularnewline
0.3 & \textbf{10.8} & \textbf{13.2} & \textbf{15.0} & \textbf{15.6} & \textbf{14.4} & \textbf{13.9} & \textbf{13.5} & \textbf{13.4} & \textbf{19.7},~3.2 & \textbf{28.1},~3.7 & \textbf{37.4},~4.1 & \textbf{42.6},~4.3 & \textbf{6.5},~1.7 & \textbf{5.3},~1.7 & \textbf{4.7},~1.8 & \textbf{4.5}~1.8\tabularnewline
\bottomrule
\end{tabular}

}

\caption{Complete Experiment Results on MNIST. The first value for each
QA (\emph{i.e.} SP-QA) experiment result is the average rank of the
chosen candidates, while the other value is the average rank of the
$C_\text{SP}$ used for SP.
}
\label{tab:mnist-complete}

\end{table}

\begin{table}[ht]
	\centering
\resizebox{1.0\columnwidth}{!}{
\setlength{\tabcolsep}{0.2em}

\begin{tabular}{c|cccc|cccc|cccc|cccc}
\toprule
\multirow{2}{*}{$\varepsilon$} & \multicolumn{4}{c|}{CA+} & \multicolumn{4}{c|}{CA-} & \multicolumn{4}{c|}{SP-QA+} & \multicolumn{4}{c}{SP-QA-}\tabularnewline
\cline{2-17} \cline{3-17} \cline{4-17} \cline{5-17} \cline{6-17} \cline{7-17} \cline{8-17} \cline{9-17} \cline{10-17} \cline{11-17} \cline{12-17} \cline{13-17} \cline{14-17} \cline{15-17} \cline{16-17} \cline{17-17}
 & \multicolumn{1}{c|}{$w=1$} & \multicolumn{1}{c|}{$2$} & \multicolumn{1}{c|}{$5$} & $10$ & \multicolumn{1}{c|}{$w=1$} & \multicolumn{1}{c|}{$2$} & \multicolumn{1}{c|}{$5$} & $10$ & \multicolumn{1}{c|}{$m=1$} & \multicolumn{1}{c|}{$2$} & \multicolumn{1}{c|}{$5$} & $10$ & \multicolumn{1}{c|}{$m=1$} & \multicolumn{1}{c|}{$2$} & \multicolumn{1}{c|}{$5$} & $10$\tabularnewline
 \midrule
\rowcolor{black!10}\multicolumn{17}{c}{(CC) Cosine Distance, Contrastive Loss (R@1=88.7\%)}\tabularnewline
0 & 50 & 50 & 50 & 50 & 2.0 & 2.0 & 2.0 & 2.0 & 50 & 50 & 50 & 50 & 0.5 & 0.5 & 0.5 & 0.5\tabularnewline
\hline
0.01 & 29.8 & 32.8 & 34.2 & 34.7 & 9.8 & 9.3 & 9.0 & 8.8 & 36.1,~1.6 & 40.3,~1.6 & 44.8,~1.5 & 46.9,~1.5 & 3.1,~1.4 & 2.8,~1.4 & 2.5,~1.4 & 2.5,~1.4\tabularnewline
0.03 & 12.5 & 16.4 & 19.1 & 20.3 & 46.0 & 45.2 & 44.7 & 44.5 & 18.5,~4.6 & 25.7,~5.6 & 35.6,~5.9 & 41.6,~5.9 & 4.9,~2.4 & 4.4,~2.4 & 4.0,~2.4 & 3.9,~2.5\tabularnewline
0.1 & 4.8 & 10.1 & 13.4 & 14.9 & 96.0 & 96.0 & 96.0 & 96.0 & 10.3,~7.0 & 17.3,~9.7 & 30.0,~11.1 & 38.2,~11.7 & 7.1,~3.7 & 6.4,~3.7 & 5.9,~3.7 & 5.7,~3.7\tabularnewline
0.3 & \textbf{3.7} & \textbf{9.6} & \textbf{12.8} & \textbf{14.2} & \textbf{100.0} & \textbf{100.0} & \textbf{100.0} & \textbf{100.0} & \textbf{9.0},~6.3 & \textbf{15.7},~9.3 & \textbf{28.5},~11.3 & \textbf{37.5},~11.6 & \textbf{7.2},~3.5 & \textbf{6.4},~3.5 & \textbf{5.8},~3.5 & \textbf{5.6},~3.4\tabularnewline
\hline
\rowcolor{black!10}\multicolumn{17}{c}{(CT) Cosine Distance, Triplet Loss (R@1=88.8\%)}\tabularnewline
0 & 50 & 50 & 50 & 50 & 1.9 & 1.9 & 1.9 & 1.9 & 50 & 50 & 50 & 50 & 0.5 & 0.5 & 0.5 & 0.5\tabularnewline
\hline
0.01 & 36.6 & 39.9 & 43.2 & 44.8 & 5.6 & 5.1 & 4.9 & 4.8 & 39.4,~0.2 & 42.0,~0.2 & 45.3,~0.2 & 47.1,~0.2 & 2.1,~0.2 & 1.6,~0.2 & 1.2,~0.2 & 1.1,~0.2\tabularnewline
0.03 & 19.7 & 25.4 & 31.7 & 35.6 & 15.5 & 14.8 & 14.4 & 14.3 & 21.7,~1.5 & 28.2,~1.6 & 35.7,~1.7 & 40.6,~1.7 & 5.6,~0.8 & 4.1,~0.8 & 3.3,~0.8 & 2.9,~0.7\tabularnewline
0.1 & 3.7 & 10.5 & 17.3 & 22.7 & 87.2 & 86.7 & 86.3 & 86.3 & 7.1,~2.4 & 12.4,~4.5 & 23.6,~6.9 & 32.5,~6.8 & 10.9,~1.9 & 8.3,~1.9 & 6.7,~1.9 & 6.0,~1.8\tabularnewline
0.3 & \textbf{1.3} & \textbf{9.4} & \textbf{16.0} & \textbf{21.5} & \textbf{100.0} & \textbf{100.0} & \textbf{100.0} & \textbf{100.0} & \textbf{6.3},~3.0 & \textbf{10.8},~5.2 & \textbf{21.8},~7.8 & \textbf{31.7},~8.3 & \textbf{12.6},~1.9 & \textbf{9.4},~1.9 & \textbf{7.5},~1.9 & \textbf{6.6},~1.8\tabularnewline
\hline
\rowcolor{black!10}\multicolumn{17}{c}{(EC) Euclidean Distance, Contrastive Loss (R@1=87.6\%)}\tabularnewline
0 & 50 & 50 & 50 & 50 & 1.3 & 1.3 & 1.3 & 1.3 & 50 & 50 & 50 & 50 & 0.5 & 0.5 & 0.5 & 0.5\tabularnewline
\hline
0.01 & 20.9 & 23.6 & 25.7 & 26.4 & 12.3 & 11.7 & 11.3 & 11.1 & 30.6,~3.7 & 36.7,~3.8 & 42.8,~3.9 & 45.8,~3.8 & 3.8,~1.6 & 3.3,~1.6 & 3.0,~1.6 & 2.9,~1.6\tabularnewline
0.03 & 7.0 & 11.5 & 14.8 & 16.0 & 74.1 & 72.9 & 71.8 & 71.4 & 15.7,~8.0 & 25.4,~9.3 & 36.9,~9.1 & 42.4,~8.8 & 4.6,~2.0 & 4.0,~2.0 & 3.6,~2.0 & 3.6,~2.1\tabularnewline
0.1 & 6.2 & 12.2 & 17.0 & 18.7 & 100.0 & 100.0 & 100.0 & 100.0 & 14.2,~9.3 & 22.6,~12.3 & 34.8,~13.2 & 41.2,~13.0 & 8.3,~4.7 & 7.8,~4.8 & 7.3,~4.8 & 7.1,~4.8\tabularnewline
0.3 & \textbf{5.7} & \textbf{12.0} & \textbf{16.5} & \textbf{18.2} & \textbf{100.0} & \textbf{100.0} & \textbf{100.0} & \textbf{100.0} & \textbf{12.8},~9.0 & \textbf{21.1},~11.9 & \textbf{33.8},~13.5 & \textbf{40.4},~13.4 & \textbf{9.1},~5.1 & \textbf{8.4},~5.2 & \textbf{7.9},~5.0 & \textbf{7.5},~4.9\tabularnewline
\hline
\rowcolor{black!10}\multicolumn{17}{c}{(ET) Euclidean Distance, Triplet Loss (R@1=88.3\%)}\tabularnewline
0 & 50 & 50 & 50 & 50 & 1.5 & 1.5 & 1.5 & 1.5 & 50 & 50 & 50 & 50 & 0.5 & 0.5 & 0.5 & 0.5\tabularnewline
\hline
0.01 & 33.2 & 34.8 & 34.7 & 36.1 & 6.5 & 5.9 & 5.5 & 5.3 & 41.7,~0.3 & 44.4,~0.3 & 47.3,~0.3 & 48.4,~0.3 & 2.5,~0.3 & 1.9,~0.3 & 1.6,~0.3 & 1.4,~0.3\tabularnewline
0.03 & 14.0 & 17.6 & 20.3 & 21.3 & 18.4 & 17.2 & 16.4 & 16.0 & 22.2,~2.5 & 30.0,~2.7 & 39.4,~2.7 & 43.6,~2.6 & 6.6,~1.3 & 5.1,~1.4 & 4.2,~1.4 & 3.9,~1.4\tabularnewline
0.1 & 1.7 & 9.5 & 14.0 & 15.6 & 88.3 & 86.7 & 85.2 & 84.5 & 8.0,~3.9 & 15.0,~6.3 & 27.6,~7.9 & 36.5,~8.1 & 10.5,~1.9 & 8.0,~1.9 & 6.4,~2.0 & 6.0,~2.0\tabularnewline
0.3 & \textbf{0.3} & \textbf{9.0} & \textbf{13.8} & \textbf{15.5} & \textbf{100.0} & \textbf{100.0} & \textbf{100.0} & \textbf{100.0} & \textbf{6.7},~3.3 & \textbf{12.6},~5.7 & \textbf{25.4},~7.9 & \textbf{34.8},~8.3 & \textbf{11.9},~1.9 & \textbf{8.9},~1.9 & \textbf{7.1},~1.9 & \textbf{6.4},~1.9\tabularnewline
\midrule
\rowcolor{blue!10}\multicolumn{17}{c}{(CCD) Cosine Distance, Contrastive Loss, Defensive (R@1=82.2\%)}\tabularnewline
0 & 50 & 50 & 50 & 50 & 2.0 & 2.0 & 2.0 & 2.0 & 50 & 50 & 50 & 50 & 0.5 & 0.5 & 0.5 & 0.5\tabularnewline
\hline
0.01 & 47.0 & 47.7 & 47.8 & 47.9 & 2.3 & 2.3 & 2.3 & 2.2 & 48.9,~0.0 & 48.7,~0.0 & 49.2,~0.0 & 49.6,~0.0 & 0.6,~0.0 & 0.5,~0.0 & 0.5,~0.0 & 0.5,~0.0\tabularnewline
0.03 & 42.2 & 43.3 & 44.1 & 44.3 & 3.1 & 3.0 & 2.9 & 2.8 & 45.2,~0.0 & 47.2,~0.0 & 48.0,~0.0 & 49.1,~0.0 & 0.8,~0.0 & 0.7,~0.0 & 0.6,~0.0 & 0.6,~0.0\tabularnewline
0.1 & 29.1 & 31.4 & 32.9 & 33.8 & 8.0 & 7.1 & 6.4 & 6.2 & 34.7,~0.1 & 39.5,~0.1 & 43.8,~0.1 & 46.5,~0.1 & 3.3,~0.1 & 2.1,~0.1 & 1.5,~0.1 & 1.3,~0.1\tabularnewline
0.3 & \textbf{11.8} & \textbf{14.8} & \textbf{17.4} & \textbf{18.4} & \textbf{28.7} & \textbf{25.5} & \textbf{23.1} & \textbf{22.4} & \textbf{13.3},~0.7 & \textbf{20.0},~0.9 & \textbf{31.0},~1.0 & \textbf{38.3},~1.0 & \textbf{21.3},~1.0 & \textbf{14.3},~1.0 & \textbf{10.3},~1.0 & \textbf{8.6},~1.1\tabularnewline
\hline
\rowcolor{blue!10}\multicolumn{17}{c}{(CTD) Cosine Distance, Triplet Loss, Defensive (R@1=79.6\%)}\tabularnewline
0 & 50 & 50 & 50 & 50 & 1.2 & 1.2 & 1.2 & 1.2 & 50 & 50 & 50 & 50 & 0.5 & 0.5 & 0.5 & 0.5\tabularnewline
\hline
0.01 & 48.9 & 48.9 & 49.3 & 49.3 & 1.4 & 1.4 & 1.4 & 1.4 & 49.4,~0.0 & 49.9,~0.0 & 49.9,~0.0 & 50.0,~0.0 & 0.5,~0.0 & 0.5,~0.0 & 0.5,~0.0 & 0.5,~0.0\tabularnewline
0.03 & 47.1 & 47.9 & 48.3 & 48.3 & 2.0 & 1.9 & 1.8 & 1.8 & 48.3,~0.0 & 49.1,~0.0 & 49.5,~0.0 & 49.8,~0.0 & 0.7,~0.0 & 0.6,~0.0 & 0.6,~0.0 & 0.6,~0.0\tabularnewline
0.1 & 42.4 & 43.5 & 44.5 & 44.8 & 4.6 & 4.2 & 4.0 & 3.9 & 45.4,~0.3 & 47.2,~0.2 & 48.7,~0.2 & 49.2,~0.2 & 1.4,~0.2 & 1.2,~0.2 & 1.1,~0.2 & 1.0,~0.2\tabularnewline
0.3 & \textbf{32.5} & \textbf{35.4} & \textbf{37.5} & \textbf{38.2} & \textbf{11.2} & \textbf{10.5} & \textbf{10.1} & \textbf{10.0} & \textbf{39.3},~1.5 & \textbf{42.6},~1.5 & \textbf{46.5},~1.3 & \textbf{47.8},~1.3 & \textbf{3.9},~1.4 & \textbf{3.3},~1.4 & \textbf{3.0},~1.4 & \textbf{2.9},~1.4\tabularnewline
\hline
\rowcolor{blue!10}\multicolumn{17}{c}{(ECD) Euclidean Distance, Contrastive Loss, Defensive (R@1=80.4\%)}\tabularnewline
0 & 50 & 50 & 50 & 50 & 1.4 & 1.4 & 1.4 & 1.4 & 50 & 50 & 50 & 50 & 0.5 & 0.5 & 0.5 & 0.5\tabularnewline
\hline
0.01 & 45.3 & 45.8 & 46.3 & 46.6 & 3.0 & 2.8 & 2.7 & 2.6 & 48.0,~0.2 & 48.7,~0.2 & 49.3,~0.3 & 49.7,~0.3 & 1.2,~0.2 & 1.0,~0.2 & 0.9,~0.2 & 0.9,~0.2\tabularnewline
0.03 & 37.6 & 39.5 & 40.4 & 40.7 & 7.5 & 7.0 & 6.8 & 6.7 & 43.9,~1.5 & 46.4,~1.6 & 48.4,~1.7 & 49.1,~1.9 & 2.7,~0.7 & 2.2,~0.7 & 1.9,~0.7 & 1.9,~0.7\tabularnewline
0.1 & 20.3 & 24.9 & 27.8 & 28.6 & 32.6 & 32.1 & 32.0 & 32.5 & 28.4,~6.6 & 36.2,~7.2 & 43.0,~7.4 & 45.9,~7.4 & 5.3,~1.5 & 4.4,~1.6 & 3.8,~1.6 & 3.7,~1.6\tabularnewline
0.3 & \textbf{7.3} & \textbf{16.0} & \textbf{21.1} & \textbf{22.5} & \textbf{78.0} & \textbf{79.0} & \textbf{80.3} & \textbf{81.2} & \textbf{14.0},~10.2 & \textbf{24.6},~11.9 & \textbf{36.5},~11.8 & \textbf{42.2},~11.2 & \textbf{7.0},~1.9 & \textbf{5.7},~1.9 & \textbf{4.9},~1.9 & \textbf{4.6},~1.9\tabularnewline
\hline
\rowcolor{blue!10}\multicolumn{17}{c}{(ETD) Euclidean Distance, Triplet Loss, Defensive (R@1=84.4\%)}\tabularnewline
0 & 50 & 50 & 50 & 50 & 1.3 & 1.3 & 1.3 & 1.3 & 50 & 50 & 50 & 50 & 0.5 & 0.5 & 0.5 & 0.5\tabularnewline
\hline
0.01 & 46.0 & 46.7 & 46.9 & 46.7 & 2.0 & 1.9 & 1.8 & 1.8 & 48.1,~0.0 & 49.1,~0.0 & 49.3,~0.0 & 49.6,~0.0 & 0.8,~0.0 & 0.7,~0.0 & 0.6,~0.0 & 0.6,~0.1\tabularnewline
0.03 & 38.9 & 40.6 & 41.2 & 41.4 & 4.3 & 3.8 & 3.5 & 3.4 & 45.2,~0.2 & 46.2,~0.1 & 48.1,~0.1 & 48.9,~0.1 & 1.9,~0.2 & 1.5,~0.2 & 1.2,~0.2 & 1.1,~0.2\tabularnewline
0.1 & 23.0 & 26.3 & 28.0 & 28.8 & 15.9 & 14.9 & 14.2 & 13.9 & 31.8,~1.8 & 38.0,~1.8 & 43.3,~1.6 & 46.2,~1.6 & 5.6,~1.0 & 4.3,~1.0 & 3.6,~1.0 & 3.3,~1.0\tabularnewline
0.3 & \textbf{7.1} & \textbf{13.9} & \textbf{18.4} & \textbf{19.9} & \textbf{54.1} & \textbf{51.8} & \textbf{50.0} & \textbf{49.2} & \textbf{10.6},~5.6 & \textbf{19.7},~7.1 & \textbf{32.5},~8.0 & \textbf{39.6},~7.6 & \textbf{9.2},~1.6 & \textbf{7.0},~1.6 & \textbf{5.6},~1.6 & \textbf{5.1},~1.6\tabularnewline
\bottomrule
\end{tabular}

}
\caption{Complete Experiments on Fashion-MNIST Dataset. Each QA result comprises
the average rank of the chosen candidates and the $C_\text{SP}$ used for SP,
respectively.}
\label{tab:fashion-complete}
\end{table}

\newpage\clearpage
\section{Complete Results on Transferability}

\subsection{Fashion-MNIST}

In addition to the transferability experiment on MNIST dataset, we also conduct the same transferability experiment on the
Fashion-MNIST dataset, as shown in
Tab.~\ref{tab:transfer-fashion}.

\begin{table}[ht]
\begin{center}
\resizebox{0.6\columnwidth}{!}{
\setlength{\tabcolsep}{0.2em}%
\begin{tabular}{c|c|c|c}

\toprule

\rowcolor{black!10}\multicolumn{4}{c}{CA+ Transfer (Black Box), $w=1$}\tabularnewline
\backslashbox{From}{To} & LeNet & C2F1 & Res18\tabularnewline
\hline
LeNet & \cellcolor{black!10}50$\rightarrow$16.0 &41.0                                   &44.3 \tabularnewline
C2F1  &  38.6                                   &\cellcolor{black!10}50$\rightarrow$1.3 &40.3 \tabularnewline
Res18 & 39.2                                    &34.3                                   &\cellcolor{black!10}50$\rightarrow$1.7 \tabularnewline
\midrule

\rowcolor{black!10}\multicolumn{4}{c}{CA- Transfer (Black Box), $w=1$}\tabularnewline
\backslashbox{From}{To} & LeNet & C2F1 & Res18\tabularnewline
\hline
LeNet &  \cellcolor{black!10}2.5$\rightarrow$84.3 & 1.9$\rightarrow$8.1                       & 1.6$\rightarrow$6.0 \tabularnewline
C2F1  &   2.5$\rightarrow$7.8                     & \cellcolor{black!10}1.9$\rightarrow$100.0 & 1.7$\rightarrow$7.7\tabularnewline
Res18 &   2.5$\rightarrow$9.5                     & 1.9$\rightarrow$14.4                      & \cellcolor{black!10}1.7$\rightarrow$80.0 \tabularnewline
\midrule

\rowcolor{black!10}\multicolumn{4}{c}{SP-QA+ Transfer (Black Box), $m=1$}\tabularnewline
\backslashbox{From}{To} & LeNet & C2F1 & Res18\tabularnewline
\hline
LeNet & \cellcolor{black!10}50$\rightarrow$18.0 &47.2 &49.3 \tabularnewline
C2F1 &  48.1 &\cellcolor{black!10}50$\rightarrow$6.4 &49.2 \tabularnewline
Res18 & 48.1 &44.8 &\cellcolor{black!10}50$\rightarrow$13.7 \tabularnewline
\midrule

\rowcolor{black!10}\multicolumn{4}{c}{SP-QA- Transfer (Black Box), $m=1$}\tabularnewline
\backslashbox{From}{To} & LeNet & C2F1 & Res18\tabularnewline
\hline
LeNet & \cellcolor{black!10}0.5$\rightarrow$13.5 & 0.5$\rightarrow$1.7 & 0.5$\rightarrow$1.5 \tabularnewline
C2F1 &  0.5$\rightarrow$1.1 & \cellcolor{black!10}0.5$\rightarrow$12.5 & 0.5$\rightarrow$1.6  \tabularnewline
Res18 & 0.5$\rightarrow$0.9 & 0.5$\rightarrow$1.3 & \cellcolor{black!10}0.5$\rightarrow$8.0  \tabularnewline
\bottomrule

\end{tabular}}
\end{center}

\caption{Transferability experiment on Fashion-MNIST dataset.
}
\label{tab:transfer-fashion}
\end{table}

\subsection{``Self-Transfer'' Attack on MNIST}

\begin{table}[ht]
\centering
\resizebox{1.0\columnwidth}{!}{
\setlength{\tabcolsep}{0.2em}%
\begin{tabular}{c|c|c|c|c|c}
\toprule
\rowcolor{black!10}\multicolumn{6}{c}{CA+ Transfer (Black Box), $w=1$}\tabularnewline
\backslashbox{From}{To} & C2F1-1 & C2F1-2 & C2F1-3 & C2F1-D1 & C2F1-D2\tabularnewline
\hline
C2F1-1 & \cellcolor{black!10}50$\rightarrow$2.1 &27.9 &25.3 &45.3 &45.1\tabularnewline
\hline
C2F1-2 & 24.4 &\cellcolor{black!10}50$\rightarrow$2.7 &23.7 &44.5 &44.0\tabularnewline
\hline
C2F1-3 & 22.6 &24.4 &\cellcolor{black!10}50$\rightarrow$2.2 &45.0 &44.5\tabularnewline
\hline
C2F1-D1 & 36.7 &37.7 &37.4 &\cellcolor{black!10}50$\rightarrow$30.9 &38.6\tabularnewline
\hline
C2F1-D2 & 37.3 &36.8 &37.3 &39.6 &\cellcolor{black!10}50$\rightarrow$30.3\tabularnewline
\midrule
\rowcolor{black!10}\multicolumn{6}{c}{CA- Transfer (Black Box), $w=1$}\tabularnewline
\backslashbox{From}{To} & C2F1-1 & C2F1-2 & C2F1-3 & C2F1-D1 & C2F1-D2\tabularnewline
\hline
C2F1-1 & \cellcolor{black!10}2.1$\rightarrow$93.7 & 2.2$\rightarrow$9.4 & 2.1$\rightarrow$11.1 & 2.0$\rightarrow$3.2 & 1.9$\rightarrow$3.1\tabularnewline
\hline
C2F1-2 & 2.1$\rightarrow$13.5 & \cellcolor{black!10}2.1$\rightarrow$88.1 & 2.2$\rightarrow$12.3 & 2.0$\rightarrow$3.4 & 1.9$\rightarrow$3.3\tabularnewline
\hline
C2F1-3 & 2.1$\rightarrow$13.4 & 2.2$\rightarrow$10.1 & \cellcolor{black!10}2.2$\rightarrow$92.4 & 2.0$\rightarrow$3.2 & 1.9$\rightarrow$3.2\tabularnewline
\hline
C2F1-D1 & 2.1$\rightarrow$9.3 & 2.2$\rightarrow$8.3 & 2.1$\rightarrow$9.1 & \cellcolor{black!10}2.0$\rightarrow$7.0 & 2.0$\rightarrow$4.9\tabularnewline
\hline
C2F1-D2 & 2.1$\rightarrow$9.2 & 2.2$\rightarrow$8.4 & 2.1$\rightarrow$9.1 & 2.0$\rightarrow$4.9 & \cellcolor{black!10}1.9$\rightarrow$7.0\tabularnewline
\midrule
\rowcolor{black!10}\multicolumn{6}{c}{SP-QA+ Transfer (Black Box), $m=1$}\tabularnewline
\backslashbox{From}{To} & C2F1-1 & C2F1-2 & C2F1-3 & C2F1-D1 & C2F1-D2\tabularnewline
\hline
C2F1-1 & \cellcolor{black!10}50$\rightarrow$6.4 &39.4 &38.1 &48.6 &48.1\tabularnewline
\hline
C2F1-2 & 36.6 &\cellcolor{black!10}50$\rightarrow$6.3 &36.4 &47.3 &48.4\tabularnewline
\hline
C2F1-3 & 36.9 &37.8 &\cellcolor{black!10}50$\rightarrow$6.2 &48.7 &48.3\tabularnewline
\hline
C2F1-D1 & 41.5 &40.3 &41.4 &\cellcolor{black!10}50$\rightarrow$32.9 &40.6\tabularnewline
\hline
C2F1-D2 & 41.7 &38.9 &41.7 &41.4 &\cellcolor{black!10}50$\rightarrow$32.7\tabularnewline
\midrule
\rowcolor{black!10}\multicolumn{6}{c}{SP-QA- Transfer (Black Box), $m=1$}\tabularnewline
\backslashbox{From}{To} & C2F1-1 & C2F1-2 & C2F1-3 & C2F1-D1 & C2F1-D2\tabularnewline
\hline
C2F1-1 & \cellcolor{black!10}0.5$\rightarrow$8.8 & 0.5$\rightarrow$1.4 & 0.5$\rightarrow$1.9 & 0.5$\rightarrow$0.6 & 0.5$\rightarrow$0.6\tabularnewline
\hline
C2F1-2 & 0.5$\rightarrow$2.3 & \cellcolor{black!10}0.5$\rightarrow$9.0 & 0.5$\rightarrow$2.2 & 0.5$\rightarrow$0.6 & 0.5$\rightarrow$0.6\tabularnewline
\hline
C2F1-3 & 0.5$\rightarrow$1.9 & 0.5$\rightarrow$1.5 & \cellcolor{black!10}0.5$\rightarrow$8.7 & 0.5$\rightarrow$0.6 & 0.5$\rightarrow$0.6\tabularnewline
\hline
C2F1-D1 & 0.5$\rightarrow$5.8 & 0.5$\rightarrow$3.1 & 0.5$\rightarrow$4.8 & \cellcolor{black!10}0.5$\rightarrow$2.4 & 0.5$\rightarrow$1.3\tabularnewline
\hline
C2F1-D2 & 0.5$\rightarrow$7.1 & 0.5$\rightarrow$3.7 & 0.5$\rightarrow$5.9 & 0.5$\rightarrow$1.3 & \cellcolor{black!10}0.5$\rightarrow$2.4\tabularnewline
\bottomrule
\end{tabular}}

\caption{Transfer experiment between models based on the same architecture but with different parameters.}
\label{tab:transfer-self}
\end{table}

In the adversarial ranking example transferability experiments, we transfer
adversarial examples between neural networks with different architectures.
In this section, we transfer adversarial examples between neural networks
based on the same architecture but with different parameters.

We train three vanilla C2F1 models (denoted as C2F1-1, C2F1-2, and C2f1-3) on MNIST dataset and two
models with our defense (denoted as C2F1-D1, C2F1-D2). All these models have 
exactly the same architecture, but different parameters. 
Transferability experimental results between these models are present in Tab.~\ref{tab:transfer-self}.


\section{Complete Results for Universal Perturbation}

\subsection{MNIST}

See Tab.~\ref{tab:iap-mnist}.

\begin{table}[t]
\centering
\resizebox{1.0\columnwidth}{!}{
\setlength{\tabcolsep}{0.2em}
\begin{tabular}{c|c|c|c|c|c|c|c|c}
\toprule
 	\multirow{2}{*}{Model} & \multicolumn{2}{c|}{I-CA+ ($w=1$)} & \multicolumn{2}{c|}{I-CA- ($w=1$)} & \multicolumn{2}{c|}{I-QA+ ($m=1$)} & \multicolumn{2}{c}{I-QA- ($m=1$)}\tabularnewline
\cline{2-9} \cline{3-9} \cline{4-9} \cline{5-9} \cline{6-9} \cline{7-9} \cline{8-9} \cline{9-9}
	& Seen & Unseen & Seen & Unseen & Seen & Unseen & Seen & Unseen \tabularnewline
\midrule
(CC) & 50 $\rightarrow$ 13.7 & 50 $\rightarrow$ 14.1 & 0.6 $\rightarrow$ 37.6 & 0.6 $\rightarrow$ 33.4 & 50 $\rightarrow$ 18.1 & 50 $\rightarrow$ 18.9 & 2.4 $\rightarrow$ 41.6 & 2.4 $\rightarrow$ 39.3\tabularnewline
\rowcolor{black!10}(CT) & 50 $\rightarrow$ 18.1 & 50 $\rightarrow$ 18.5 & 0.6 $\rightarrow$ 9.5 & 0.7 $\rightarrow$ 9.4 & 50 $\rightarrow$ 20.5 & 50 $\rightarrow$ 21.0 & 2.1 $\rightarrow$ 7.6 & 2.2 $\rightarrow$ 7.4\tabularnewline
(EC) & 50 $\rightarrow$ 9.1 & 50 $\rightarrow$ 10.3 & 1.9 $\rightarrow$ 94.6 & 2.0 $\rightarrow$ 79.4 & 50 $\rightarrow$ 3.9 & 50 $\rightarrow$ 6.9 & 3.3 $\rightarrow$ 87.1 & 3.5 $\rightarrow$ 74.4 \tabularnewline
\rowcolor{black!10}(ET) & 50 $\rightarrow$ 11.5 & 50 $\rightarrow$ 12.6 & 2.1 $\rightarrow$ 10.6 & 2.1 $\rightarrow$ 9.6 & 50 $\rightarrow$ 21.6 & 50 $\rightarrow$ 23.6 & 3.2 $\rightarrow$ 28.6 & 3.2 $\rightarrow$ 19.5\tabularnewline
\bottomrule
\end{tabular}}
\caption{Universal Adversarial Ranking Perturbation on MNIST. Each pair of result
	presents the original rank of chosen candidate(s), and the rank of the chosen candidate(s)
	after adding adversarial perturbation to the candidate or to the query.
	``Seen'' samples are those used for generating the universal perturbation,
	while ``Unseen'' samples are a set of other non-overlapping samples.
   }
\label{tab:iap-mnist}
\end{table}

\subsection{Fashion-MNIST}

As shown in Tab.~\ref{tab:iap-fashion},
Image-agnostic adversarial perturbation has better effect on seen data from
Fashion-MNIST, but the gap between the effect on seen samples and that on
unseen samples is slightly larger, which may due to the higher
intra-class variance of Fashion-MNIST than MNIST.

\begin{table}[t]
\centering
\resizebox{1.0\columnwidth}{!}{
\setlength{\tabcolsep}{0.2em}
\begin{tabular}{c|c|c|c|c|c|c|c|c}
\toprule
	\multirow{2}{*}{Model} & \multicolumn{2}{c|}{I-CA+ ($w=1$)} & \multicolumn{2}{c|}{I-CA- ($w=1$)} & \multicolumn{2}{c|}{I-QA+ ($m=1$)} & \multicolumn{2}{c}{I-QA- ($m=1$)}\tabularnewline
\cline{2-9} \cline{3-9} \cline{4-9} \cline{5-9} \cline{6-9} \cline{7-9} \cline{8-9} \cline{9-9}
 & Seen & Unseen & Seen & Unseen & Seen & Unseen & Seen & Unseen \tabularnewline
\midrule
(CC) & 50 $\rightarrow$ 7.0 & 50 $\rightarrow$ 7.3 & 0.6 $\rightarrow$ 91.8 & 0.6 $\rightarrow$ 84.9 & 50 $\rightarrow$ 4.4 & 50 $\rightarrow$ 4.9 & 2.1 $\rightarrow$ 87.5 & 2.1 $\rightarrow$ 84.4\tabularnewline
\rowcolor{black!10}(CT) & 50 $\rightarrow$ 9.8 & 50 $\rightarrow$ 9.9 & 0.6 $\rightarrow$ 72.3 & 0.6 $\rightarrow$ 69.7 & 50 $\rightarrow$ 8.2 & 50 $\rightarrow$ 8.4 & 2.0 $\rightarrow$ 76.3 & 2.0 $\rightarrow$ 72.9\tabularnewline
(EC) & 50 $\rightarrow$ 5.8 & 50 $\rightarrow$ 9.6 & 2.0 $\rightarrow$ 97.5 & 1.9 $\rightarrow$ 83.7 & 50 $\rightarrow$ 1.8 & 50 $\rightarrow$ 7.1 & 2.9 $\rightarrow$ 87.9 & 2.8 $\rightarrow$ 78.4\tabularnewline
\rowcolor{black!10}(ET) & 50 $\rightarrow$ 5.7 & 50 $\rightarrow$ 8.5 & 2.0 $\rightarrow$ 84.4 & 1.9 $\rightarrow$ 69.9 & 50 $\rightarrow$ 3.3 & 50 $\rightarrow$ 6.3 & 3.1 $\rightarrow$ 88.0 & 3.0 $\rightarrow$ 78.0\tabularnewline
\bottomrule
\end{tabular}}
\caption{Image-agnostic Adversarial Perturbation on Fashion-MNIST.}
\label{tab:iap-fashion}
\end{table}

\newpage\clearpage
\section{Complete Parameter Search on $\xi$}

See Tab.~\ref{tab:xi-ablation}.

\begin{table}[t]
\centering
\resizebox{1.0\columnwidth}{!}{
\setlength{\tabcolsep}{0.2em}%
\begin{tabular}{c|cccc|cccc}
\toprule
\multirow{2}{*}{$\xi$} & \multicolumn{4}{c|}{SP-QA+} & \multicolumn{4}{c}{SP-QA-}\tabularnewline
\cline{2-9} \cline{3-9} \cline{4-9} \cline{5-9} \cline{6-9} \cline{7-9} \cline{8-9} \cline{9-9}
 & \multicolumn{1}{c|}{m=1} & \multicolumn{1}{c|}{2} & \multicolumn{1}{c|}{5} & 10 & \multicolumn{1}{c|}{m=1} & \multicolumn{1}{c|}{2} & \multicolumn{1}{c|}{5} & 10\tabularnewline
 \midrule

\rowcolor{black!10}\multicolumn{9}{c}{(CC) Cosine, Contrastive}\tabularnewline
0 & 1.4,~43.9 & 13.8,~38.1 & 30.7,~37.9 & 39.4,~36.3 & 94.7,~93.3 & 94.4,~93.5 & 94.2,~93.7 & 94.1,~93.8\tabularnewline
$10^0$ & 14.3,~9.7 & 23.3,~12.5 & 36.3,~13.8 & 42.6,~13.9 & 40.6,~37.6 & 40.6,~38.1 & 40.4,~38.1 & 39.8,~37.7\tabularnewline
$10^2$ & 25.4,~2.0 & 33.2,~2.1 & 41.7,~2.1 & 45.4,~2.1 & 6.1,~4.4 & 5.8,~4.4 & 5.7,~4.4 & 5.7,~4.4\tabularnewline
$10^4$ & 48.1,~0.3 & 49.2,~0.2 & 49.6,~0.2 & 49.8,~0.3 & 1.2,~0.3 & 1.1,~0.3 & 1.1,~0.3 & 1.1,~0.3\tabularnewline
\midrule
\rowcolor{black!10}\multicolumn{9}{c}{(CT) Cosine, Triplet}\tabularnewline
0 & 0.2,~33.6 & 6.3,~23.7 & 18.5,~26.5 & 29.6,~25.7 & 94.1,~89.4 & 93.2,~90.3 & 92.6,~90.9 & 92.3,~91.2\tabularnewline
$10^0$ & 6.3,~3.6 & 11.2,~5.7 & 22.5,~7.7 & 32.1,~7.7 & 55.5,~35.6 & 52.4,~37.6 & 50.2,~39.3 & 49.4,~40.0\tabularnewline
$10^2$ & 14.1,~0.6 & 20.8,~0.7 & 31.2,~0.7 & 38.1,~0.7 & 8.6,~1.6 & 6.6,~1.6 & 5.3,~1.5 & 4.8,~1.5\tabularnewline
$10^4$ & 37.9,~0.1 & 42.6,~0.1 & 46.3,~0.1 & 47.8,~0.1 & 1.9,~0.1 & 1.4,~0.1 & 1.2,~0.1 & 1.1,~0.1\tabularnewline
\midrule
\rowcolor{black!10}\multicolumn{9}{c}{(EC) Euclidean, Contrastive}\tabularnewline
0 & 0.7,~44.5 & 13.4,~39.9 & 31.0,~41.8 & 39.7,~41.6 & 94.0,~92.6 & 93.9,~93.0 & 93.9,~93.4 & 93.9,~93.5\tabularnewline
$10^0$ & 14.4,~9.4 & 23.9,~12.5 & 36.7,~13.6 & 42.6,~13.8 & 39.9,~37.6 & 39.0,~37.2 & 39.2,~37.7 & 38.7,~37.3\tabularnewline
$10^2$ & 30.1,~1.2 & 37.3,~1.1 & 43.9,~0.9 & 46.7,~0.9 & 5.9,~3.2 & 5.4,~3.2 & 5.0,~3.2 & 4.9,~3.1\tabularnewline
$10^4$ & 50.1,~0.8 & 50.2,~0.8 & 49.8,~0.8 & 50.0,~0.8 & 1.6,~0.7 & 1.6,~0.7 & 1.7,~0.8 & 1.6,~0.8\tabularnewline
\midrule
\rowcolor{black!10}\multicolumn{9}{c}{(ET) Euclidean, Triplet}\tabularnewline
0 & 0.6,~37.4 & 8.2,~35.1 & 22.6,~38.2 & 33.2,~38.8 & 93.0,~88.4 & 92.5,~89.9 & 92.3,~90.9 & 92.1,~90.9\tabularnewline
$10^0$ & 7.5,~5.2 & 15.0,~7.4 & 28.1,~9.0 & 36.3,~8.9 & 50.9,~37.3 & 49.2,~39.4 & 47.7,~40.6 & 47.2,~40.9\tabularnewline
$10^2$ & 20.3,~0.5 & 28.6,~0.5 & 38.2,~0.5 & 43.1,~0.4 & 7.4,~1.7 & 5.8,~1.7 & 4.9,~1.7 & 4.6,~1.7\tabularnewline
$10^4$ & 46.2,~0.1 & 48.2,~0.1 & 49.1,~0.1 & 49.6,~0.1 & 1.7,~0.1 & 1.4,~0.1 & 1.2,~0.1 & 1.1,~0.1\tabularnewline
\bottomrule
\end{tabular}
	}

\caption{Parameter search on Semantics-Preserving balancing parameter $\xi$ with MNIST.
	We report two mean ranks in each cell: one for the chosen candidate(s) $C$, the other
	for the $C_\text{SP}$ used for SP.}
\label{tab:xi-ablation}
\end{table}

%

\newpage\clearpage
\section{Analysis on Our Defense}

\begin{figure}[h]
\centering
\includegraphics[width=0.9\columnwidth]{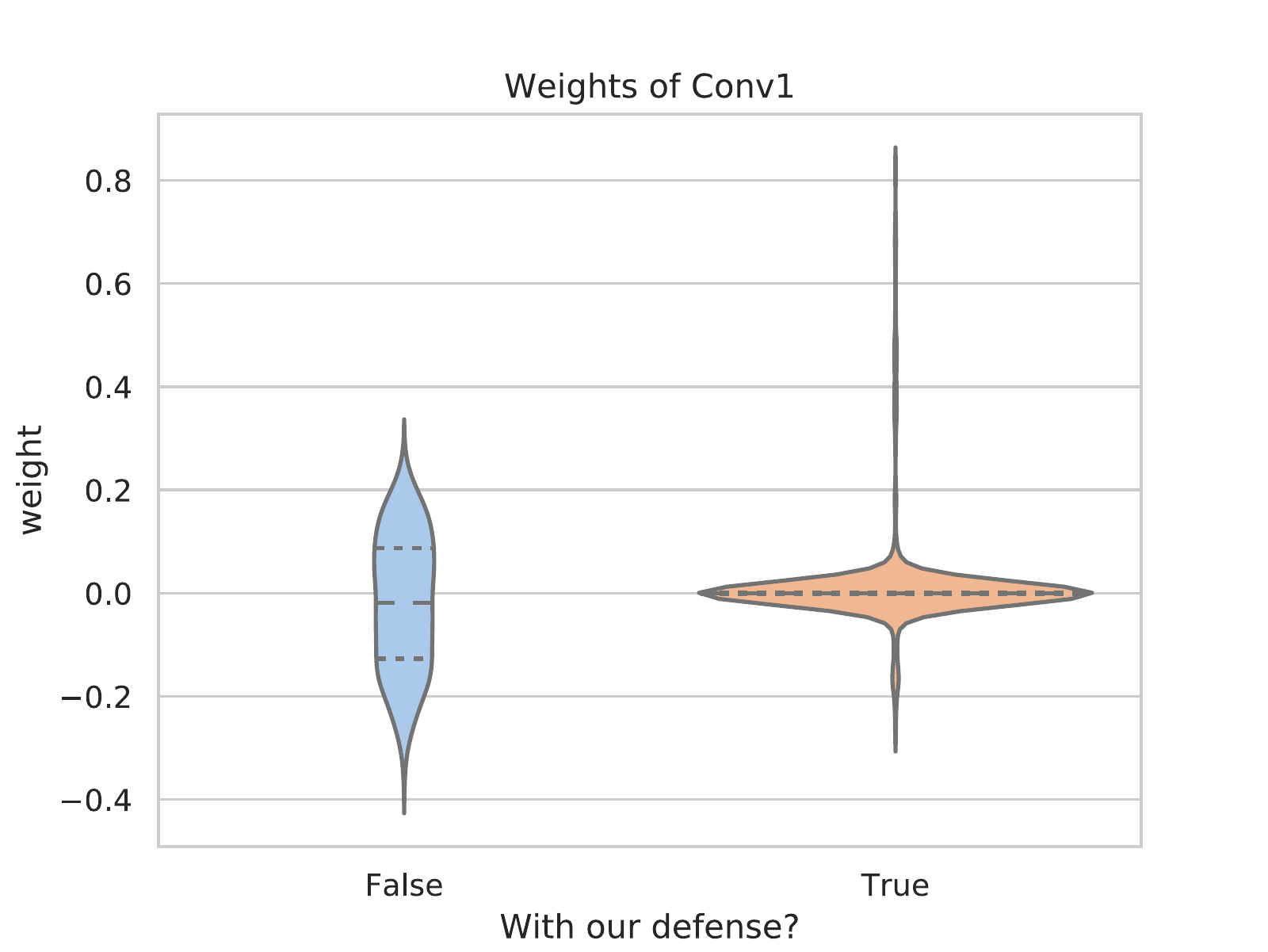}
\includegraphics[width=0.9\columnwidth]{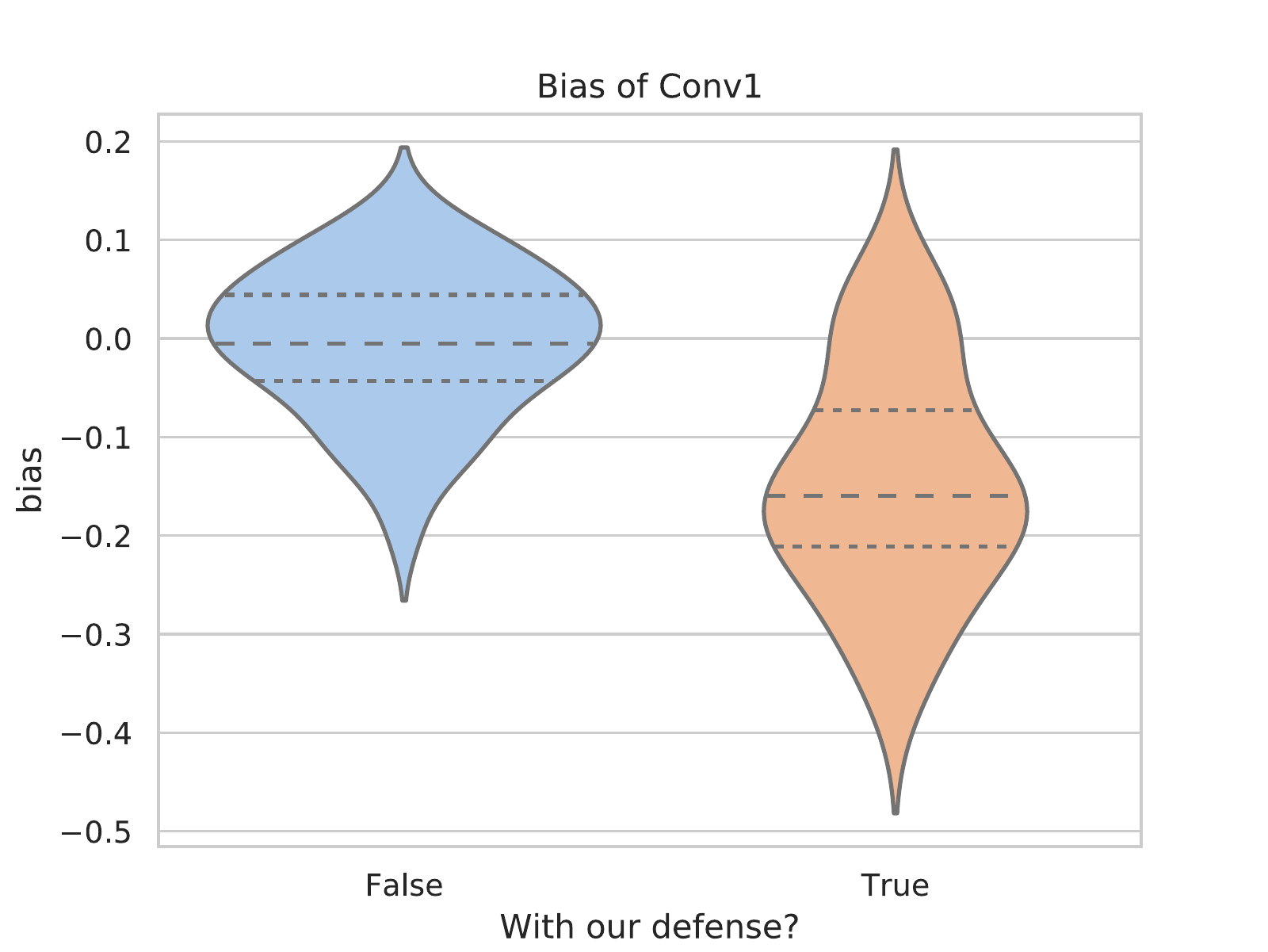}
\caption{Weights and bias of the first convolutional layer (conv1)
	with/without our defense.}
\label{fig:conv1}
\end{figure}

\begin{figure}[h]
\centering
	\minipage{0.49\columnwidth}
	\includegraphics[width=\linewidth]{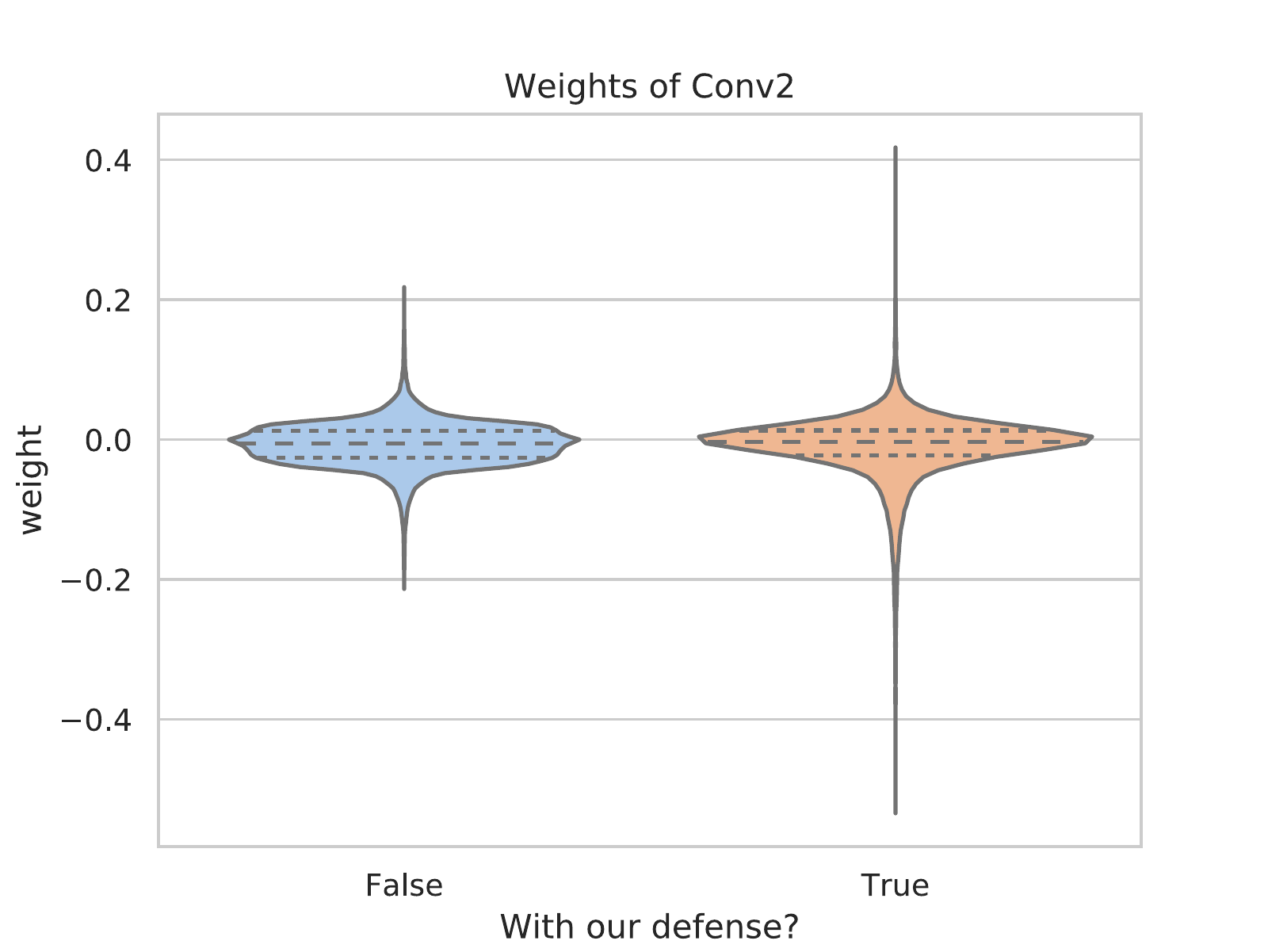}
	\endminipage\hfill
	\minipage{0.49\columnwidth}
	\includegraphics[width=\linewidth]{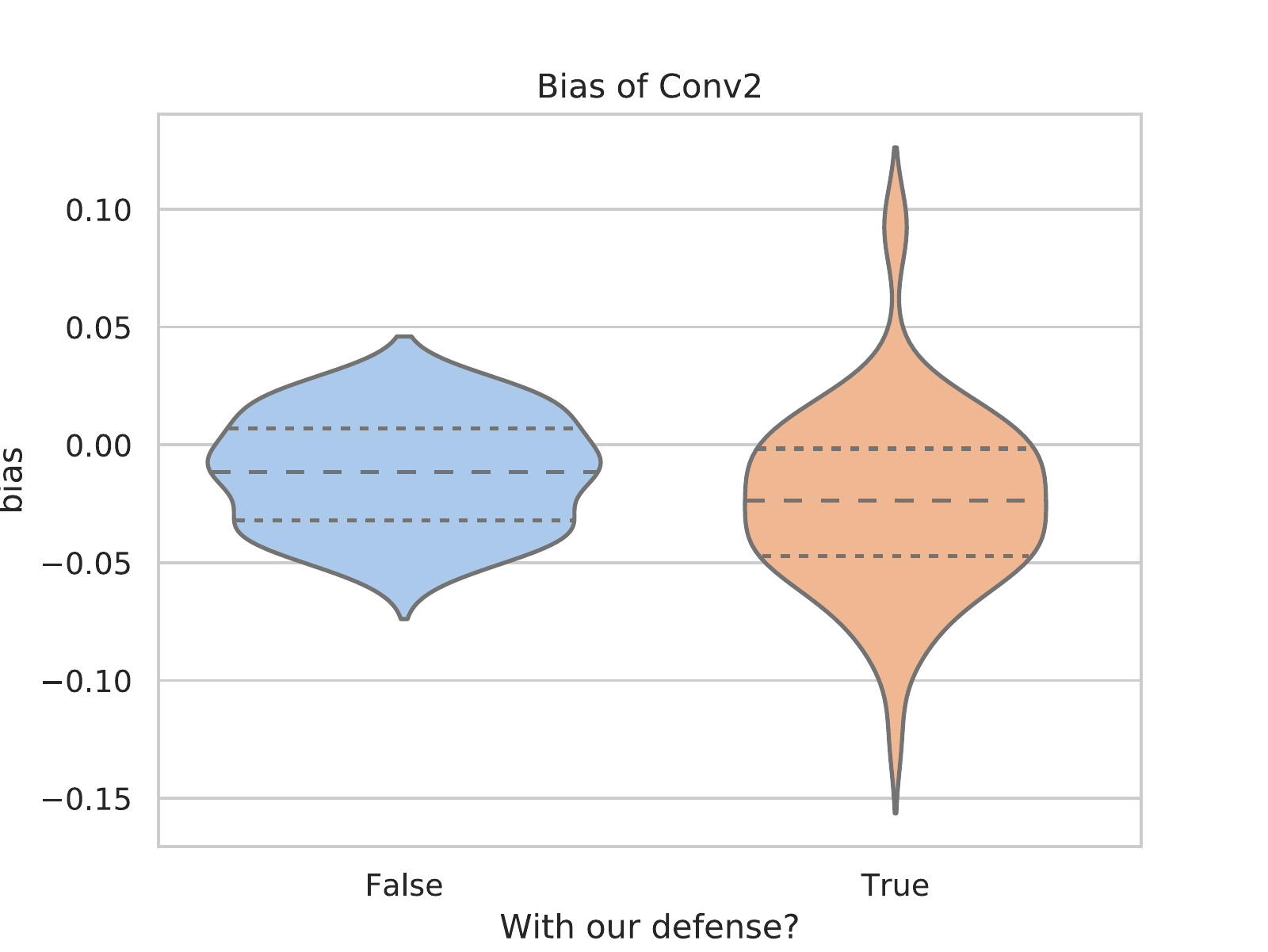}
	\endminipage
\caption{Weights and bias of the second convolutional layer (conv2)
	with/without our defense.}
\label{fig:conv2}
\end{figure}

\begin{figure}[h]
	\centering
	\minipage{0.49\columnwidth}
	\includegraphics[width=\columnwidth]{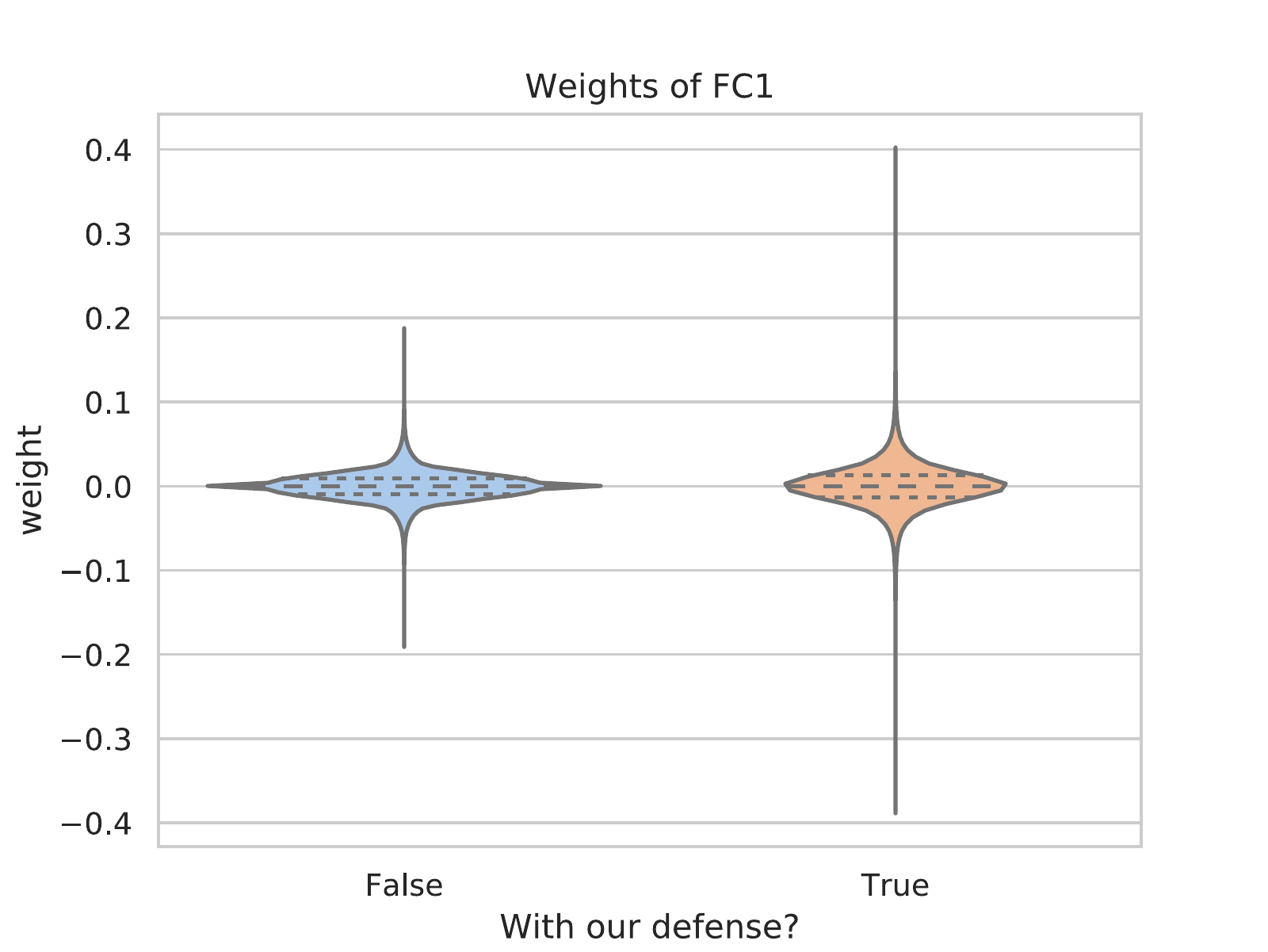}
	\endminipage\hfill
	\minipage{0.49\columnwidth}
	\includegraphics[width=\columnwidth]{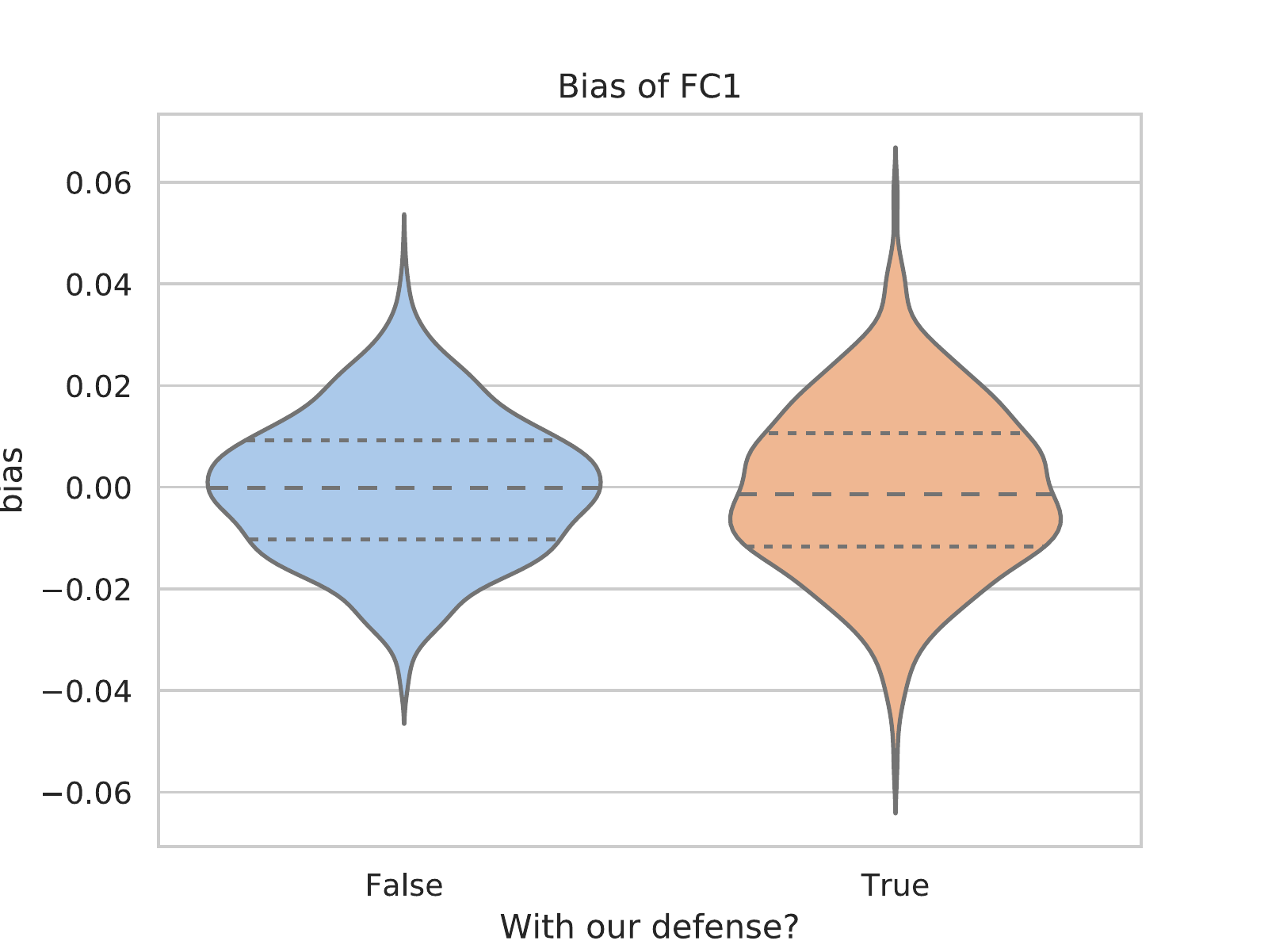}
	\endminipage
\caption{Weights and bias of the first fully-connected layer (fc1) with/without our defense.}
\label{fig:fc1}
\end{figure}

Ian \etal~\cite{fgsm} proposed a linear explanation of adversarial examples:
\begin{framed}
\begin{quote}
Consider the dot product between a weight vector $w$ and an adversarial
example $\tilde{x}=x+r~(\|r\|_\infty < \varepsilon)$:
\begin{equation}
	w^T\tilde{x} = w^Tx + w^Tr.
\end{equation}
The adversarial perturbation causes the activation to grow by $w^Tr$.
We can maximize this increase subject to the max norm constraint on $r$
by assigning $r=\varepsilon \text{sign}(w)$. If $w$ has $k$ dimensions
and the average magnitude of an element of the weight vector is $h$, then
the activation will grow by $\varepsilon k h$. Since $\|r\|_\infty$ does
not grow with the dimensionality of the problem but the change in activation
caused by perturbation by $r$ can grow linearly with $k$,
then for high dimensional problems, we can make many infinitesimal changes
to the input that add up to one large change to the output.
\end{quote}
\end{framed}

In our experiments, we analysed the parameters in models trained on MNIST,
as shown in Fig.~\ref{fig:conv1},~\ref{fig:conv2},~\ref{fig:fc1}.
These violin plots suggest that
\begin{itemize}
\item the weights in the first convolution layer of the defensive
model are closer to $0$ and have smaller variance than those
of the vanilla model. That means the $h$ in the above quotation is
decreased with our defense, and it will be harder for the $r$
to incur a large increase in activation $\varepsilon k h$.
\item the bias in the first convolution layer of the defensive model tend
to be negative values instead of being nearly ``zero-mean''. These negative
bias could help further suppress the increase in activation caused by
purturbation $r$.
\end{itemize}
Therefore, Ian \etal's~\cite{fgsm} theory could explain why our defense works,
as the resulting differences in network parameters may help reduce the
probability for adversarial perturbation
to increase the layer outputs into the local linear area of ReLU.

\newpage\clearpage
\section{Alternative Attack}

\subsection{Distance-based Ranking Attack}

In order to implement the proposed ranking attack, alternative
attacking objectives are possible.
Some related works such as \textbf{Feature Adversary}~\cite{featureadversary} generates
untargeted adversarial examples against classifiers by maximizing the distance
shift of representation vectors off their original locations.
This may inspire an alternative version of CA or QA objective functions which
are directly based on distance. For example, such alternative objective for
CA+ and QA- could be as follows:
%
\begin{align}
	r&=\argmin_{r\in\Gamma}\sum_{q\in Q}d(q,c+r)\\
	r&=\argmax_{r\in\Gamma}\sum_{c\in C}d(q+r,c).
\end{align}
%
However, it must be pointed out that
our method significantly differs from feature adversary~\cite{featureadversary}:
\textbf{(1)} Feature adversary concerns the \emph{pairwise} similarity of source-target
representations, while our image ranking problem concerns the ranking
order of multiple candidates;
\textbf{(2)} Feature adversary attempts to reduce the $\ell_2$ distance as much as
possible, while our triplet-like loss attempts to make positive candidates
closer to query than the negative ones, which well fits the objective of ranking
order optimization. Such \emph{relative} distance optimization
becomes more important since our attack simultaneously involves multiple queries
and multiple candidates;
\textbf{(3)} The $\ell_2$ distance based methods suffer from inevitable disadvantages.
Specifically, distance-based objectives are suboptimal,
because they disregard the relative
positions among the candidates and queries.

\begin{figure}[t]
	\centering
	\includegraphics[width=1.0\columnwidth]{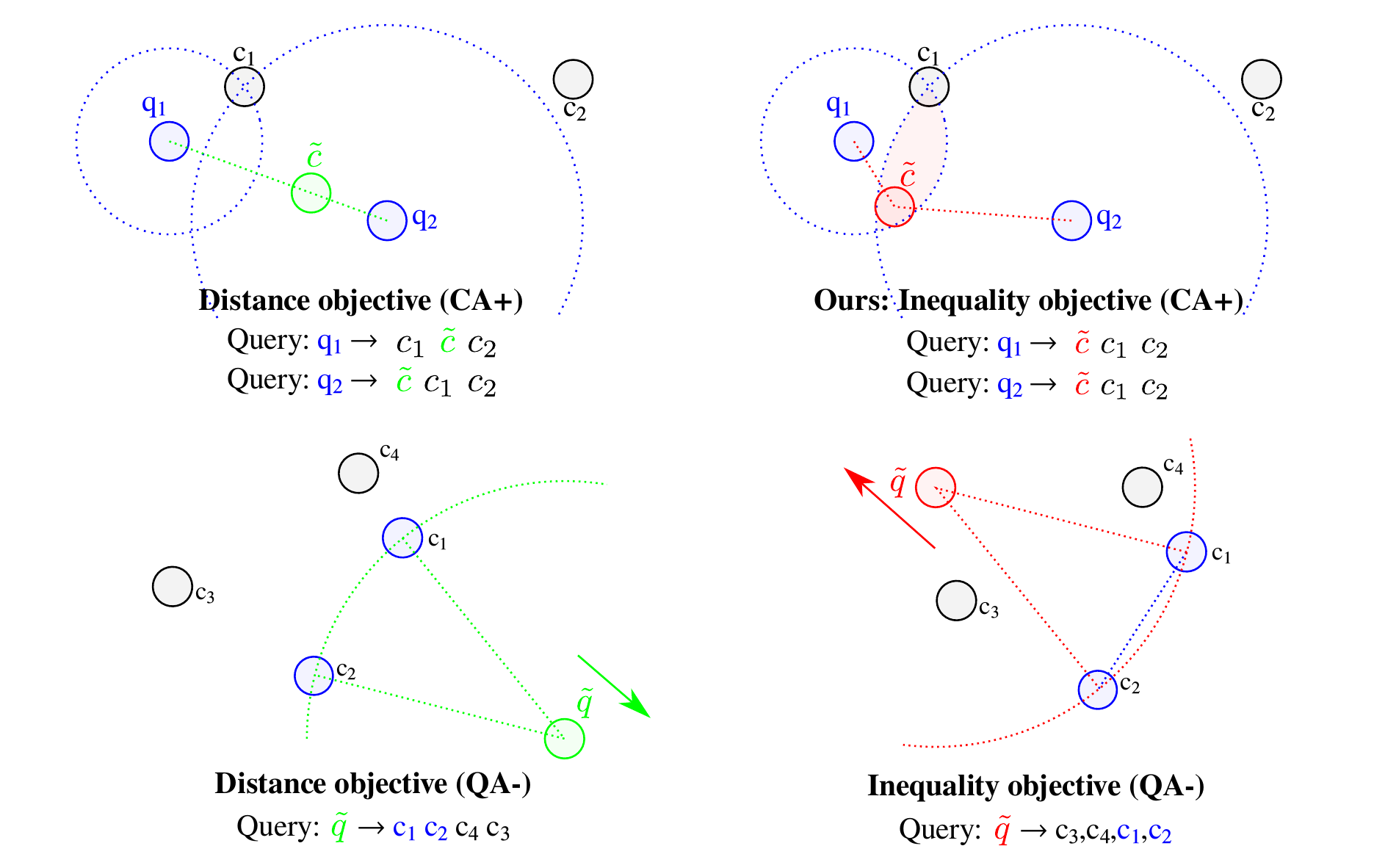}
	\caption{Distance-based CA+ (top-left) and QA- (bottom-left) objective
	functions do not lead to desired attacking result compared to our inequality-based
	method (top-right and bottom-right). In the top-left diagram for CA+, $q_1$ and $q_2$
	are the chosen queries, while $\tilde{c}$ is the adversarial candidate found
	by the distance objective. In the bottom-left
	diagram of QA-, $c_1$ and $c_2$ are the chosen candidates, while $\tilde{q}$ is
	the adversarial query found by the distance objective.
	It is noted that the distance objective is not optimal as shown in the top-left
	part, where the solution set for minimizing the distance objective contains
	suboptimal results. Distance-based objective
	may even fail to change the ranking result as shown in the bottom-left part,
	where optimizing the objective further cannot change the ranking result.
	In contrast, our proposed
	inequality-based method does not suffer from these issues, as shown in
	the top-right and bottom-right parts.}

	\label{fig:ineqdist}
\end{figure}

As shown in the top-left part of Fig.~\ref{fig:ineqdist}, the solution set for
for distance-based CA+ (the green dotted line) contains suboptimal solutions.
Similarly, in the bottom-left part of Fig.~\ref{fig:ineqdist} the
distance-based objective for QA- tends to maximize the sum of distance neglecting
the ranking result, and further optimization (moving $\tilde{q}$ along the green
arrow) will not change the ranking result.
In contrast, our proposed
inequality-based method does not suffer from these issues, as shown in
the top-right and bottom-right parts.

We also implemented such distance-based method and compared it with our
triplet-like method, as shown in Tab.~\ref{tab:additional1}. Experimental
results suggest that our method always outperforms distance-based method
by a margin. Especially for QA-, the distance-based objective is very difficult
to optimize because the distance-based Semantics-Preserving term contradicts
with the other term in the loss function. In summary, distance-based method
is not well-suited for our proposed \textit{adversarial ranking attack},
especially in the scenario of QA-.

\begin{table}[h]
\centering
\resizebox{1.0\columnwidth}{!}{%
\setlength{\tabcolsep}{0.2em}
\begin{tabular}{c|cccc|cccc|cccc|cccc}
\toprule
\multirow{2}{*}{$\varepsilon$} & \multicolumn{4}{c|}{CA+} & \multicolumn{4}{c|}{CA-} & \multicolumn{4}{c|}{QA+} & \multicolumn{4}{c}{QA-}\tabularnewline
\cline{2-17} \cline{3-17} \cline{4-17} \cline{5-17} \cline{6-17} \cline{7-17} \cline{8-17} \cline{9-17} \cline{10-17} \cline{11-17} \cline{12-17} \cline{13-17} \cline{14-17} \cline{15-17} \cline{16-17} \cline{17-17}
 & \multicolumn{1}{c|}{$w=1$} & \multicolumn{1}{c|}{$2$} & \multicolumn{1}{c|}{$5$} & $10$ & \multicolumn{1}{c|}{$w=1$} & \multicolumn{1}{c|}{$2$} & \multicolumn{1}{c|}{$5$} & $10$ & \multicolumn{1}{c|}{$m=1$} & \multicolumn{1}{c|}{$2$} & \multicolumn{1}{c|}{$5$} & $10$ & \multicolumn{1}{c|}{$m=1$} & \multicolumn{1}{c|}{$2$} & \multicolumn{1}{c|}{$5$} & $10$\tabularnewline
\hline
0.3 & 3.0 & 9.5 & 15.9 & 22.2 & 86.0 & 85.2 & 84.7 & 84.6 & 7.4 & 20.2 & 34.9 & 41.7 & 0.8 & 0.8 & 0.8 & 0.8\tabularnewline
\bottomrule
\end{tabular}
}
\caption{Attack Based on L-$2$ Distance Loss with MNIST.}
\label{tab:additional1}
\end{table}

\newpage\clearpage
\section{Alternative Defense}

Apart from the defense provided in the manuscript, we also tried some other
loss functions for adversarial training. In literature, there is no predominant choice
for the adversarial training loss function. The only common trait among these choices
is that all of them involve adversarial examples. Inspired by previous works,
we also implement some alternative defenses for ranking systems as follows.

\subsection{Straightforward Adaptation of Madry Defense}

Madry~\cite{madry} formularised improving neural network classifier robustness
as a min-max optimization problem, where the inner maximization
seeks to generate adversarial examples that lead to maximum cross-entropy
loss $L_\text{CE}$, while the outer minimization tunes the neural network parameters
$\theta$ to suppress the cross-entropy loss:
\begin{equation}
	\min_\theta \Big\{ \mathbb{E}_{(x,y)\sim D} \big[ \max_{r\in\Gamma} L_\text{CE} (x+r,y) \big] \Big\}
\end{equation}
where $(x,y)$ is a pair of image and ground-truth class label.

Similarly, we follow the idea and use a similar defense for ranking models:
\begin{equation}
	\min_\theta \Big\{ \mathbb{E}_{(q,c_p,c_n)\sim D} \big[ \max_{r\in\Gamma}L_{\text{triplet}}(q+r,c_p,c_n) \big] \Big\}
	\label{eq:mintmaxt}
\end{equation}
where the inner maximization aims to generate strongest adversarial
examples that could lead to triplet ranking error and a large loss value,
while the outer minimization seeks network parameters
that could reduce such error.

However, during experiments, we observe that such defensive loss function always
diverges, possibly due to the adversarial examples generated by the inner
problem being too ``strong''. We leave further investigation into this problem
for future work.

\subsection{Straightforward Adaptation of Ian Defense}

Ian~\cite{fgsm} proposed the following loss function for adversarial training:
\begin{align}
	L_\text{Ian}(x, y) &= \alpha L_\text{CE}(x,y) \\
	&+
	(1-\alpha)L_\text{CE}(x + \varepsilon \text{sign}(\nabla_x L_\text{CE}
	(x,y) ))
\end{align}
where the first term is a normal Cross-Entropy loss, and the second term
is the cross-entropy loss with untargeted adversarial example that aims
to increase the loss value. Constant $\alpha$ is a balancing parameter.

When adapted to a deep ranking system, this defense method also suffers
from the diverging issue similar to Madry's defense.

\begin{table}[t]
	\centering
\resizebox{1.0\columnwidth}{!}{
\setlength{\tabcolsep}{0.2em}

\begin{tabular}{c|cccc|cccc|cccc|cccc}
\toprule
\multirow{2}{*}{$\varepsilon$} & \multicolumn{4}{c|}{CA+} & \multicolumn{4}{c|}{CA-} & \multicolumn{4}{c|}{QA+} & \multicolumn{4}{c}{QA-}\tabularnewline
\cline{2-17} \cline{3-17} \cline{4-17} \cline{5-17} \cline{6-17} \cline{7-17} \cline{8-17} \cline{9-17} \cline{10-17} \cline{11-17} \cline{12-17} \cline{13-17} \cline{14-17} \cline{15-17} \cline{16-17} \cline{17-17}
 & \multicolumn{1}{c|}{$w=1$} & \multicolumn{1}{c|}{$2$} & \multicolumn{1}{c|}{$5$} & $10$ & \multicolumn{1}{c|}{$w=1$} & \multicolumn{1}{c|}{$2$} & \multicolumn{1}{c|}{$5$} & $10$ & \multicolumn{1}{c|}{$m=1$} & \multicolumn{1}{c|}{$2$} & \multicolumn{1}{c|}{$5$} & $10$ & \multicolumn{1}{c|}{$m=1$} & \multicolumn{1}{c|}{$2$} & \multicolumn{1}{c|}{$5$} & $10$\tabularnewline
 \midrule
\rowcolor{black!10}\multicolumn{17}{c}{(CCD) Cosine Distance, Contrastive Loss, Defensive (R@1=95.3\%)}\tabularnewline
\hline
0 & 50 & 50 & 50 & 50 & 1.2 & 1.2 & 1.2 & 1.2 & 50 & 50 & 50 & 50 & 0.5 & 0.5 & 0.5 & 0.5\tabularnewline
\hline
0.01 & 49.1 & 49.3 & 49.6 & 49.5 & 1.3 & 1.3 & 1.3 & 1.3 & 49.7,~0.0 & 49.9,~0.0 & 50.0,~0.0 & 49.8,~0.0 & 0.5,~0.0 & 0.5,~0.0 & 0.5,~0.0 & 0.5,~0.0\tabularnewline
\hline
0.03 & 48.0 & 48.2 & 48.5 & 48.6 & 1.6 & 1.5 & 1.5 & 1.5 & 48.7,~0.0 & 49.2,~0.0 & 49.7,~0.0 & 49.7,~0.0 & 0.6,~0.0 & 0.5,~0.0 & 0.5,~0.0 & 0.5,~0.0\tabularnewline
\hline
0.1 & 43.1 & 44.4 & 45.2 & 45.5 & 2.4 & 2.3 & 2.1 & 2.1 & 45.3,~0.1 & 47.4,~0.1 & 48.6,~0.1 & 49.4,~0.1 & 0.8,~0.1 & 0.7,~0.1 & 0.6,~0.1 & 0.6,~0.1\tabularnewline
\hline
0.3 & \textbf{33.3} & \textbf{35.8} & \textbf{37.4} & \textbf{38.0} & \textbf{5.6} & \textbf{5.1} & \textbf{4.8} & \textbf{4.7} & \textbf{38.2},~0.3 & \textbf{42.3},~0.3 & \textbf{45.7},~0.3 & \textbf{47.6},~0.3 & \textbf{2.1},~0.4 & \textbf{1.7},~0.4 & \textbf{1.5},~0.4 & \textbf{1.5},~0.4\tabularnewline
\midrule
\rowcolor{black!10}\multicolumn{17}{c}{(CTD) Cosine Distance, Triplet Loss, Defensive (R@1=97.4\%)}\tabularnewline
\hline
0 & 50 & 50 & 50 & 50 & 1.5 & 1.5 & 1.5 & 1.5 & 50 & 50 & 50 & 50 & 0.5 & 0.5 & 0.5 & 0.5\tabularnewline
\hline
0.01 & 49.3 & 49.6 & 49.5 & 49.7 & 1.6 & 1.6 & 1.6 & 1.6 & 49.5,~0.0 & 49.7,~0.0 & 50.0,~0.0 & 50.0,~0.0 & 0.5,~0.0 & 0.5,~0.0 & 0.5,~0.0 & 0.5,~0.0\tabularnewline
\hline
0.03 & 48.2 & 48.3 & 48.8 & 48.8 & 1.8 & 1.8 & 1.8 & 1.7 & 49.3,~0.0 & 48.9,~0.0 & 49.5,~0.0 & 49.7,~0.0 & 0.6,~0.0 & 0.5,~0.0 & 0.5,~0.0 & 0.5,~0.0\tabularnewline
\hline
0.1 & 44.7 & 45.4 & 46.3 & 46.4 & 2.8 & 2.6 & 2.5 & 2.5 & 46.3,~0.1 & 47.4,~0.1 & 48.6,~0.1 & 49.2,~0.1 & 0.7,~0.1 & 0.7,~0.1 & 0.6,~0.1 & 0.6,~0.1\tabularnewline
\hline
0.3 & \textbf{35.5} & \textbf{38.5} & \textbf{40.3} & \textbf{40.9} & \textbf{5.7} & \textbf{5.3} & \textbf{5.1} & \textbf{5.0} & \textbf{39.3},~0.4 & \textbf{43.1},~0.3 & \textbf{46.0},~0.3 & \textbf{47.5},~0.3 & \textbf{1.8},~0.4 & \textbf{1.6},~0.4 & \textbf{1.4},~0.4 & \textbf{1.4},~0.4\tabularnewline
\bottomrule
\end{tabular}
	}
	 \caption{The loss that directly suppresses embedding shift distance with MNIST dataset.}
   \label{tab:trip-es}
\end{table}

\subsection{Directly Suppressing Shift Distance}

As discussed in the manuscript, another possible adversarial training
method could be to directly suppress the maximum shift distance of embedding
vectors, \ie:
\begin{small}
\begin{equation} L_{\text{trip-es}} = L_{\text{triplet}}(q,c_p,c_n) +
\sum_{x\in\{q,c_p,c_n\}} \big(\max_{r\in\Gamma} d(x+r, x)\big),
\label{eq:trip-es} \end{equation}
\end{small}
where the most severe distance shift incurred by adversarial perturbations is
explicitly suppressed, in addition to a standard ranking loss term.

Experimental results (Tab.~\ref{tab:trip-es}) show that cosine distance-based
ranking models are more robust with this defense.
However, we note that this loss may numerically explode on an
Euclidean distance-based embedding model, as a strong adversary can
gradually cause very large embedding shift distance.
%
To mitigate such divergence issue, we
also tried to add a balancing parameter
to greatly scale down the second term of the loss function,
but the instability problem was
still not alleviated.
%

Due to the lack of universality, we leave this alternative defense in supplementary
material as a pure discussion, and possible improvements as future work.

\end{document}